\documentclass[chapter,biber]{nowfnt} 

\usepackage[utf8]{inputenc}

\usepackage[leqno]{amsmath}
\usepackage{amssymb}
\usepackage{amsthm}
\usepackage{amsfonts}
\usepackage{epsfig}
\usepackage{latexsym}
\usepackage{url}
\usepackage{hyperref}

\usepackage{ifthen}
\usepackage{comment}
\usepackage{color}
\usepackage{caption}
\usepackage{subcaption}
\usepackage{bm}
\usepackage{algorithm}
\usepackage{algorithmicx}
\usepackage{algpseudocode}
\usepackage{multirow}

\newcommand*{\Scale}[2][4]{\scalebox{#1}{$#2$}}%

\graphicspath{{old_plots/}{new_plots/}{new_plots/old_plots/}}

\newcommand{\field}[1]{\mathbb{#1}}
\newcommand{\R}{\field{R}} 
\newcommand{\E}{\mathbb{E}} 
\renewcommand{\Re}{\R} 
\newcommand{\Xc}{\mathcal{X}}

\newcommand{\Sc}{\mathcal{S}}
\newcommand{\Ac}{\mathcal{A}}

\newcommand{\Prob}{\mathbb{P}}

\newcommand{\hist}{\mathbb{H}}
\newcommand{\regret}{\text{regret}}

\DeclareMathOperator*{\argmax}{\arg\!\max}

\title{A Tutorial on Thompson Sampling}

\maintitleauthorlist{
Daniel J. Russo \\
Columbia University
\and
Benjamin Van Roy \\
Stanford University
\and
Abbas Kazerouni \\
Stanford University
\and
Ian Osband \\
Google DeepMind
\and
Zheng Wen \\
Adobe Research
}

 \issuesetup
{%
copyrightowner={D.~J.~Russo, B.~Van~Roy, A.~Kazerouni, I.~Osband and Z.~Wen},
volume        = 11,
issue         = 1,
pubyear       = 2018,
isbn          = 978-1-68083-470-3,
eisbn         = 978-1-68083-471-0,
doi           = 10.1561/2200000070,
firstpage     = 1, 
lastpage      = 96 
}

\usepackage{mwe}

\author[1]{Daniel J. Russo}
\author[2]{Benjamin Van Roy}
\author[2]{Abbas Kazerouni}
\author[3]{Ian Osband}
\author[4]{Zheng Wen}

\affil[1]{Columbia University}
\affil[2]{Stanford University}
\affil[3]{Google DeepMind}
\affil[4]{Adobe Research}

\articledatabox{\nowfntstandardcitation}

\begin{document}

\makeabstracttitle

\begin{abstract}
Thompson sampling is an algorithm for online decision problems where actions are taken sequentially in a manner
that must balance between exploiting what is known to maximize immediate performance and investing to accumulate new information
that may improve future performance.  The algorithm addresses a broad range of problems in a computationally efficient manner
and is therefore enjoying wide use.  This tutorial covers the algorithm and its application, illustrating concepts
through a range of examples, including Bernoulli bandit problems, shortest path problems, product recommendation, assortment,
active learning with neural networks, and reinforcement learning in Markov decision processes.
Most of these problems involve complex information structures, where information revealed by
taking an action informs beliefs about other actions.  We will also discuss when and why Thompson sampling is or is not effective
and relations to alternative algorithms.
\end{abstract}

\newpage

$ $

\vspace{2in}

\begin{center}
In memory of Arthur F. Veinott, Jr.
\end{center}

\chapter{Introduction}

The multi-armed bandit problem has been the subject of decades of intense study in statistics, operations research, electrical engineering, computer science, and economics.
A ``one-armed bandit'' is a somewhat antiquated term for a slot machine, which tends to ``rob'' players of their money. The colorful name for our problem comes from a motivating story in which a gambler enters a casino and sits down at a slot machine with multiple levers, or arms, that can be pulled. When pulled, an arm produces a random payout drawn independently of the past. Because the distribution of payouts corresponding to each arm is not listed, the player can learn it only by experimenting. As the gambler learns about the arms' payouts, she faces a dilemma: in the immediate future she expects to earn more by \emph{exploiting} arms that yielded high payouts in the past, but by continuing to \emph{explore} alternative arms she may learn how to earn higher payouts in the future. Can she develop a sequential strategy for pulling arms that balances this tradeoff and maximizes the cumulative payout earned? The following Bernoulli bandit problem is a canonical example.
\begin{example}(\emph{Bernoulli Bandit}) \label{ex:bernoulli}
	Suppose there are $K$ actions, and when played, any action yields either a success or a failure. Action $k \in \{1,...,K\}$ produces a success with probability $\theta_k \in [0,1]$. The success probabilities $(\theta_1,..,\theta_K)$ are unknown to the agent, but are fixed over time, and therefore can be learned by experimentation. The objective, roughly speaking, is to maximize the cumulative number of successes over $T$ periods, where $T$ is relatively large compared to the number of arms $K$.

	The ``arms'' in this problem might represent different banner ads that can be displayed on a website. Users arriving at the site are shown versions of the website with different banner ads. A success is associated either with a click on the ad, or with a conversion (a sale of the item being advertised). The parameters $\theta_k$ represent either the click-through-rate or conversion-rate among the population of users who frequent the site. The website hopes to balance exploration and exploitation in order to maximize the total number of successes.

	A naive approach to this problem involves allocating some fixed fraction of time periods to exploration and in each such period sampling an arm uniformly at random, while aiming to select successful actions in other time periods.  We will observe that such an approach can be quite wasteful even for the simple Bernoulli bandit problem described above and can fail completely for more complicated problems.
\end{example}
Problems like the Bernoulli bandit described above have been studied in the decision sciences since the second world war, as they crystallize the fundamental trade-off between exploration and exploitation in sequential decision making. But the information revolution has created significant new opportunities and challenges, which have spurred a particularly intense interest in this problem in recent years. To understand this, let us contrast the Internet advertising  example given above with the problem of choosing a banner ad to display on a highway. A physical banner ad might be changed only once every few months, and once posted will be seen by every individual who drives on the road. There is value to experimentation, but data is limited, and the cost of of trying a potentially ineffective ad is enormous. Online, a different banner ad can be shown to each individual out of a large pool of users, and data from each such interaction is stored. Small-scale experiments are now a core tool at most leading Internet companies.


Our interest in this problem is motivated by this broad phenomenon. Machine learning is increasingly used to make rapid data-driven decisions.  While standard algorithms in supervised machine learning learn passively from historical data, these systems often drive the generation of their own training data through interacting with users. An online recommendation system, for example, uses historical data to optimize current recommendations, but the outcomes of these recommendations are then fed back into the system and used to improve future recommendations. As a result, there is enormous potential benefit in the design of algorithms that not only learn from past data, but also explore systemically to generate useful data that improves future performance. There are significant challenges in extending algorithms designed to address Example \ref{ex:bernoulli} to treat more realistic and complicated decision problems. To understand some of these challenges, consider the problem of learning by experimentation to solve a shortest path problem.

\begin{example}(Online Shortest Path)\label{ex:shortest-path}
	An agent commutes from home to work every morning. She would like to commute along the path that requires the least average travel time, but she is uncertain of the travel time along different routes. How can she learn efficiently and minimize the total travel time over a large number of trips?

\begin{figure}[htpb]
\centering
\includegraphics[width=4in]{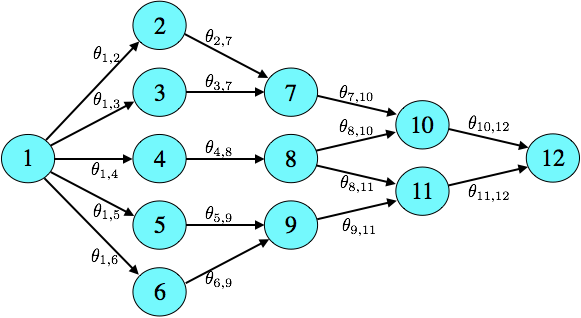}
\caption{Shortest path problem.}
\label{fig:shortest-path}
\end{figure}

	We can formalize this as a shortest path problem on a graph $G=(V, E)$ with vertices $V=\{1,...,N\}$ and edges $E$.  An example is illustrated in Figure \ref{fig:shortest-path}.  Vertex $1$ is the source (home) and vertex $N$ is the destination (work). Each vertex can be thought of as an intersection, and for two vertices $i,j\in V$, an edge $(i,j) \in E$ is present if there is a direct road connecting the two intersections. Suppose that traveling along an edge $e\in E$ requires time $\theta_e$ on average. If these parameters were known, the agent would select a path $(e_1,..,e_n)$, consisting of a sequence of adjacent edges connecting vertices $1$ and $N$, such that the expected total time $\theta_{e_1}+...+\theta_{e_n}$ is minimized. Instead, she chooses paths in a sequence of periods. In period $t$, the realized time $y_{t,e}$ to traverse edge $e$ is drawn independently from a distribution with mean $\theta_e$. The agent sequentially chooses a path $x_t$, observes the realized travel time $(y_{t,e})_{e\in x_t}$ along each edge in the path, and incurs cost $c_t = \sum_{e\in x_t} y_{t,e}$ equal to the total travel time. By exploring intelligently, she hopes to minimize cumulative travel time $\sum_{t=1}^{T} c_t$ over a large number of periods $T$.

	This problem is conceptually similar to the Bernoulli bandit in Example \ref{ex:bernoulli}, but here the number of actions is the number of paths in the graph, which generally scales exponentially in the number of edges. This raises substantial challenges. For moderate sized graphs, trying each possible path would require a prohibitive number of samples, and algorithms that require enumerating and searching through the set of all paths to reach a decision will be computationally intractable. An efficient approach therefore needs to leverage the statistical and computational structure of problem.

	 In this model, the agent observes the travel time along each edge traversed in a given period. Other feedback models are also natural: the agent might start a timer as she leaves home and checks it once she arrives, effectively only tracking the total travel time of the chosen path. This is closer to the Bernoulli bandit model, where only the realized reward (or cost) of the chosen arm was observed. We have also taken the random edge-delays $y_{t,e}$ to be independent, conditioned on $\theta_e$.  A more realistic model might treat these as correlated random variables, reflecting that neighboring roads are likely to be congested at the same time.  Rather than design a specialized algorithm for each possible statistical model, we seek a general approach to exploration that accommodates flexible modeling and works for a broad array of problems. We will see that Thompson sampling accommodates such flexible modeling, and offers an elegant and efficient approach to exploration in a wide range of structured decision problems, including the shortest path problem described here.
\end{example}
Thompson sampling -- also known as {\it posterior sampling} and {\it probability matching} -- was first proposed in 1933 \citep{thompson1933, thompson1935theory} for allocating experimental effort in two-armed bandit problems arising in clinical trials. The algorithm was largely ignored in the academic literature until recently, although it was independently rediscovered several times in the interim \citep{wyatt1997exploration, strens2000bayesian} as an effective heuristic. Now, more than eight decades after it was introduced, Thompson sampling has seen a surge of interest among industry practitioners and academics. This was spurred partly by two influential articles that displayed the algorithm's strong empirical performance \citep{chapelle2011empirical, scott2010modern}. In the subsequent five years, the literature on Thompson sampling has grown rapidly. Adaptations of Thompson sampling have now been successfully applied in a wide variety of domains, including revenue management \citep{ferreira2016online}, marketing \citep{schwartz2017customer}, web site optimization \citep{hill2017}, Monte Carlo tree search \citep{bai2013bayesian}, A/B testing \citep{graepel2010web}, Internet advertising \citep{graepel2010web, agarwal2013computational, agarwal2014laser}, recommendation systems \citep{kawale2015efficient}, hyperparameter tuning \citep{kandasamy2018parallel}, and arcade games \citep{osband2016deep}; and have been used at several companies, including Adobe, Amazon \citep{hill2017}, Facebook, Google \citep{scott2010modern, scott2015multi}, LinkedIn \citep{agarwal2013computational,agarwal2014laser}, Microsoft \citep{graepel2010web}, Netflix, and Twitter.

The objective of this tutorial is to explain when, why, and how to apply Thompson sampling.
A range of examples are used to demonstrate how the algorithm can be used to solve a variety of problems and provide clear insight into why it works and when it offers substantial benefit over naive alternatives.
The tutorial also provides guidance on approximations to Thompson sampling that can simplify computation as well as practical considerations like prior distribution specification, safety constraints and nonstationarity.
Accompanying this tutorial we also release a Python package\footnote{Python code and documentation is available at \url{https://github.com/iosband/ts_tutorial}.} that reproduces all experiments and figures presented.
This resource is valuable not only for reproducible research, but also as a reference implementation that may help practioners build intuition for how to practically implement some of the ideas and algorithms we discuss in this tutorial.  A concluding section discusses theoretical results that aim to develop an understanding of why Thompson sampling works, highlights settings
where Thompson sampling performs poorly, and discusses alternative approaches studied in recent literature.
As a baseline and backdrop for our discussion of Thompson sampling, we begin with an alternative approach that does not actively explore.

\chapter{Greedy Decisions}

\begin{figure}[htpb]
\centering
\includegraphics[width=4.5in]{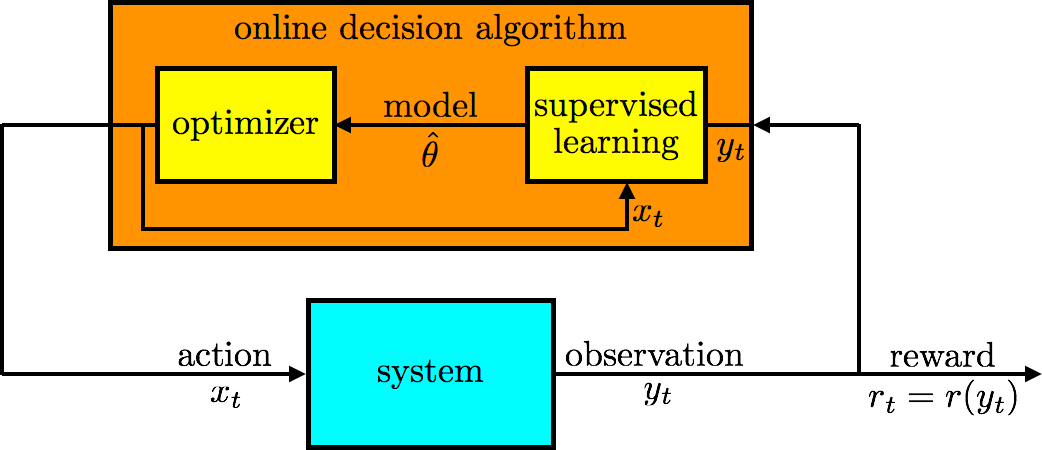}
\caption{Online decision algorithm.}
\label{fig:online-learning}
\end{figure}

{\it Greedy} algorithms serve as perhaps the simplest and most common approach to online decision problems.
The following two steps are taken to generate each action: (1) estimate a model
from historical data and (2) select the action that is optimal for the estimated model, breaking ties in an arbitrary manner.
Such an algorithm is greedy in the sense that an action is chosen solely to maximize immediate reward.
Figure \ref{fig:online-learning} illustrates such a scheme.  At each time $t$, a supervised learning algorithm fits
a model to historical data pairs $\hist_{t-1} = ((x_1,y_1), \ldots, (x_{t-1},y_{t-1}))$, generating an estimate $\hat{\theta}$
of model parameters.  The resulting model can then be used to predict the reward $r_t = r(y_t)$ from applying action
$x_t$.  Here, $y_t$ is an observed outcome, while $r$ is a known function that represents the agent's preferences.
Given estimated model parameters $\hat{\theta}$, an optimization algorithm selects the action $x_t$ that maximizes expected
reward, assuming that $\theta =\hat{\theta}$.  This action is then applied to the exogenous system and an outcome $y_t$ is observed.

A shortcoming of the greedy approach, which can severely curtail performance, is that it does not actively explore.
To understand this issue, it is helpful to focus on the Bernoulli bandit setting of Example \ref{ex:bernoulli}.  In that context,
the observations are rewards, so $r_t = r(y_t) = y_t$.  At each time $t$, a greedy algorithm would generate
an estimate $\hat{\theta}_k$ of the mean reward for each $k$th action, and
select the action that attains the maximum among these estimates.

\begin{figure}[htpb]
\centering
\includegraphics[scale=0.4]{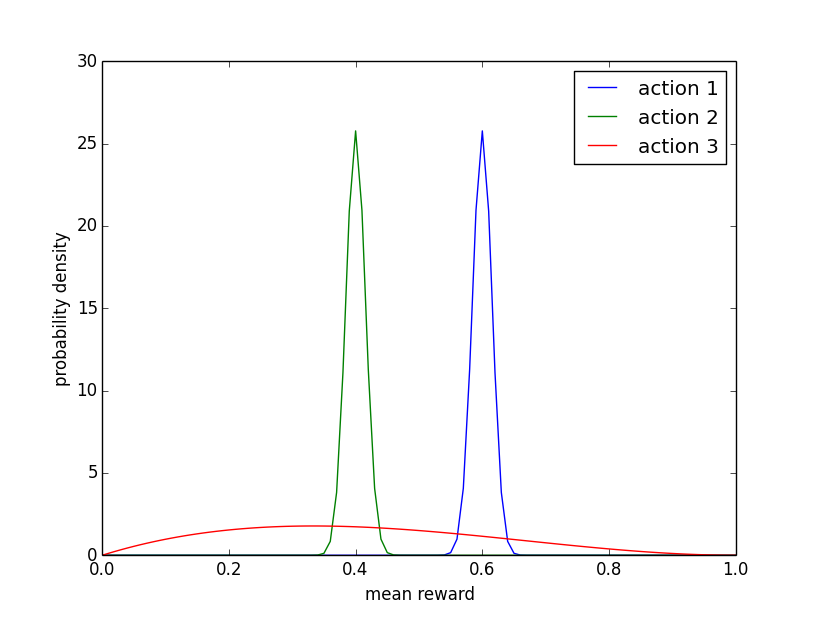}
\caption{Probability density functions over mean rewards.}
\label{fig:bernoulli-pdf}
\end{figure}

Suppose there are three actions with mean rewards $\theta \in \Re^3$.  In particular, each time
an action $k$ is selected, a reward of $1$ is generated with probability $\theta_k$.  Otherwise, a reward of
$0$ is generated.  The mean rewards are not known to the agent.  Instead, the agent's beliefs
in any given time period about these mean rewards can be expressed in terms of posterior distributions.
Suppose that, conditioned on the observed history
$\hist_{t-1}$, posterior distributions are represented by the probability density
functions plotted in Figure \ref{fig:bernoulli-pdf}.  These distributions represent beliefs after the
agent tries actions 1 and 2 one thousand times each, action 3 three times, receives cumulative rewards
of $600$, $400$, and $1$, respectively, and synthesizes these observations with uniform prior distributions
over mean rewards of each action.  They indicate that the agent is confident
that mean rewards for actions 1 and 2 are close to their expectations of approximately $0.6$ and
$0.4$.  On the other hand, the agent is highly uncertain about the mean reward of action 3,
though he expects $0.4$.

The greedy algorithm would select action 1, since that offers the maximal expected mean reward.
Since the uncertainty around this expected mean reward is small, observations are unlikely to change the expectation
substantially, and therefore, action 1 is likely to be selected {\it ad infinitum}.  It seems reasonable to avoid action 2,
since it is extremely unlikely that $\theta_2 > \theta_1$.  On the other hand,
if the agent plans to operate over many time periods, it should try action 3.  This is because
there is some chance that $\theta_3 > \theta_1$, and if this turns out to be the case,
the agent will benefit from learning that and applying action 3.  To learn whether $\theta_3 > \theta_1$, the agent
needs to try action 3, but the greedy algorithm will unlikely ever do that.
The algorithm fails to account for uncertainty in the mean reward of action
3, which should entice the agent to explore and learn about that action.

{\it Dithering} is a common approach to exploration that operates through randomly perturbing actions
that would be selected by a greedy algorithm.  One version of dithering, called
 {\it $\epsilon$-greedy exploration}, applies the greedy action with probability $1-\epsilon$
and otherwise selects an action uniformly at random.  Though this form of exploration can improve behavior
relative to a purely greedy approach, it wastes resources by failing to ``write off'' actions regardless of how unlikely
they are to be optimal.  To understand why, consider again the posterior distributions
of Figure \ref{fig:bernoulli-pdf}.  Action 2 has almost no chance of being optimal, and therefore, does not deserve
experimental trials, while the uncertainty surrounding action 3 warrants exploration.  However, $\epsilon$-greedy exploration
would allocate an equal number of experimental trials to each action.  Though only half of the
exploratory actions are wasted in this example, the issue is exacerbated as the number of possible actions
increases.  Thompson sampling, introduced more than eight decades ago \citep{thompson1933},
provides an alternative to dithering that more intelligently allocates exploration effort.

\chapter{Thompson Sampling for the Bernoulli Bandit}
\label{se:bernoulli}
To digest how Thompson sampling (TS) works, it is helpful to begin with a simple context that builds on the Bernoulli bandit of Example \ref{ex:bernoulli}
and incorporates a Bayesian model to represent uncertainty.
\begin{example}(Beta-Bernoulli Bandit)\label{ex:beta-bernoulli}
Recall the Bernoulli bandit of Example \ref{ex:bernoulli}.  There are $K$ actions. When played, an action $k$ produces a reward of one with probability $\theta_k$ and a reward of zero with probability $1-\theta_k$.
Each $\theta_k$ can be interpreted as an action's success probability or mean reward.  The mean rewards
$\theta=(\theta_1,...,\theta_K)$ are unknown, but fixed over time.  In the first period, an action $x_1$ is applied, and a reward $r_1 \in \{0,1\}$
 is generated with success probability $\mathbb{P}(r_1=1 | x_1, \theta) = \theta_{x_1}$.  After observing $r_1$, the agent applies another
 action $x_2$, observes a reward $r_2$, and this process continues.

Let the agent begin with an independent prior belief over each $\theta_k$.
Take these priors to be beta-distributed with parameters $\alpha = (\alpha_1,\ldots,\alpha_K)$ and $\beta \in (\beta_1,\ldots,\beta_K)$.
In particular, for each action $k$, the prior probability density function of $\theta_k$ is
$$p(\theta_k) = \frac{\Gamma(\alpha_k+\beta_k)}{\Gamma(\alpha_k)\Gamma(\beta_k)} \theta_k^{\alpha_k-1} (1-\theta_k)^{\beta_k-1},$$
where $\Gamma$ denotes the gamma function.
As observations are gathered, the distribution is updated according to Bayes' rule.  It is particularly convenient to work
with beta distributions because of their conjugacy properties.  In particular, each action's posterior distribution is also beta with parameters
that can be updated according to a simple rule:
\begin{align*}
 (\alpha_k, \beta_k) &\leftarrow
 \begin{cases}
 	(\alpha_k, \beta_k)    & \text{if } x_t \neq k \\
 	(\alpha_k, \beta_k)+(r_t, 1-r_t)      & \text{if } x_t =k.
 \end{cases}
 \end{align*}
\end{example}
Note that for the special case of $\alpha_k=\beta_k=1$, the prior $p(\theta_k)$ is uniform over $[0,1]$.
Note that only the parameters of a selected action are updated. The parameters $(\alpha_k, \beta_k)$ are sometimes
called pseudo-counts, since $\alpha_k$ or $\beta_k$ increases by one with each observed success or failure, respectively.
A beta distribution with parameters $(\alpha_k,\beta_k)$
has mean $\alpha_k/(\alpha_k + \beta_k)$, and the distribution becomes more concentrated as $\alpha_k+\beta_k$ grows.
Figure \ref{fig:bernoulli-pdf} plots probability density functions of beta distributions
with parameters $(\alpha_1, \beta_1) = (601,401)$, $(\alpha_2, \beta_2) = (401,601)$, and $(\alpha_3, \beta_3) = (2,3)$.

Algorithm \ref{alg:BernoulliGreedy} presents a greedy algorithm for the beta-Bernoulli bandit.
In each time period $t$, the algorithm generates an estimate $\hat{\theta}_k = \alpha_k/(\alpha_k+\beta_k)$, equal to its
current expectation of the success probability $\theta_k$.  The action $x_t$ with the largest estimate $\hat{\theta}_k$ is then applied, after which a
reward $r_t$ is observed and the distribution parameters $\alpha_{x_t}$ and $\beta_{x_t}$ are updated.

TS, specialized to the case of a beta-Bernoulli bandit, proceeds similarly,
as presented in  Algorithm \ref{alg:BernoulliTS}.  The only difference is that the success probability estimate
$\hat{\theta}_k$ is randomly sampled from the posterior distribution, which is a beta distribution
with parameters $\alpha_k$ and $\beta_k$, rather than taken to be the expectation $\alpha_k / (\alpha_k+\beta_k)$.
To avoid a common misconception, it is worth emphasizing TS does {\it not} sample $\hat{\theta}_k$
from the posterior distribution of the binary value $y_t$ that would be observed if action $k$ is selected.
In particular, $\hat{\theta}_k$ represents a statistically plausible success probability rather than a statistically
plausible observation.

\noindent
\begin{minipage}[t]{2.2in}
\vspace{0pt}
\begin{algorithm}[H]
\begin{scriptsize}
\caption{{\small  $\text{BernGreedy}(K, \alpha, \beta)$}}\label{alg:BernoulliGreedy}
\begin{algorithmic}[1]
\For{$t=1,2,\ldots $}
\State \textcolor{blue}{\#estimate model:}
\For{$k=1, \ldots, K$}
\State $\hat{\theta}_k \leftarrow \alpha_k / (\alpha_k + \beta_k)$
\EndFor \\
\State \textcolor{blue}{\#select and apply action:}
\State $x_t \leftarrow \argmax_k \hat{\theta}_k$
\State Apply $x_t$ and observe $r_t$ \\
\State \textcolor{blue}{\#update distribution:}
\State $(\alpha_{x_t}, \beta_{x_t}) \leftarrow  (\alpha_{x_t} + r_t, \beta_{x_t} + 1-r_t)$
\EndFor
\end{algorithmic}
\end{scriptsize}
\end{algorithm}
\end{minipage}%
\hspace{0.2in}
\begin{minipage}[t]{2.2in}
  \vspace{0pt}
\begin{algorithm}[H]
\begin{scriptsize}
\caption{{\small $\text{BernTS}(K, \alpha, \beta)$}}\label{alg:BernoulliTS}
\begin{algorithmic}[1]
\For{$t=1,2,\ldots $}
\State \textcolor{blue}{\#sample model:}
\For{$k=1, \ldots, K$}
\State Sample $\hat{\theta}_k \sim \text{beta}(\alpha_k, \beta_k)$
\EndFor \\
\State \textcolor{blue}{\#select and apply action:}
\State $x_t \leftarrow \argmax_k \hat{\theta}_k$
\State Apply $x_t$ and observe $r_t$ \\
\State \textcolor{blue}{\#update distribution:}
\State $(\alpha_{x_t}, \beta_{x_t}) \leftarrow  (\alpha_{x_t} + r_t, \beta_{x_t} + 1-r_t)$
\EndFor
\end{algorithmic}
\end{scriptsize}
\end{algorithm}
\end{minipage}
\vspace{0.3in}

To understand how TS improves on greedy actions with or without dithering, recall
the three armed Bernoulli bandit with posterior distributions illustrated in Figure \ref{fig:bernoulli-pdf}.
In this context, a greedy action would forgo the potentially valuable opportunity to learn about action 3.
With dithering, equal chances would be assigned to probing actions 2 and 3, though probing action 2 is
virtually futile since it is extremely unlikely to be optimal.  TS, on the other hand would
sample actions 1, 2, or 3, with probabilities approximately equal to $0.82$, $0$, and $0.18$, respectively.
In each case, this is the probability that the random estimate drawn for the action exceeds those
drawn for other actions.  Since these estimates are drawn from posterior distributions, each of
these probabilities is also equal to the probability that the corresponding action is optimal, conditioned
on observed history.  As such, TS explores to resolve uncertainty where
there is a chance that resolution will help the agent identify the optimal action, but avoids
probing where feedback would not be helpful.

\begin{figure}[htpb]
\centering
    \begin{subfigure}{.49\textwidth}
        \centering
	\includegraphics[width=\linewidth]{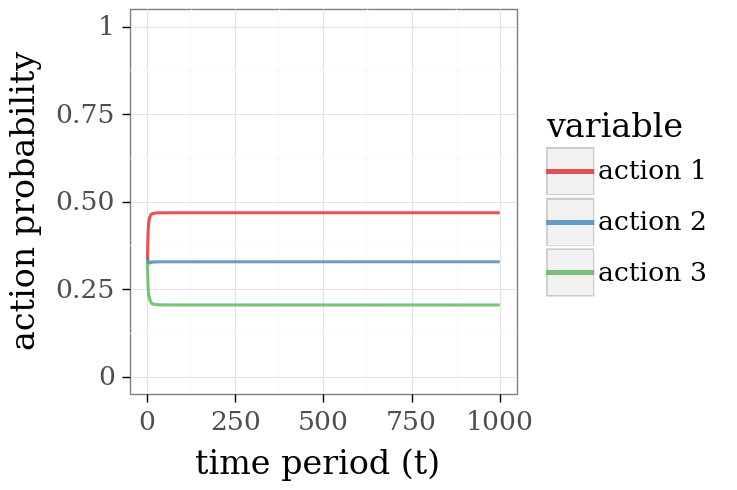}
	\caption{greedy algorithm}
	 \label{fig:bernoulli-action-probs-greedy}
    \end{subfigure}
    \begin{subfigure}{.49\textwidth}
        \centering
	\includegraphics[width=\linewidth]{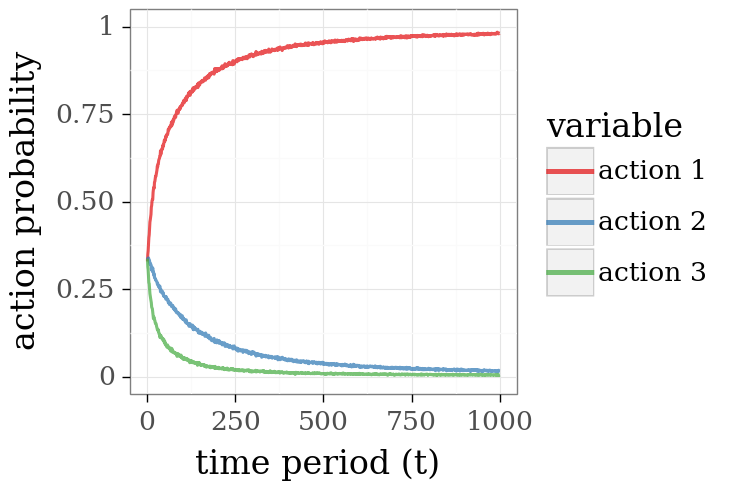}
	\caption{Thompson sampling}
	 \label{fig:bernoulli-action-probs-ts}
    \end{subfigure}
\caption{Probability that the greedy algorithm and Thompson sampling selects an action.}
\label{fig:bernoulli-action-probs}
\end{figure}

It is illuminating to compare simulated behavior of TS to that of a greedy algorithm.
Consider a three-armed beta-Bernoulli bandit with mean rewards $\theta_1=0.9$, $\theta_2=0.8$, and $\theta_3=0.7$.
Let the prior distribution over each mean reward be uniform.  Figure \ref{fig:bernoulli-action-probs}
plots results based on ten thousand independent simulations of each algorithm.
Each simulation is over one thousand time periods.  In each simulation, actions are randomly rank-ordered
for the purpose of tie-breaking so that the greedy algorithm is not biased toward selecting any particular action.
Each data point represents the fraction of simulations for which a particular action is selected at a particular time.

From the plots, we see that the greedy algorithm does not always converge on action 1, which is the optimal action.
This is because the algorithm can get stuck, repeatedly applying a poor action.  For example, suppose the algorithm
applies action 3 over the first couple time periods and receives a reward of $1$ on both occasions.
The algorithm would then continue to select action 3, since the expected mean reward
of either alternative remains at $0.5$.  With repeated selection of action 3, the expected mean reward
converges to the true value of $0.7$, which reinforces the agent's commitment to action 3.
TS, on the other hand, learns to select action 1 within the thousand periods.
This is evident from the fact that, in an overwhelmingly large fraction of simulations, TS selects action 1
in the final period.

\begin{figure}[h!]
\centering
    \begin{subfigure}{.49\textwidth}
        \centering
	\includegraphics[width=\linewidth]{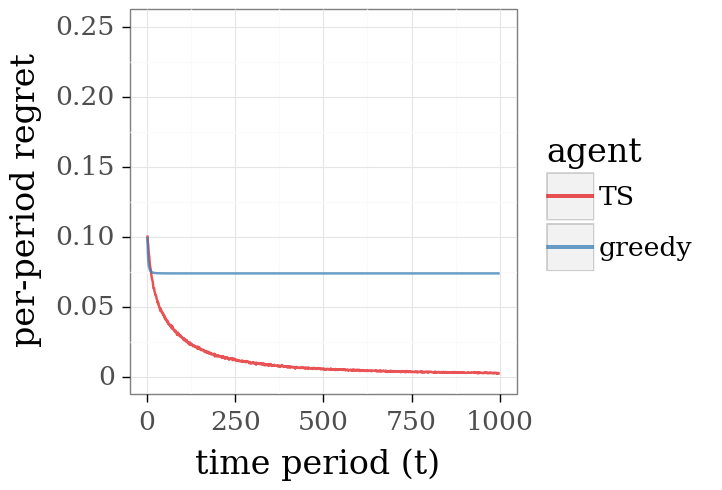}
	\caption{$\theta=(0.9,0.8,0.7)$}
	  \label{fig:bernoulli-regret-conditional}
    \end{subfigure}
    \begin{subfigure}{.49\textwidth}
        \centering
	\includegraphics[width=\linewidth]{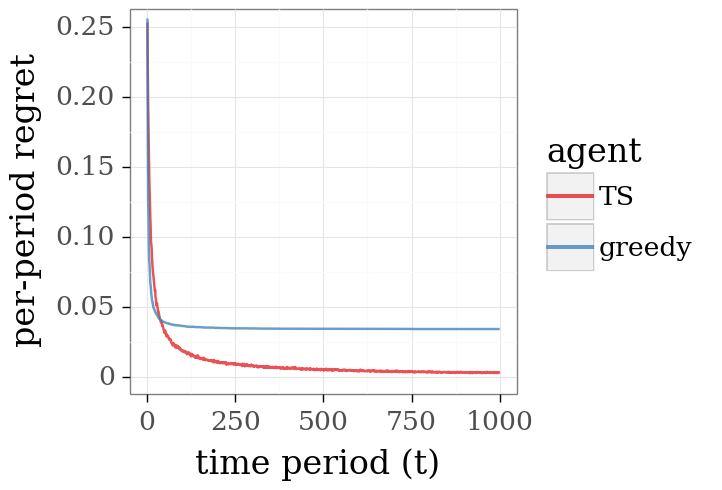}
	\caption{average over random $\theta$}
	\label{fig:bernoulli-regret-unconditional}
    \end{subfigure}
\caption{Regret from applying greedy and Thompson sampling algorithms to the three-armed Bernoulli bandit.}
\label{fig:bernoulli-regret}
\end{figure}

The performance of online decision algorithms is often studied and compared through plots of regret.
The {\it per-period regret} of an algorithm over a time period $t$ is the difference between the mean reward of an optimal
action and the action selected by the algorithm.  For the Bernoulli bandit problem, we can write this
as $\regret_t(\theta) = \max_k \theta_k - \theta_{x_t}$.
Figure \ref{fig:bernoulli-regret-conditional} plots per-period regret realized by the greedy algorithm and TS,
again averaged over ten thousand simulations.  The average per-period regret of TS
vanishes as time progresses.  That is not the case for the greedy algorithm.

Comparing algorithms with fixed mean rewards raises questions about the extent to which the results depend
on the particular choice of $\theta$.  As such, it is often useful to also examine regret averaged over plausible values of $\theta$.
A natural approach to this involves sampling many instances of $\theta$ from the prior distributions and generating an independent
simulation for each.  Figure \ref{fig:bernoulli-regret-unconditional} plots averages over ten thousand such simulations,
with each action reward sampled independently from a uniform prior for each simulation.  Qualitative features
of these plots are similar to those we inferred from Figure \ref{fig:bernoulli-regret-conditional}, though regret in Figure \ref{fig:bernoulli-regret-conditional}
is generally smaller over early time periods and larger over later time periods, relative to Figure \ref{fig:bernoulli-regret-unconditional}.
The smaller regret in early time periods is due to the fact that with $\theta = (0.9,0.8,0.7)$,
mean rewards are closer than for a typical randomly sampled $\theta$, and therefore the regret of randomly selected actions
is smaller.  The fact that per-period regret of TS is larger in Figure \ref{fig:bernoulli-regret-conditional} than Figure \ref{fig:bernoulli-regret-unconditional}
over later time periods, like period $1000$, is also a consequence of proximity among rewards with $\theta = (0.9,0.8,0.7)$.
In this case, the difference is due to the fact that it takes longer to differentiate actions than it would for a typical randomly sampled $\theta$.

\chapter{General Thompson Sampling}
\label{se:GeneralTS}

TS can be applied fruitfully to a broad array of online decision problems beyond the Bernoulli bandit, and
we now consider a more general setting.
Suppose the agent applies a sequence of actions $x_1, x_2, x_3,\ldots$ to a system, selecting each from a set
$\mathcal{X}$.  This action set could be finite, as in the case of the Bernoulli bandit, or infinite.
After applying action $x_t$, the agent observes an outcome
$y_t$, which the system randomly generates according to a conditional probability measure $q_\theta(\cdot | x_t)$.
The agent enjoys a reward $r_t = r(y_t)$, where $r$ is a known function.  The agent is initially uncertain about the
value of $\theta$ and represents his uncertainty using a prior distribution $p$.

Algorithms \ref{alg:GeneralGreedy} and \ref{alg:GeneralTS} present greedy and TS approaches in an abstract
form that accommodates this very general problem.  The two differ in the way they generate model parameters $\hat{\theta}$.
The greedy algorithm takes $\hat{\theta}$ to be the expectation of $\theta$ with respect to the distribution $p$, while TS
draws a random sample from $p$.  Both algorithms then apply actions that maximize expected reward for their respective models.
Note that, if there are a finite set of possible observations $y_t$, this expectation is given by
\begin{equation}
\label{eq:finite-expectation}
\E_{q_{\hat{\theta}}}[r(y_t) | x_t = x] = \sum_o q_{\hat{\theta}}(o|x) r(o).
\end{equation}
The distribution $p$ is updated by conditioning on the realized observation $\hat{y}_t$.  If $\theta$
is restricted to values from a finite set, this conditional distribution can be written by Bayes rule as
\begin{equation}
\label{eq:finite-conditioning}
\mathbb{P}_{p, q}(\theta = u | x_t, y_t) = \frac{p(u) q_u(y_t | x_t)}{\sum_v p(v) q_v(y_t | x_t)}.
\end{equation}

\begin{centering}

\begin{minipage}[t]{2.2in}
\vspace{0pt}
\begin{algorithm}[H]
\begin{scriptsize}
\caption{{\small $\text{Greedy}(\mathcal{X}, p, q, r)$}}\label{alg:GeneralGreedy}
\begin{algorithmic}[1]
\For{$t=1,2,\ldots $}
\State \textcolor{blue}{\#estimate model:}
\State $\hat{\theta} \leftarrow \E_p[\theta]$ \\
\State \textcolor{blue}{\#select and apply action:}
\State $x_t \leftarrow \argmax_{x\in\mathcal{X}} \E_{q_{\hat{\theta}}}[r(y_t) | x_t = x]$
\State Apply $x_t$ and observe $y_t$ \\
\State \textcolor{blue}{\#update distribution:}
\State $p \leftarrow \mathbb{P}_{p, q}(\theta \in \cdot | x_t, y_t)$
\EndFor
\end{algorithmic}
\end{scriptsize}
\end{algorithm}
\end{minipage}%
\hspace{0.2in}
\begin{minipage}[t]{2.2in}
\vspace{0pt}
\begin{algorithm}[H]
\begin{scriptsize}
\caption{{\small $\text{Thompson}(\mathcal{X}, p, q, r)$}}\label{alg:GeneralTS}
\begin{algorithmic}[1]
\For{$t=1,2,\ldots $}
\State \textcolor{blue}{\#sample model:}
\State Sample $\hat{\theta} \sim p$ \\
\State \textcolor{blue}{\#select and apply action:}
\State $x_t \leftarrow \argmax_{x \in \mathcal{X}} \E_{q_{\hat{\theta}}}[r(y_t) | x_t =x]$
\State Apply $x_t$ and observe $y_t$ \\
\State \textcolor{blue}{\#update distribution:}
\State $p \leftarrow \mathbb{P}_{p, q}(\theta \in \cdot | x_t, y_t)$
\EndFor
\end{algorithmic}
\end{scriptsize}
\end{algorithm}
\end{minipage}

\end{centering}
\vspace{0.3in}

The Bernoulli bandit with a beta prior serves as a special case of this more general formulation.  In this special case, the set of actions
is $\mathcal{X} = \{1,\ldots,K\}$ and only rewards are observed, so $y_t = r_t$.  Observations and rewards are modeled by conditional probabilities
$q_\theta(1|k) = \theta_k$ and $q_\theta(0|k) = 1-\theta_k$.  The prior distribution is encoded by vectors $\alpha$ and $\beta$,
with probability density function given by:
$$p(\theta) = \prod_{k=1}^K \frac{\Gamma(\alpha+\beta)}{\Gamma(\alpha_k)\Gamma(\beta_k)} \theta_k^{\alpha_k-1} (1-\theta_k)^{\beta_k-1},$$
where $\Gamma$ denotes the gamma function.
In other words, under the prior distribution, components of $\theta$ are independent and beta-distributed, with parameters $\alpha$ and $\beta$.

For this problem, the greedy algorithm (Algorithm \ref{alg:GeneralGreedy}) and TS (Algorithm \ref{alg:GeneralTS})
begin each $t$th iteration with posterior parameters $(\alpha_k,\beta_k)$ for $k \in \{1,\ldots,K\}$.  The greedy algorithm sets $\hat{\theta}_k$ to the expected value
$\E_p[\theta_k] = \alpha_k/(\alpha_k+\beta_k)$, whereas TS randomly draws $\hat{\theta}_k$ from
a beta distribution with parameters $(\alpha_k,\beta_k)$.  Each algorithm then selects the action $x$ that maximizes
$\E_{q_{\hat{\theta}}}[r(y_t) | x_t = x] = \hat{\theta}_x$.
After applying the selected action, a reward $r_t = y_t$ is observed, and belief distribution parameters are updated according to
$$(\alpha, \beta) \leftarrow (\alpha + r_t {\bf 1}_{x_t}, \beta + (1-r_t) {\bf 1}_{x_t}),$$
where ${\bf 1}_{x_t}$ is a vector with component $x_t$ equal to $1$ and all other components equal to $0$.

Algorithms \ref{alg:GeneralGreedy} and \ref{alg:GeneralTS} can also be applied to much more complex problems.
As an example, let us consider a version of the shortest path problem presented in Example \ref{ex:shortest-path}.
\begin{example}(Independent Travel Times)\label{ex:log-shortest-path}
Recall the shortest path problem of Example \ref{ex:shortest-path}.
The model is defined with respect to a directed graph $G = (V, E)$, with vertices
$V = \{1,\ldots, N\}$, edges $E$, and mean travel times $\theta \in \Re^{N}$.  Vertex $1$ is the source and vertex
$N$ is the destination.  An action is a sequence of distinct edges leading from source to destination.  After applying
action $x_t$, for each traversed edge $e \in x_t$, the agent observes a travel time $y_{t,e}$ that is independently
sampled from a distribution with mean $\theta_e$.  Further, the agent incurs
a cost of $\sum_{e \in x_t} y_{t,e}$, which can be thought of as a reward $r_t = -\sum_{e \in x_t} y_{t,e}$.

Consider a prior for which each $\theta_e$ is independent and log-Gaussian-distributed with parameters $\mu_e$ and $\sigma^2_e$. That is, $\ln(\theta_e) \sim N(\mu_e, \sigma_e^2)$ is Gaussian-distributed. Hence, $\E[\theta_e] = e^{\mu_e + \sigma_e^2/2}$.  Further, take $y_{t,e}|\theta$ to be independent across edges $e \in E$
and log-Gaussian-distributed with parameters $\ln(\theta_e) - \tilde{\sigma}^2/2$ and $\tilde{\sigma}^2$, so that $\E[y_{t,e}|\theta_e] = \theta_e$.  Conjugacy
properties accommodate a simple rule for updating the distribution of $\theta_e$
upon observation of $y_{t,e}$:
\begin{equation}
\label{eq:lognormal-update}
(\mu_e, \sigma_e^2) \leftarrow \left(\frac{\frac{1}{\sigma_e^2} \mu_e + \frac{1}{\tilde{\sigma}^2} \left(\ln(y_{t,e}) +\frac{\tilde{\sigma}^2}{2}\right)}{\frac{1}{\sigma_e^2} + \frac{1}{\tilde{\sigma}^2}},  \frac{1}{\frac{1}{\sigma_e^2} + \frac{1}{\tilde{\sigma}^2}}\right).
\end{equation}
\end{example}

To motivate this formulation, consider an agent who commutes from
home to work every morning.  Suppose possible paths are represented by a graph $G = (V, E)$.  Suppose the agent knows
the travel distance $d_e$ associated with each edge $e \in E$ but is uncertain about average travel times.
It would be natural for her to construct a prior for which expectations are equal to travel distances.
With the log-Gaussian prior, this can be accomplished by setting $\mu_e = \ln(d_e) -\sigma_e^2/2$.
Note that the parameters $\mu_e$ and $\sigma_e^2$ also express a degree of uncertainty; in particular, the prior variance
of mean travel time along an edge is $(e^{\sigma_e^2}-1) d_e^2$.

The greedy algorithm (Algorithm \ref{alg:GeneralGreedy}) and TS (Algorithm \ref{alg:GeneralTS}) can be applied to
Example  \ref{ex:log-shortest-path} in a computationally efficient manner.  Each algorithm begins each $t$th iteration with
posterior parameters $(\mu_e,\sigma_e)$ for each $e \in E$.  The greedy algorithm sets $\hat{\theta}_e$ to the expected value
$\E_p[\theta_e] = e^{\mu_e + \sigma_e^2/2}$, whereas TS randomly draws $\hat{\theta}_e$ from
 a log-Gaussian distribution with parameters $\mu_e$ and $\sigma_e^2$.  Each algorithm then selects its action $x$
 to maximize $\E_{q_{\hat{\theta}}}[r(y_t) | x_t = x] = -\sum_{e \in x_t} \hat{\theta}_e$.
This can be cast as a deterministic shortest path problem, which can be solved efficiently, for example,
via Dijkstra's algorithm.  After applying the selected action, an outcome $y_t$ is observed, and belief distribution
parameters $(\mu_e, \sigma_e^2)$, for each $e \in E$, are updated according to (\ref{eq:lognormal-update}).

\begin{figure}[h!]
\centering
    \begin{subfigure}{.49\textwidth}
        \centering
	\includegraphics[width=\linewidth]{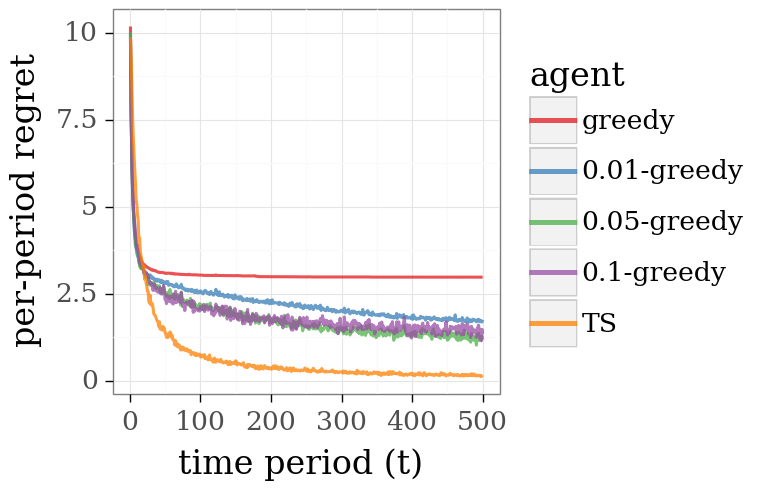}
	\caption{regret}
	  \label{fig:shortest_path_regret_lognorm}
    \end{subfigure}
    \begin{subfigure}{.49\textwidth}
        \centering
	\includegraphics[width=\linewidth]{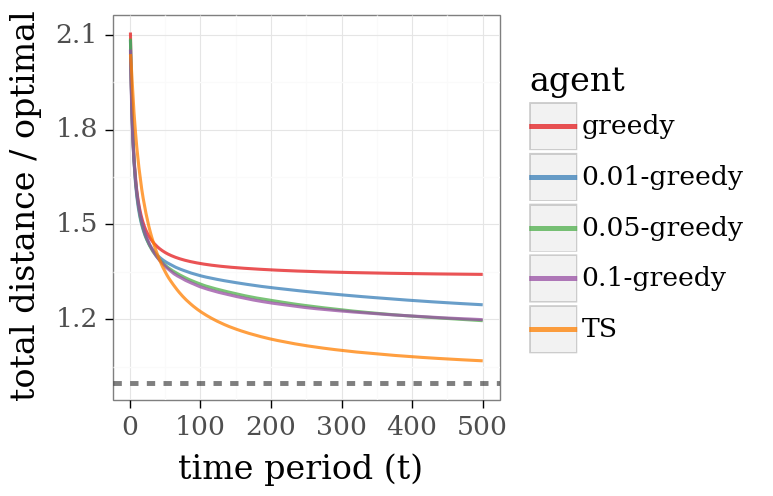}
	\caption{cumulative travel time vs. optimal}
	\label{fig:shortest_path_ratio_lognorm}
    \end{subfigure}
\caption{Performance of Thompson sampling and $\epsilon$-greedy algorithms in the shortest path problem.}
\label{fig:shortest_path_lognorm}
\end{figure}

\begin{figure}[htpb]
\centering
\includegraphics[scale=0.4]{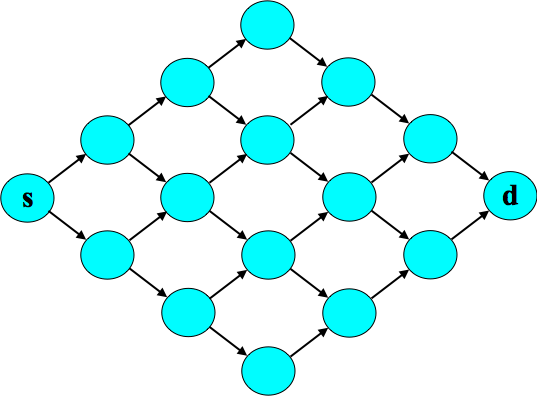}
\caption{A binomial bridge with six stages.}
\label{fig:binomial-bridge}
\end{figure}

Figure \ref{fig:shortest_path_lognorm} presents results from applying greedy and TS
algorithms to Example \ref{ex:log-shortest-path},
with the graph taking the form of a binomial bridge, as shown in Figure \ref{fig:binomial-bridge},
except with twenty rather than six stages, so there are 184,756 paths from source to destination.
Prior parameters are set to $\mu_e = -\frac{1}{2}$ and $\sigma_e^2 = 1$ so that $\mathbb{E}[\theta_e] = 1$, for each $e \in E$, and
the conditional distribution parameter is $\tilde{\sigma}^2 = 1$.  Each data point represents an average
over ten thousand independent simulations.

The plots of regret demonstrate that the performance of TS converges quickly to optimal,
while that is far from true for the greedy algorithm.
We also plot results generated by $\epsilon$-greedy exploration, varying $\epsilon$.  For each trip,
with probability $1-\epsilon$, this algorithm traverses a path produced by a greedy algorithm.
Otherwise, the algorithm samples a path randomly.  Though this form of exploration
can be helpful, the plots demonstrate that learning progresses at a far slower pace than
with TS.  This is because $\epsilon$-greedy exploration is not
judicious in how it selects paths to explore.  TS, on the other hand,
orients exploration effort towards informative rather than entirely random paths.

Plots of cumulative travel time relative to optimal offer a sense for the fraction of driving time wasted due
to lack of information.  Each point plots an average of the ratio between the time incurred
over some number of days and the minimal expected travel time given $\theta$.  With TS,
this converges to one at a respectable rate.  The same can not be said for $\epsilon$-greedy approaches.

Algorithm \ref{alg:GeneralTS} can be applied to problems with complex information structures,
and there is often substantial value to careful modeling of such structures.
As an example, we consider a more complex variation of the binomial bridge example.
\begin{example}(Correlated Travel Times)\label{ex:log-shortest-path-cf}
As with Example \ref{ex:log-shortest-path}, let each $\theta_e$ be independent and log-Gaussian-distributed with
parameters $\mu_e$ and $\sigma_e^2$.  Let the observation distribution be characterized by
$$y_{t,e} = \zeta_{t,e} \eta_t \nu_{t,\ell(e)} \theta_e,$$
where each $\zeta_{t,e}$ represents an idiosyncratic factor associated with edge $e$,
$\eta_t$ represents a factor that is common to all edges,
$\ell(e)$ indicates whether edge $e$ resides in the lower half of the binomial bridge,
and $\nu_{t,0}$ and $\nu_{t,1}$ represent factors that bear a common influence on edges in the upper and lower halves,
respectively.  We take each $\zeta_{t,e}$, $\eta_t$, $\nu_{t,0}$, and $\nu_{t,1}$ to be independent log-Gaussian-distributed with parameters
$-\tilde{\sigma}^2/6$ and $\tilde{\sigma}^2/3$.  The distributions of the shocks $\zeta_{t,e}$, $\eta_t$, $\nu_{t,0}$ and $\nu_{t,1}$ are known, and only the parameters $\theta_e$ corresponding to each individual edge must be learned through experimentation. Note that, given these parameters, the marginal distribution of $y_{t,e} | \theta$ is identical to that of Example \ref{ex:log-shortest-path}, though
the joint distribution over $y_t | \theta$ differs.

The common factors induce correlations among travel times in the binomial bridge:
 $\eta_t$ models the impact of random events that influence traffic
conditions everywhere, like the day's weather, while $\nu_{t,0}$ and $\nu_{t,1}$
each reflect events that bear influence only on traffic conditions along edges in half of the binomial bridge.
Though mean edge travel times are independent under the prior, correlated observations induce
dependencies in posterior distributions.

Conjugacy properties again facilitate efficient updating of posterior parameters.
Let $\phi, z_t \in \Re^N$ be defined by
$$\phi_e = \ln(\theta_e)
\qquad \text{and} \qquad
z_{t,e} = \left\{\begin{array}{ll}
\ln(y_{t,e}) \qquad & \text{if } e \in x_t \\
0 \qquad & \text{otherwise.}
\end{array}\right.
$$
Note that it is with some abuse of notation that we index vectors and matrices using edge indices.  Define a $|x_t| \times |x_t|$ covariance matrix $\tilde{\Sigma}$ with
elements
$$\tilde{\Sigma}_{e,e'} = \left\{\begin{array}{ll}
\tilde{\sigma}^2 \qquad & \text{for } e=e' \\
2 \tilde{\sigma}^2/3 \qquad & \text{for } e \neq e', \ell(e) = \ell(e') \\
\tilde{\sigma}^2/3 \qquad & \text{otherwise,}
\end{array}\right.$$
for $e,e' \in x_t$, and a $N \times N$ concentration matrix
$$\tilde{C}_{e,e'} = \left\{\begin{array}{ll}
\tilde{\Sigma}^{-1}_{e,e'} \qquad & \text{if } e, e' \in x_t\\
0 \qquad & \text{otherwise,}
\end{array}\right.$$
for $e,e' \in E$.  Then, the posterior distribution of $\phi$ is Gaussian with a mean vector $\mu$ and covariance matrix $\Sigma$ that can be updated according to
\begin{equation}
\label{eq:lognormal-cf-update}
(\mu, \Sigma) \leftarrow \left( \left(\Sigma^{-1} + \tilde{C} \right)^{-1} \left(\Sigma^{-1} \mu + \tilde{C} z_t\right),  \left(\Sigma^{-1} + \tilde{C}\right)^{-1}\right).
\end{equation}
\end{example}
TS (Algorithm \ref{alg:GeneralTS}) can again be applied in a computationally efficient manner.
Each $t$th iteration begins with posterior parameters $\mu \in \Re^N$ and $\Sigma\in \Re^{N\times N}$.  The sample $\hat{\theta}$
can be drawn by first sampling  a vector $\hat{\phi}$ from a Gaussian distribution with mean $\mu$ and covariance matrix $\Sigma$,
and then setting $\hat{\theta}_e = \hat{\phi}_e$ for each $e \in E$.  An action $x$ is selected to maximize
$\E_{q_{\hat{\theta}}}[r(y_t) | x_t = x] = -\sum_{e \in x_t} \hat{\theta}_e$, using Djikstra's algorithm or an alternative.
After applying the selected action, an outcome $y_t$ is observed, and belief distribution
parameters $(\mu, \Sigma)$ are updated according to (\ref{eq:lognormal-cf-update}).

\begin{figure}[h!]
\centering
    \begin{subfigure}{.49\textwidth}
        \centering
	\includegraphics[width=\linewidth]{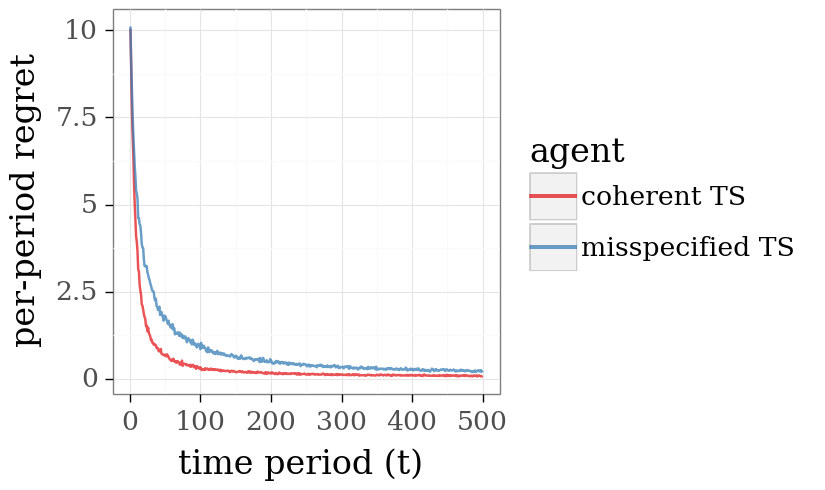}
	\caption{regret}
	  \label{fig:shortest_path_regret_factor}
    \end{subfigure}
    \begin{subfigure}{.49\textwidth}
        \centering
	\includegraphics[width=\linewidth]{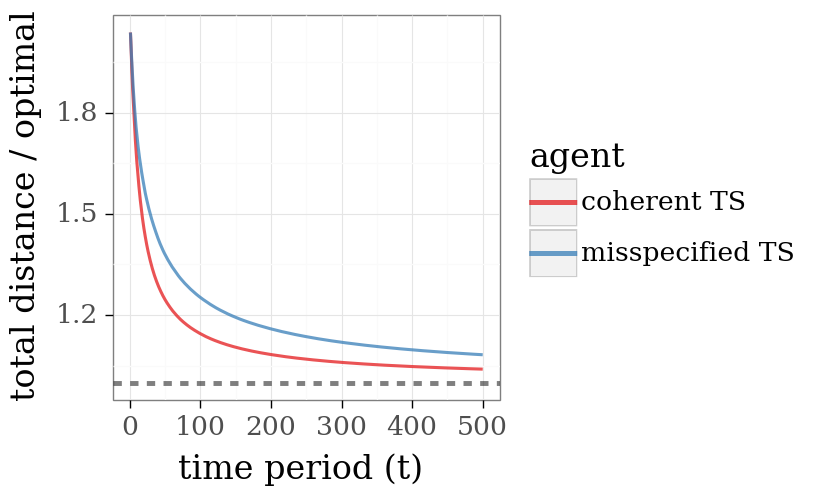}
	\caption{cumulative travel time vs. optimal}
	\label{fig:shortest_path_ratio_factor}
    \end{subfigure}
\caption{Performance of two versions of Thompson sampling in the shortest path problem with correlated travel times.}
\label{fig:shortest_path_factor}
\end{figure}

Figure \ref{fig:shortest_path_factor} plots results from applying TS to Example \ref{ex:log-shortest-path-cf},
again with the binomial bridge, $\mu_e = -\frac{1}{2}$, $\sigma_e^2 = 1$, and $\tilde{\sigma}^2 = 1$.
Each data point represents an average over ten thousand independent simulations.
Despite model differences, an agent can pretend that observations made in this new context are
generated by the model described in Example \ref{ex:log-shortest-path}.
In particular, the agent could maintain an independent log-Gaussian posterior for each $\theta_e$,
updating parameters $(\mu_e,\sigma_e^2)$ as though each $y_{t,e}|\theta$ is independently drawn from a log-Gaussian
distribution.  As a baseline for comparison,
Figure \ref{fig:shortest_path_factor} additionally plots results from application of this approach,
which we will refer to here as {\it misspecified TS}.
The comparison demonstrates substantial improvement that results from accounting
for interdependencies among edge travel times, as is done by what we refer to here as {\it coherent TS}.
Note that we have assumed here that the agent must select a path before initiating
each trip.  In particular, while the agent may be able to reduce travel times in contexts with correlated delays by adjusting the path during
the trip based on delays experienced so far, our model does not allow this behavior.

\chapter{Approximations}\label{se:approximations}

Conjugacy properties in the Bernoulli bandit and shortest path examples that we have considered so far
facilitated simple and computationally efficient Bayesian inference.  Indeed,
computational efficiency can be an important consideration when formulating a model.  However,
many practical contexts call for more complex models for which
exact Bayesian inference is computationally intractable.  Fortunately, there are reasonably efficient and
accurate methods that can be used to approximately sample from posterior distributions.

\begin{figure}[htpb]
\centering
\includegraphics[scale=0.4]{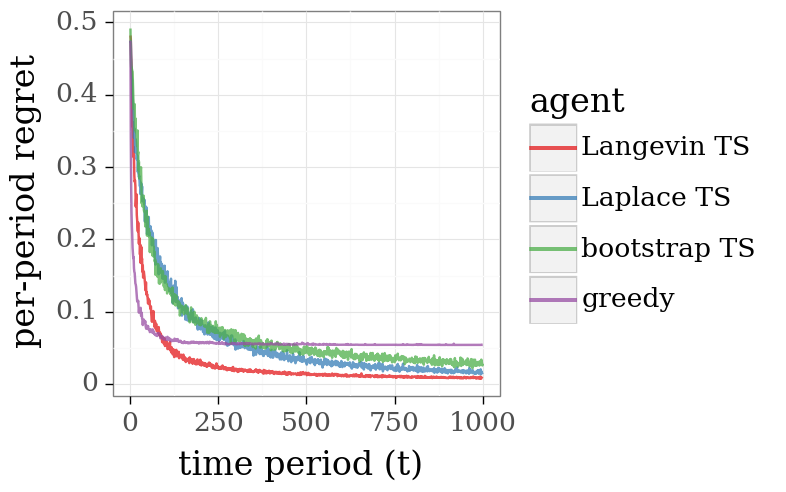}
\caption{Regret experienced by approximation methods applied to the path recommendation problem with binary feedback.}
\label{fig:path-recommendation-approximations}
\end{figure}

In this section we discuss four approaches to approximate posterior sampling: Gibbs sampling, Langevin Monte Carlo,
sampling from a Laplace approximation, and the bootstrap.  Such methods are called for when dealing with
problems that are not amenable to efficient Bayesian inference.
As an example, we consider a variation of the online shortest path problem.
\begin{example}(Binary Feedback)\label{ex:path-recommendation}
Consider Example \ref{ex:log-shortest-path-cf}, except with deterministic travel times and noisy binary observations.
Let the graph represent a binomial bridge with $M$ stages.
Let each $\theta_e$ be independent and gamma-distributed with $\E[\theta_e] = 1$, $\E[\theta_e^2] = 1.5$, and
observations be generated according to
$$y_t | \theta \sim \left\{ \begin{array}{ll}
1 \qquad & \text{with probability } \frac{1}{1 + \exp\left(\sum_{e \in x_t} \theta_e - M\right)} \\
0 \qquad & \text{otherwise.}
\end{array}\right.
$$
We take the reward to be the rating $r_t = y_t$.
This information structure could be used to model, for example, an Internet route recommendation service.
Each day, the system recommends a route $x_t$ and receives feedback $y_t$ from the driver, expressing
whether the route was desirable.
When the realized travel time $\sum_{e \in x_t} \theta_e$ falls short of the prior expectation $M$, the feedback tends to be positive, and
vice versa.
\end{example}
This new model does not enjoy conjugacy properties leveraged in Section \ref{se:GeneralTS}
and is not amenable to efficient exact Bayesian inference.  However, the problem may be addressed
via approximation methods.  To illustrate, Figure \ref{fig:path-recommendation-approximations} plots results from application
of three approximate versions of TS to an online shortest path problem on a twenty-stage binomial bridge with binary feedback.
The algorithms leverage Langevin Monte Carlo, the Laplace approximation, and the bootstrap, three approaches we will discuss, and the
results demonstrate effective learning, in the sense that regret vanishes over time.  Also plotted as a baseline for comparison
are results from application of the greedy algorithm.

In the remainder of this section, we will describe several approaches to approximate TS.  It is worth mentioning
that we do not cover an exhaustive list, and further, our descriptions do not serve as comprehensive or definitive treatments
of each approach.  Rather, our intent is to offer simple descriptions that convey key ideas that may be extended or combined
to serve needs arising in any specific application.

Throughout this section, let $f_{t-1}$ denote the posterior density of $\theta$ conditioned on the history $\hist_{t-1} = ((x_1, y_1),\ldots,(x_{t-1},y_{t-1}))$ of observations. TS generates an action $x_t$ by sampling a parameter vector $\hat{\theta}$ from $f_{t-1}$ and solving for the optimal path under $\hat{\theta}$. The methods we describe generate a sample $\hat{\theta}$ whose distribution approximates the posterior $\hat{f}_{t-1}$, which enables approximate implementations of
TS when exact posterior sampling is infeasible.

\section{Gibbs Sampling}

Gibbs sampling is a general Markov chain Monte Carlo (MCMC) algorithm for drawing approximate samples from multivariate probability distributions. It produces a sequence of sampled parameters $(\hat{\theta}^{n} : n =0,1,2, \ldots)$ forming a Markov chain with stationary distribution $f_{t-1}$. Under reasonable technical conditions, the limiting distribution of this Markov chain is its stationary distribution, and the distribution of $\hat{\theta}^n$ converges to $f_{t-1}$.

Gibbs sampling starts with an initial guess $\hat{\theta}^{0}$.
Iterating over sweeps $n=1,\ldots,N$, for each $n$th sweep, the algorithm iterates over the components $k=1,\ldots,K$,
for each $k$ generating a one-dimensional marginal distribution
$$f^{n,k}_{t-1}(\theta_k) \propto f_{t-1}((\hat{\theta}^n_1, \ldots, \hat{\theta}^n_{k-1}, \theta_k, \hat{\theta}^{n-1}_{k+1}, \ldots, \hat{\theta}^{n-1}_K)),$$
and sampling the $k$th component according to $\hat{\theta}^n_k \sim f^{n,k}_{t-1}$. After $N$ of sweeps, the prevailing vector $\hat{\theta}^{N}$ is taken to be the approximate posterior sample. We refer to \citep{casella1992explaining} for a more thorough introduction to the algorithm.

Gibbs sampling applies to a broad range of problems, and is often computationally viable even when sampling from
$f_{t-1}$ is not.  This is because sampling from a one-dimensional distribution is simpler.  That said, for complex problems,
Gibbs sampling can still be computationally demanding.  This is the case, for example, with our path recommendation problem with binary feedback.
In this context, it is easy to implement a version of Gibbs sampling that generates a close approximation to a posterior sample within
well under a minute.  However, running thousands of simulations
each over hundreds of time periods can be quite time-consuming.  As such, we turn to more efficient approximation methods.

\section{Laplace Approximation}
\label{se:laplace}

We now discuss an approach that approximates a potentially complicated posterior distribution by a Gaussian distribution. Samples from this simpler Gaussian distribution can then serve as approximate samples from the posterior distribution of interest.  Chapelle and Li \citep{chapelle2011empirical} proposed this method to approximate TS in a display advertising problem with a logistic regression model of ad-click-through rates.

Let $g$ denote a probability density function over $\Re^K$ from which we wish to sample. If $g$ is unimodal, and its log density $\ln(g(\phi))$ is strictly concave around its mode $\overline{\phi}$, then $g(\phi)= e^{\ln(g(\phi))}$ is sharply peaked around $\overline{\phi}$. It is therefore natural to consider approximating $g$ locally around its mode. A second-order Taylor approximation to the log-density gives
$$\ln(g(\phi)) \approx \ln(g(\overline{\phi})) - \frac{1}{2} (\phi - \overline{\phi})^\top C (\phi - \overline{\phi}),$$
where
$$C = -\nabla^2 \ln(g(\overline{\phi})).$$
As an approximation to the density $g$, we can then use
$$\tilde{g}(\phi) \propto e^{-\frac{1}{2} (\phi - \overline{\phi})^\top C (\phi - \overline{\phi})}.$$ This is proportional to the density of a Gaussian distribution with mean $\overline{\phi}$ and covariance $C^{-1}$, and hence
$$\tilde{g}(\phi) = \sqrt{|C/2\pi|} e^{-\frac{1}{2} (\phi - \overline{\phi})^\top C (\phi - \overline{\phi})}.$$
We refer to this as the Laplace approximation of $g$.  Since there are efficient algorithms for generating
Gaussian-distributed samples, this offers a viable means to approximately sampling from $g$.

As an example, let us consider application of the Laplace approximation to Example \ref{ex:path-recommendation}.
Bayes rule implies that the posterior density $f_{t-1}$ of $\theta$ satisfies
\[
\Scale[0.95]{
f_{t-1}(\theta) \propto f_0(\theta) \prod_{\tau=1}^{t-1} \left(\frac{1}{1 + \exp\left(\sum_{e \in x_\tau} \theta_e - M\right)}\right)^{y_\tau}  \left(\frac{\exp\left(\sum_{e \in x_\tau} \theta_e - M\right)}{1 + \exp\left(\sum_{e \in x_\tau} \theta_e - M\right)}\right)^{1-y_\tau}.}
\]
The mode $\overline{\theta}$ can be efficiently computed via maximizing $f_{t-1}$, which is log-concave.  An approximate posterior sample $\hat{\theta}$ is then drawn from
a Gaussian distribution with mean $\overline{\theta}$ and covariance matrix $(- \nabla^2 \ln(f_{t-1}(\overline{\theta})))^{-1}$.

Laplace approximations are well suited for Example \ref{ex:path-recommendation} because the log-posterior density is strictly concave and its gradient and Hessian can be computed efficiently. Indeed, more broadly, Laplace approximations tend to be effective for posterior distributions with smooth densities that are sharply peaked around their mode. They tend to be computationally efficient when one can efficiently compute the posterior mode, and can efficiently form the Hessian of the log-posterior density.

The behavior of the Laplace approximation is not invariant to a substitution of variables, and it can sometimes be helpful to apply such a substitution. To illustrate this point, let us revisit the online shortest path problem of Example \ref{ex:log-shortest-path-cf}.  For this problem,
posterior distributions components of $\theta$ are log-Gaussian.  However, the distribution of $\phi$, where $\phi_e = \ln(\theta_e)$ for each edge $e \in E$, is Gaussian.  As such, if the Laplace approximation approach is applied to generate a sample $\hat{\phi}$ from the posterior distribution of $\phi$,
the Gaussian approximation is no longer an approximation, and,
letting $\hat{\theta}_e = \exp(\hat{\phi}_e)$ for each $e \in E$, we obtain a sample $\hat{\theta}$ exactly from the posterior distribution of $\theta$. In this case, through a variable substitution, we can sample in a manner that makes the Laplace approximation exact.  More broadly, for any given problem, it may be possible to introduce variable substitutions that enhance the efficacy of the Laplace approximation.

To produce the computational results reported in Figure \ref{fig:path-recommendation-approximations},
we applied Newton's method with a backtracking line search to maximize $\ln(f_{t-1})$.  Though regret decays and should eventually vanish,
it is easy to see from the figure that, for our example, the performance of the Laplace approximation falls short of Langevin Monte Carlo, which we will discuss in the next section.
This is likely due to the fact that the posterior distribution is not sufficiently close to Gaussian.  It is interesting that, despite serving as a popular
approach in practical applications of TS \citep{chapelle2011empirical,Gomez-Uribe-2016}, the Laplace approximation
can leave substantial value on the table.

\section{Langevin Monte Carlo}
\label{sec::langevin}
We now describe an alternative Markov chain Monte Carlo method that uses gradient information about the target distribution.  Let $g(\phi)$ denote a log-concave probability density function over $\mathbb{R}^{K}$ from which we wish to sample. Suppose that $\ln(g(\phi))$ is differentiable and its gradients are efficiently computable. Arising first in physics, Langevin dynamics refer to the diffusion process
\begin{equation}\label{eq: langevin sde}
d\phi_t = \nabla \ln(g(\phi_t)) dt + \sqrt{2} dB_t
\end{equation}
where $B_t$ is a standard Brownian motion process. This process has $g$ as its unique stationary distribution, and under reasonable technical conditions, the distribution of $\phi_t$ converges rapidly to this stationary distribution \citep{roberts1996exponential,mattingly2002ergodicity}. Therefore simulating the process \eqref{eq: langevin sde} provides a means of approximately sampling from $g$.

Typically, one instead implements a Euler discretization of this stochastic differential equation
\begin{equation}\label{eq: langevin mc}
\phi_{n+1}= \phi_{n} + \epsilon \nabla \ln(g(\phi_n)) + \sqrt{2\epsilon} W_n \qquad n \in \mathbb{N},
\end{equation}
where $W_1, W_2,\ldots$ are i.i.d. standard Gaussian random variables and $\epsilon>0$ is a small step size. Like a gradient ascent method, under this method $\phi_n$ tends to drift in directions of increasing density $g(\phi_n)$. However, random Gaussian noise $W_n$ is injected at each step so that, for large $n$, the position of $\phi_n$ is random and captures the uncertainty in the distribution $g$. A number of papers establish rigorous guarantees for the rate at which this Markov chain converges to its stationary distribution \citep{roberts1998optimal, bubeck2015sampling, durmus2016sampling, cheng2017convergence}. These papers typically require $\epsilon$ is sufficiently small, or that a decaying sequence of step sizes $(\epsilon_1, \epsilon_2,\ldots)$ is used.

We make two standard modifications to this method to improve computational efficiency. First, following recent work \citep{welling2011bayesian}, we implement \emph{stochastic gradient} Langevin Monte Carlo, which uses sampled minibatches of data to compute  approximate rather than exact gradients. Our implementation uses a mini-batch size of 100; this choice seems to be effective but has not been carefully optimized. When fewer than 100 observations are available, we follow the Markov chain \eqref{eq: langevin mc} with exact gradient computation. When more than 100 observations have been gathered, we follow \eqref{eq: langevin mc} but use an estimated gradient $\nabla \ln(\hat{g}_n(\phi_n))$ at each step based on a random subsample of 100 data points. Some work provides rigorous guarantees for stochastic gradient Langevin Monte Carlo by arguing the cumulative impact of the noise in gradient estimation is second order relative to the additive Gaussian noise \citep{teh2016consistency}.

Our second modification involves the use of a preconditioning matrix to improve the mixing rate of the Markov chain \eqref{eq: langevin mc}. For the path recommendation problem in Example \ref{ex:path-recommendation}, we have found that the log posterior density becomes ill-conditioned in later time periods. For this reason, gradient ascent converges very slowly to the posterior mode. Effective optimization methods should leverage second order information. Similarly, due to poor conditioning, we may need to choose an extremely small step size $\epsilon$, causing the Markov chain in \ref{eq: langevin mc} to mix slowly. We have found that preconditioning substantially improves performance. Langevin MCMC can be implemented with a symmetric positive definite preconditioning matrix $A$ by simulating the Markov chain
\[
\phi_{n+1}= \phi_{n} + \epsilon A \nabla \ln(g(\phi_n)) + \sqrt{2\epsilon} A^{1/2} W_n \qquad n \in \mathbb{N},
\]
where $A^{1/2}$ denotes the matrix square root of  $A$. In our implementation, we take $\phi_0 = \argmax_{\phi} \ln(g(\phi))$, so the chain is initialized at the posterior mode, computed via means discussed in Section \ref{se:laplace}, and take the preconditioning matrix $A= -(\nabla^2 \ln(g(\phi)) \rvert_{\phi=\phi_0})^{-1}$ to be the negative inverse Hessian at that point. It may be possible to improve computational efficiency by constructing an incremental approximation to the Hessian, as we will discuss in Subsection \ref{se:incremental}, but we do not explore that improvement here.

\section{Bootstrapping}
\label{se:bootstrap}

As an alternative, we discuss an approach based on the statistical bootstrap, which accommodates even very complex densities.  Use of the bootstrap for TS was first considered in \citep{DBLP:journals/corr/EcklesK14}, though the version studied there applies to Bernoulli bandits and does not naturally generalize to more complex problems.
There are many other versions of the bootstrap approach that can be used to approximately sample from a posterior distribution.
For concreteness, we introduce a specific one that is suitable for examples we cover in this tutorial.

Like the Laplace approximation approach, our bootstrap method assumes that $\theta$ is drawn from a Euclidean space $\Re^K$. Consider first a standard bootstrap method for evaluating the sampling distribution of the maximum likelihood estimate of $\theta$. The method generates a hypothetical history $\hat{\hist}_{t-1} = ((\hat{x}_1,\hat{y}_1), \ldots, (\hat{x}_{t-1}, \hat{y}_{t-1}))$, which is made up of $t-1$ action-observation pairs, each sampled uniformly with replacement from $\hist_{t-1}$. We then maximize the likelihood of $\theta$ under the hypothetical history, which for our shortest path recommendation problem is given by
\[
\Scale[1.0]{
\hat{L}_{t-1}(\theta) = \prod_{\tau=1}^{t-1} \left(\frac{1}{1 + \exp\left(\sum_{e \in \hat{x}_\tau} \theta_e - M\right)}\right)^{\hat{y}_\tau}  \left(\frac{\exp\left(\sum_{e \in \hat{x}_\tau} \theta_e - M\right)}{1 + \exp\left(\sum_{e \in \hat{x}_\tau} \theta_e - M\right)}\right)^{1-\hat{y}_\tau}.}
\]
The randomness in the maximizer of $\hat{L}_{t-1}$ reflects the randomness in the sampling distribution of the maximum likelihood estimate. Unfortunately, this method does not take the agent's prior into account. A more severe issue is that it grossly underestimates the agent's real uncertainty in initial periods. The modification described here is intended to overcome these shortcomings in a simple way.

The method proceeds as follows. First, as before, we draw a hypothetical history $\hat{\hist}_{t-1} = ((\hat{x}_1,\hat{y}_1), \ldots, (\hat{x}_{t-1}, \hat{y}_{t-1}))$, which is made up of $t-1$ action-observation pairs, each sampled uniformly with replacement from $\hist_{t-1}$.
Next, we draw a sample $\theta^0$ from the prior distribution $f_0$. Let $\Sigma$ denote the covariance matrix of the prior $f_0$. Finally, we solve the maximization problem
\[
\hat{\theta} = \argmax_{\theta \in \mathbb{R}^k}  \,\, e^{-(\theta-\theta^0)^\top \Sigma (\theta - \theta^0)}  \hat{L}_{t-1}(\theta)
\]
and treat $\hat{\theta}$ as an approximate posterior sample. This can be viewed as maximizing a randomized approximation $\hat{f}_{t-1}$ to the posterior density, where $\hat{f}_{t-1}(\theta) \propto e^{-(\theta-\theta^0)^\top \Sigma (\theta - \theta^0)}  \hat{L}_{t-1}(\theta)$ is what the posterior density would be if the prior were Gaussian with mean $\theta^0$ and covariance matrix $\Sigma$, and the history of observations were $\hat{\hist}_{t-1}$. When very little data has been gathered, the randomness in the samples mostly stems from the randomness in the prior sample $\theta_0$. This random prior sample encourages the agent to explore in early periods. When $t$ is large, so a lot of a data has been gathered, the likelihood typically overwhelms the prior sample and randomness in the samples mostly stems from the random selection of the history $\hat{\hist}_{t-1}$.

In the context of the shortest path recommendation problem, $\hat{f}_{t-1}(\theta)$ is log-concave and can therefore be efficiently maximized.  Again, to produce our computational results reported in Figure \ref{fig:path-recommendation-approximations}, we applied Newton's method with a backtracking line search to maximize $\ln(\hat{f}_{t-1})$.  Even when it is not possible to efficiently maximize $\hat{f}_{t-1}$, however, the bootstrap approach can be applied with heuristic optimization methods that  identify local or approximate maxima.

As can be seen from Figure \ref{fig:path-recommendation-approximations}, for our example, bootstrapping performs about as well as the Laplace approximation.  One
advantage of the bootstrap is that it is nonparametric, and may work reasonably regardless of the functional form of the posterior distribution, whereas the
Laplace approximation relies on a Gaussian approximation and Langevin Monte Carlo relies on log-concavity and other regularity assumptions.
That said, it is worth mentioning that there is a lack of theoretical justification for bootstrap approaches or even understanding of whether there
are nontrivial problem classes for which they are guaranteed to perform well.

\section{Sanity Checks}

Figure \ref{fig:path-recommendation-approximations} demonstrates that Laplace approximation, Langevin Monte Carlo, and bootstrap approaches, when applied
to the path recommendation problem, learn from binary feedback to improve performance over time.  This may leave one wondering, however,
whether exact TS would offer substantially better performance.  Since we do not have a tractable means of carrying out
exact TS for this problem, in this section, we apply our approximation methods to problems for
which exact TS is tractable.  This enables comparisons between performance of exact and approximate methods.

\begin{figure}[h!]
\centering
    \begin{subfigure}{.49\textwidth}
        \centering
	\includegraphics[width=\linewidth]{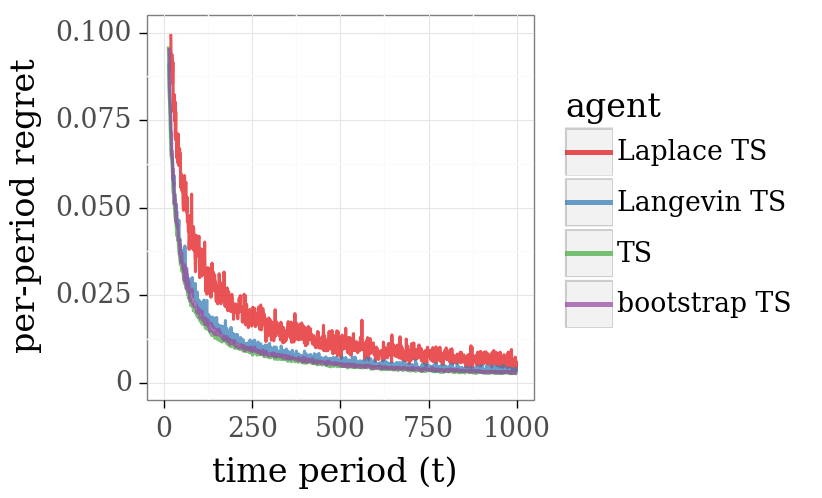}
	\caption{Bernoulli bandit}
	  \label{fig:bernoulli-approximations}
    \end{subfigure}
    \begin{subfigure}{.46\textwidth}
        \centering
	\includegraphics[width=\linewidth]{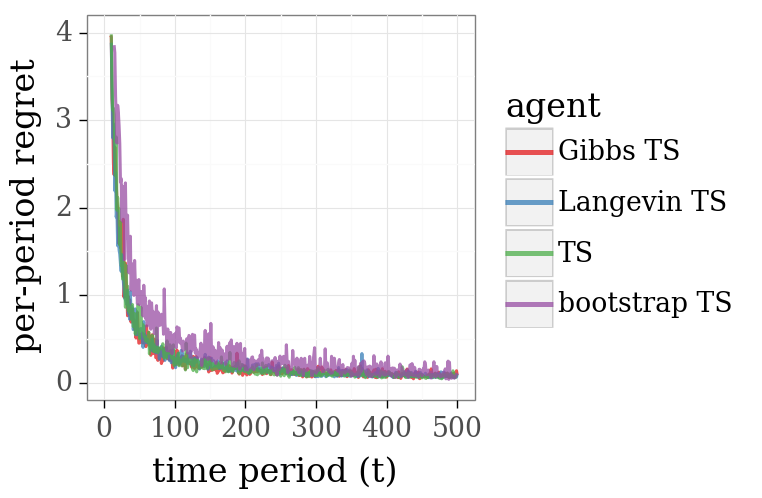}
	\caption{online shortest path}
	\label{fig:sp-approximations}
    \end{subfigure}
\caption{Regret of approximation methods versus exact Thompson sampling.}
\label{fig:sanity-checks}
\end{figure}

Recall the three-armed beta-Bernoulli bandit problem for which results from application of greedy and TS
algorithms were reported in Figure \ref{fig:bernoulli-regret}(b).
For this problem, components of $\theta$ are independent under posterior distributions,
and as such, Gibbs sampling yields exact posterior samples.  Hence, the performance of an approximate version that
uses Gibbs sampling would be identical to that of exact TS.
Figure \ref{fig:bernoulli-approximations} plots results from applying Laplace approximation, Langevin Monte Carlo, and bootstrap approaches.
For this problem, our approximation methods offer performance that is qualitatively similar to exact TS, though
the Laplace approximation performs marginally worse than alternatives in this setting.

Next, consider the online shortest path problem with correlated edge delays.  Regret experienced by TS applied to such a problem
were reported in Figure \ref{fig:shortest_path_regret_factor}.  As discussed in Section \ref{se:laplace}, applying the Laplace approximation approach
with an appropriate variable substitution leads to the same results as exact TS.  Figure \ref{fig:sp-approximations} compares
those results to what is generated by Gibbs sampling, Langevin Monte Carlo, and bootstrap approaches.
Again, the approximation methods yield competitive results, although bootstrapping is marginally less effective than others.

It is easy to verify that for the online shortest path problem and specific choices of step size $\epsilon = 1/2$ and conditioning matrix $A = \Sigma_t$, a
single Langevin Monte Carlo iteration offers an exact posterior sample.  However, our simulations
do not use this step size and carry out multiple iterations.
The point here is not to optimize results for our specific problem but rather to offer a sanity check for the approach.

\section{Incremental Implementation}
\label{se:incremental}

For each of the three approximation methods we have discussed, the computation time required per time period grows as time progresses.
This is because each past observation must be accessed to generate the next action.  This differs from
exact TS algorithms we discussed earlier, which maintain parameters that encode a posterior distribution,
and update these parameters over each time period based only on the most recent observation.

In order to keep the computational burden manageable, it can be important to consider incremental variants of our approximation methods.
We refer to an algorithm as {\it incremental} if it operates with fixed rather than growing per-period compute time.
There are many ways to design incremental variants of approximate posterior sampling algorithms we have presented.
As concrete examples, we consider here particular incremental versions of Laplace approximation and bootstrap approaches.

For each time $t$, let $\ell_t(\theta)$ denote the likelihood of $y_t$ conditioned on $x_t$ and $\theta$.  Hence, conditioned on
$\hist_{t-1}$, the posterior density satisfies
$$f_{t-1}(\theta) \propto f_0(\theta) \prod_{\tau=1}^{t-1} \ell_\tau(\theta).$$
Let $g_0(\theta) = \ln(f_0(\theta))$ and $g_t(\theta) = \ln(\ell_t(\theta))$ for $t > 0$.  To identify the mode of $f_{t-1}$, it suffices to maximize
$\sum_{\tau=0}^{t-1} g_\tau(\theta)$.

Consider an incremental version of the Laplace approximation.  The algorithm maintains statistics $H_t$, and $\overline{\theta}_t$, initialized with
$\overline{\theta}_0 = \argmax_{\theta} g_0(\theta)$, and $H_0 = \nabla^2 g_0(\overline{\theta}_0)$,
and updating according to
$$H_t = H_{t-1} + \nabla^2 g_t(\overline{\theta}_{t-1}),$$
$$\overline{\theta}_t = \overline{\theta}_{t-1} - H_t^{-1} \nabla g_t(\overline{\theta}_{t-1}).$$
This algorithm is a type of online newton method for computing the posterior mode $\overline{\theta}_{t-1}$ that maximizes $\sum_{\tau=0}^{t-1} g_{\tau}(\theta)$. Note that if each function $g_{t-1}$ is strictly concave and quadratic, as would be the case if the prior is Gaussian and observations are linear in $\theta$ and perturbed only by Gaussian noise, each pair $\overline{\theta}_{t-1}$ and $H_{t-1}^{-1}$
represents the mean and covariance matrix of $f_{t-1}$. More broadly, these iterates can be viewed as the mean and covariance matrix of a Gaussian approximation to the posterior, and used to generate an approximate posterior sample $\hat{\theta} \sim N(\overline{\theta}_{t-1}, H_{t-1}^{-1})$. It is worth noting that for linear and generalized linear models, the matrix $\nabla^2 g_t(\overline{\theta}_{t-1})$ has rank one, and therefore $H_{t}^{-1}=(H_{t-1} + \nabla^2 g_t(\overline{\theta}_{t-1}))^{-1}$ can be updated incrementally using the Sherman-Woodbury-Morrison formula. This incremental version of the Laplace approximation is closely related to the notion of an extended Kalman filter, which has been explored in greater depth by G\'{o}mez-Uribe \citep{Gomez-Uribe-2016} as a means for incremental approximate TS with exponential families of distributions.

Another approach involves incrementally updating each of an ensemble of models to behave like a sample from the posterior distribution.
The posterior can be interpreted as a distribution of ``statistically plausible'' models, by which we mean models that are sufficiently consistent with prior beliefs and
the history of observations.  With this interpretation in mind, TS can be thought of as randomly drawing from the range of
statistically plausible models.  {\it Ensemble sampling} aims to maintain, incrementally update, and sample from a finite set of such models.
In the spirit of particle filtering, this set of models approximates the posterior distribution.  The workings of ensemble sampling are in some ways more intricate
than conventional uses of particle filtering, however, because interactions between the ensemble of models and selected actions can skew the distribution.
{\it Ensemble sampling} is presented in more depth in \citep{LuVR2017}, which draws inspiration from work on exploration in deep reinforcement
learning \citep{osband2016deep}.

There are multiple ways of generating suitable model ensembles.  One builds on the aforementioned bootstrap method and involves
fitting each model to a different bootstrap sample.  To elaborate, consider maintaining $N$ models with parameters
$(\overline{\theta}_t^n, H_t^n : n =1,\ldots,N)$.  Each set is initialized with
$\overline{\theta}^n_0 \sim g_0$, $H^n_0 = \nabla g_0(\overline{\theta}^n_0)$, $d^n_0 = 0$,
and updated according to
$$H_t^n = H_{t-1}^n + z_t^n \nabla^2 g_t(\overline{\theta}^n_{t-1}),$$
$$\overline{\theta}_t^n = \overline{\theta}_{t-1}^n -z_t^n  (H_t^n)^{-1} \nabla g_t(\overline{\theta}_{t-1}^n),$$
where each $z_t^n$ is an independent Poisson-distributed sample with mean one.
Each $\overline{\theta}_t^n$ can be viewed as a random statistically plausible model, with randomness stemming from the initialization of $\overline{\theta}_0^n$ and the random weight $z_t^n$ placed on each observation.
The variable, $z_\tau^n$ can loosely be interpreted as a number of replicas of the data sample $(x_\tau, y_\tau)$
placed in a hypothetical history $\hat{\hist}_t^n$. Indeed, in a data set of size $t$,  the number of replicas of a particular bootstrap data sample follows a ${\rm Binomial}(t, 1/t)$ distribution, which is approximately ${\rm Poisson}(1)$ when $t$ is large. With this view, each $\overline{\theta}^n_t$ is effectively fit to a different data set $\hat{\hist}_t^n$, distinguished by the random number of replicas assigned to each data sample.  To generate an action $x_t$, $n$ is sampled uniformly from $\{1,\ldots,N\}$, and the action is chosen to maximize $\E[r_t | \theta = \overline{\theta}_{t-1}^n]$.
Here, $\overline{\theta}_{t-1}^n$ serves as the approximate posterior sample.
Note that the per-period compute time grows with $N$, which is an algorithm tuning parameter.

This bootstrap approach offers one mechanism for incrementally updating an ensemble of models.  In Section \ref{se:ensemble}, we will discuss another, which we
apply to active learning with neural networks.

\chapter{Practical Modeling Considerations}

Our narrative over previous sections has centered around a somewhat idealized view of TS,
which ignored the process of prior specification and assumed a simple model in which the system and set of feasible
actions is constant over time and there is no side information on decision context.
In this section, we provide greater perspective on the process of prior specification and on extensions of
TS that serve practical needs arising in some applications.

\section{Prior Distribution Specification}
\label{se:prior-specification}

The algorithms we have presented require as input a prior distribution over model parameters.
The choice of prior can be important, so let us now discuss its role and how it might be selected.
In designing an algorithm for an online decision problem, unless the value of $\theta$ were known with certainty,
it would not make sense to optimize performance for a single value, because that could
lead to poor performance for other plausible values.  Instead, one might design the algorithm
to perform well on average across a collection of possibilities.  The prior can be thought
of as a distribution over plausible values, and its choice directs the algorithm to perform well
on average over random samples from the prior.

For a practical example of prior selection, let us revisit the banner ad placement problem introduced in
Example \ref{ex:bernoulli}.  There are $K$ banner ads for a single product, with unknown
click-through probabilities $(\theta_1,\ldots,\theta_K)$.  Given a prior, TS
can learn to select the most successful ad.  We could use a uniform or, equivalently,
a $\text{beta}(1,1)$ distribution over each $\theta_k$.  However, if some values
of $\theta_k$ are more likely than others, using a uniform prior sacrifices performance.
In particular, this prior represents no understanding of the context, ignoring any useful
knowledge from past experience.  Taking knowledge into account reduces what must be learned
and therefore reduces the time it takes for TS to identify the most effective ads.

\begin{figure}[htpb]
\centering
\includegraphics[scale=0.4]{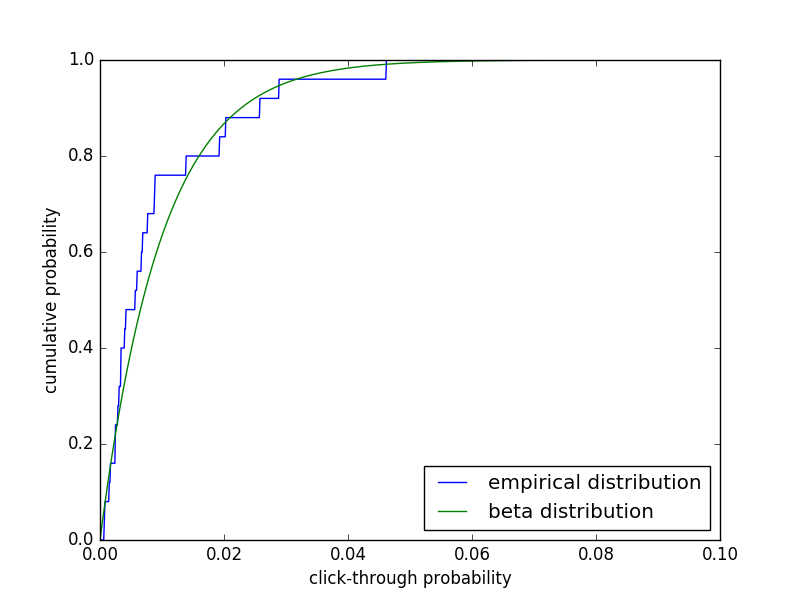}
\caption{An empirical cumulative distribution and an approximating beta distribution.}
\label{fig:beta-data}
\end{figure}

Suppose we have a data set collected from experience with previous products and their ads,
each distinguished by stylistic features such as language, font, and background, together
with accurate estimates of click-through probabilities.  Let us consider an empirical approach to prior selection
that leverages this data.  First, partition past ads into $K$ sets, with each $k$th partition
consisting of those with stylistic features most similar to the $k$th ad under current consideration.
Figure \ref{fig:beta-data} plots a hypothetical
empirical cumulative distribution of click-through probabilities for ads in the $k$th set.
It is then natural to consider as a prior a smoothed approximation of this distribution,
such as the $\text{beta}(1,100)$ distribution also plotted in Figure \ref{fig:beta-data}.
Intuitively, this process assumes that click-through probabilities of past ads in set $k$
represent plausible values of $\theta_k$.  The resulting prior is informative; among other things,
it virtually rules out click-through probabilities greater than $0.05$.

\begin{figure}[htpb]
\centering
    \begin{subfigure}{.49\textwidth}
        \centering
	\includegraphics[width=\linewidth]{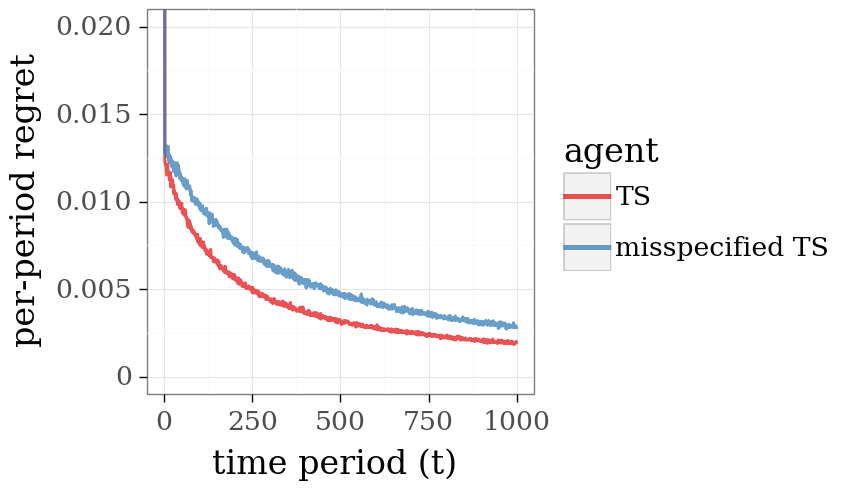}
	\caption{regret}
	 \label{fig:bernoulli-regret-misspecified}
    \end{subfigure}
    \begin{subfigure}{.49\textwidth}
        \centering
	\includegraphics[width=\linewidth]{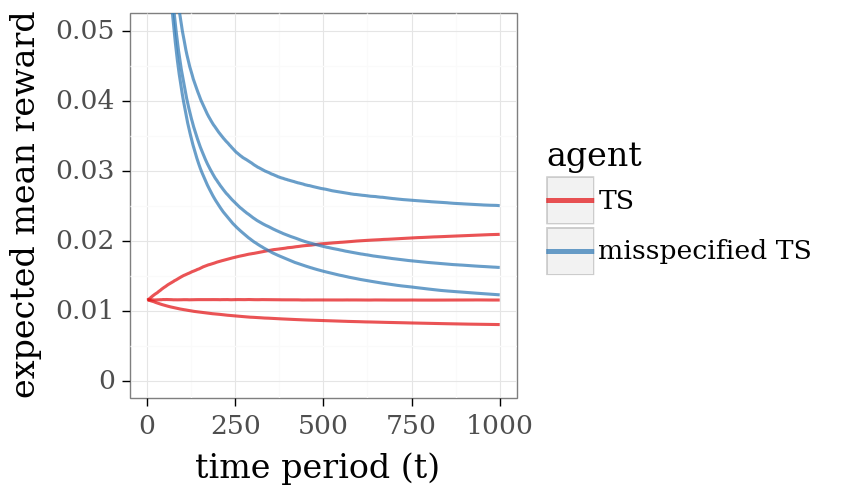}
	\caption{expected mean rewards}
	 \label{fig:bernoulli-means-misspecified}
    \end{subfigure}
\caption{Comparison of TS for the Bernoulli bandit problem with coherent versus misspecified priors.}
\label{fig:bernoulli-misspecified}
\end{figure}

A careful choice of prior can improve learning performance.  Figure \ref{fig:bernoulli-misspecified}
presents results from simulations of a three-armed Bernoulli bandit.  Mean rewards of the three
actions are sampled from $\text{beta}(1,50)$, $\text{beta}(1,100)$, and $\text{beta}(1,200)$ distributions, respectively.
TS is applied with these as prior distributions and with a uniform prior distribution.  We refer to the latter as
a {\it misspecified prior} because it is not consistent
with our understanding of the problem.  A prior that is consistent in this sense is termed {\it coherent}.
Each plot represents an average over ten thousand independent
simulations, each with independently sampled mean rewards.  Figure \ref{fig:bernoulli-regret-misspecified}
plots expected regret, demonstrating that the misspecified prior increases regret.
Figure \ref{fig:bernoulli-regret-misspecified} plots the evolution of the agent's mean reward conditional
expectations.  For each algorithm, there are three curves corresponding to the best, second-best,
and worst actions, and they illustrate how starting with a misspecified prior delays learning.

\section{Constraints, Context, and Caution}
\label{se:constraints}

Though Algorithm \ref{alg:GeneralTS}, as we have presented it, treats a very general model, straightforward extensions accommodate even
broader scope.  One involves imposing {\bf time-varying constraints} on the actions.  In particular, there could
be a sequence of admissible action sets $\mathcal{X}_t$ that constrain actions $x_t$.
To motivate such an extension, consider our shortest path example.  Here, on any given day, the drive to work may be constrained
by announced road closures.
If $\mathcal{X}_t$ does not depend on $\theta$ except through possible dependence on the history of observations,
TS (Algorithm \ref{alg:GeneralTS}) remains an effective approach, with the only required modification being to constrain
the maximization problem in Line 6.

Another extension of practical import addresses {\bf contextual online decision problems}.  In such problems, the response $y_t$
to action $x_t$ also depends on an independent random variable $z_t$ that the agent observes prior to making her decision.
In such a setting, the conditional distribution of $y_t$ takes the form $p_\theta(\cdot | x_t,z_t)$.  To motivate this,
consider again the shortest path example, but with the agent observing a weather report $z_t$ from a news channel
before selecting a path $x_t$.  Weather may affect delays along different edges differently, and the agent can take
this into account before initiating her trip.  Contextual problems of this flavor can be addressed through
augmenting the action space and introducing time-varying constraint sets.  In particular, if we view
$\tilde{x}_t = (x_t,z_t)$ as the action and constrain its choice to $\mathcal{X}_t = \{(x,z_t): x \in \mathcal{X}\}$,
where $\mathcal{X}$ is the set from which $x_t$ must be chosen, then it is straightforward to apply TS
to select actions $\tilde{x}_1,\tilde{x}_2,\ldots$.  For the shortest path problem, this can be interpreted as allowing the agent to dictate
both the weather report and the path to traverse, but constraining the agent to provide a weather report
identical to the one observed through the news channel.

In some applications, it may be important to ensure that expected performance exceeds some prescribed baseline.  This can
be viewed as a level of {\bf caution} against poor performance.  For
example, we might want each action applied to offer expected reward of at least some level $\underline{r}$.  This can
again be accomplished through constraining actions: in each $t$th time period, let the action set be
$\mathcal{X}_t = \{x \in \mathcal{X} : \mathbb{E}[r_t | x_t=x] \geq \underline{r}\}$.  Using such an action set ensures
that expected average reward exceeds $\underline{r}$.  When actions are related, an actions that is initially omitted from the
set can later be included if what is learned through experiments with similar actions increases the agent's expectation
of reward from the initially omitted action.

\section{Nonstationary Systems}
\label{se:nonstationary}

Problems we have considered involve model parameters $\theta$ that are constant
over time.  As TS hones in on an optimal action, the frequency
of exploratory actions converges to zero.
In many practical applications, the agent faces a nonstationary system,
which is more appropriately modeled by time-varying parameters $\theta_1, \theta_2, \ldots$,
such that the response $y_t$ to action $x_t$ is generated according to
$p_{\theta_t}(\cdot|x_t)$.
In such contexts, the agent should never stop exploring, since it needs to track
changes as the system drifts.
With minor modification, TS remains an effective approach
so long as model parameters change little over durations that are sufficient to identify
effective actions.

In principle, TS could be applied to a broad range of problems where the parameters $\theta_1, \theta_2, \theta_3, ...$ evolve according to a stochastic process
by using techniques from filtering and sequential Monte Carlo to generate posterior samples. Instead we describe below some much simpler approaches to such problems.

One simple approach to addressing nonstationarity involves ignoring historical observations
made beyond some number $\tau$ of time periods in the past.  With such an approach,
at each time $t$, the agent would produce a posterior distribution based on the prior
and conditioned only on the most recent $\tau$ actions and observations.
Model parameters are sampled from this distribution, and an action is selected to
optimize the associated model.  The agent never ceases to explore, since the degree to which the posterior
distribution can concentrate is limited by the number of observations taken into account.
Theory supporting such an approach is developed in \citep{besbes2017nipsMAB}. 

An alternative approach involves modeling evolution of a belief distribution in a manner that discounts
the relevance of past observations and tracks a time-varying parameters $\theta_t$.  We now consider such a model
and a suitable modification of TS.  Let us start with the simple context
of a Bernoulli bandit.  Take the prior for each $k$th mean reward to be $\text{beta}(\alpha,\beta)$.
Let the algorithm update parameters to identify the belief distribution
of $\theta_t$ conditioned on the history $\hist_{t-1} = ((x_1,y_1), \ldots, (x_{t-1},y_{t-1}))$
according to
\begin{equation}
\label{eq:nonstationary-bernoulli-bandit}
(\alpha_k, \beta_k) \leftarrow \left\{\begin{array}{ll}
\big((1-\gamma) \alpha_k + \gamma \overline{\alpha}, (1-\gamma) \beta_k + \gamma \overline{\beta}\big) \qquad & x_t \neq k \\
\big((1-\gamma) \alpha_k + \gamma \overline{\alpha} + r_t, (1-\gamma) \beta_k + \gamma \overline{\beta} + 1-r_t\big) \qquad & x_t = k,
 \end{array}\right.
 \end{equation}
where $\gamma \in [0,1]$ and $\overline{\alpha}_k, \overline{\beta}_k > 0$.  This models a process for which the
belief distribution converges to $\text{beta}(\overline{\alpha}_k,\overline{\beta}_k)$ in the absence of observations.
Note that, in the absence of observations, if $\gamma > 0$ then $(\alpha_k,\beta_k)$ converges to $(\overline{\alpha}_k, \overline{\beta}_k)$.
Intuitively, the process can be thought of as randomly perturbing model parameters in each time period, injecting uncertainty.
The parameter $\gamma$ controls how quickly uncertainty is injected.  At one extreme, when $\gamma=0$,
no uncertainty is injected.  At the other extreme, $\gamma=1$ and each $\theta_{t,k}$ is an independent
$\text{beta}(\overline{\alpha}_k,\overline{\beta}_k)$-distributed process.
A modified version of Algorithm \ref{alg:BernoulliTS} can be applied to this nonstationary Bernoulli bandit problem,
the differences being in the additional arguments $\gamma$, $\overline{\alpha}$, and $\overline{\beta}$ ,
and the formula used to update distribution parameters.

The more general form of TS presented in Algorithm \ref{alg:GeneralTS} can be modified in an analogous
manner.  For concreteness, let us focus on the case where $\theta$ is restricted to a finite set; it is straightforward to extend
things to infinite sets.  The conditional distribution update in Algorithm \ref{alg:GeneralTS}  can be written as
$$p(u) \leftarrow \frac{p(u) q_u(y_t | x_t)}{\sum_v p(v) q_v(y_t | x_t)}.$$
To model nonstationary model parameters, we can use the following alternative:
$$p(u) \leftarrow \frac{\overline{p}^\gamma(u) p^{1-\gamma}(u) q_u(y_t | x_t)}{\sum_v \overline{p}^\gamma(v) p^{1-\gamma}(v) q_v(y_t | x_t)}.$$
This generalizes the formula provided earlier for the Bernoulli bandit case.
Again, $\gamma$ controls the rate at which uncertainty is injected.
The modified version of Algorithm \ref{alg:BernoulliTS}, which we refer to as {\it nonstationary TS},
takes $\gamma$ and $\overline{p}$ as additional arguments and replaces the distribution update formula.

\begin{figure}[htpb]
\centering
\includegraphics[scale=0.4]{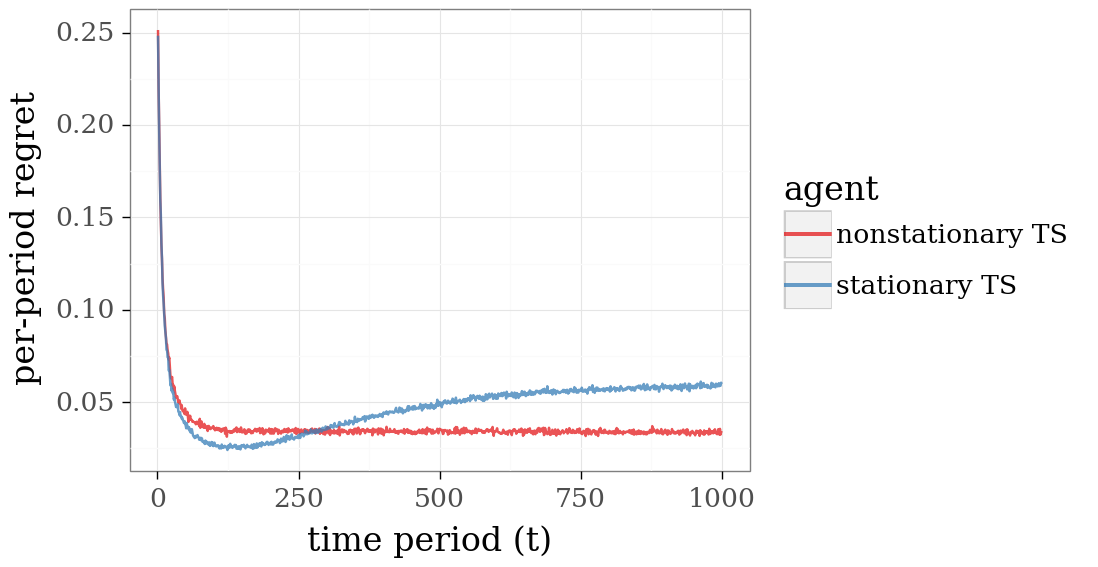}
\caption{Comparison of TS versus nonstationary TS with a nonstationary Bernoulli bandit problem.}
\label{fig:nonstationary-TS}
\end{figure}

Figure \ref{fig:nonstationary-TS} illustrates potential benefits of nonstationary TS when dealing with a nonstationairy
Bernoulli bandit problem.  In these simulations, belief distributions evolve according to Equation (\ref{eq:nonstationary-bernoulli-bandit}).
The prior and stationary distributions are specified by $\alpha = \overline{\alpha} = \beta = \overline{\beta} = 1$.  The
decay rate is $\gamma = 0.01$.  Each plotted point represents an average over 10,000 independent simulations.
Regret here is defined by $\text{regret}_t(\theta_t) = \max_k \theta_{t,k} - \theta_{t,x_t}$.
While nonstationary TS updates its belief distribution in a manner consistent with the
underlying system, TS pretends that the success probabilities are constant over time
and updates its beliefs accordingly.  As the system drifts over time, TS becomes less effective,
while nonstationary TS retains reasonable performance.  Note, however, that due to nonstationarity,
no algorithm can promise regret that vanishes with time.

\section{Concurrence}
\label{se:concurrence}

In many applications, actions are applied concurrently.   As an example, consider a variation of the online shortest path
problem of Example \ref{ex:log-shortest-path}.  In the original version of this problem, over each period, an agent selects and traverses
 a path from origin to destination, and upon completion, updates a posterior distribution based on observed edge traversal times.
Now consider a case in which, over each period, multiple agents travel between the same origin and destination, possibly
along different paths, with the travel time experienced by agents along each edge $e$ to conditionally independent, conditioned on $\theta_e$.
At the end of the period, agents update a common posterior distribution based on their collective experience. The paths represent concurrent
actions, which should be selected in a manner that diversifies experience.

\begin{figure}[h!]
\centering
    \begin{subfigure}{.49\textwidth}
        \centering
	\includegraphics[width=\linewidth]{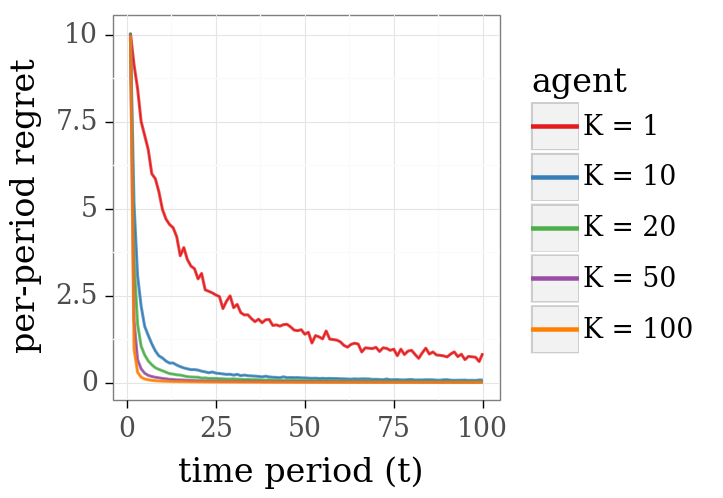}
	\caption{per-action regret over time}
	  \label{fig:cocurrence_a}
    \end{subfigure}
    \begin{subfigure}{.49\textwidth}
        \centering
	\includegraphics[width=\linewidth]{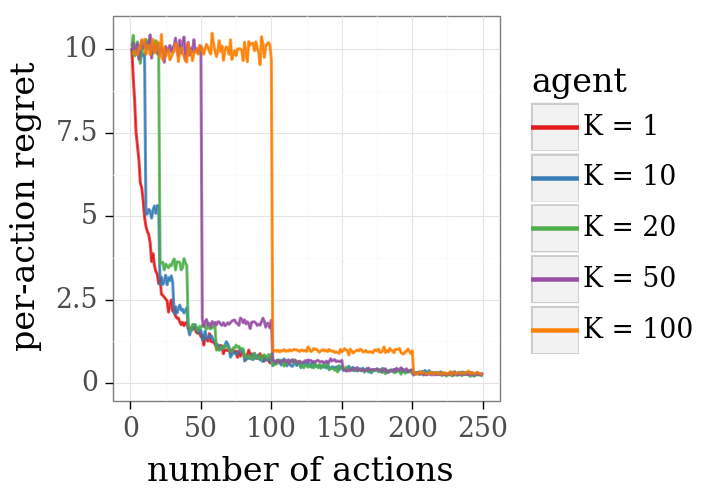}
	\caption{per-action regret over actions}
	\label{fig:concurrence_b}
    \end{subfigure}
\caption{Performance of concurrent Thompson sampling.}
\label{fig:concurrence}
\end{figure}

TS naturally suits this concurrent mode of operation.  Given the posterior distribution available
at the beginning of a time period, multiple independent samples can be drawn to produce paths for multiple agents.
Figure \ref{fig:concurrence} plots results from applying TS in this manner.
Each simulation was carried out with $K$ agents navigating over each time period through a twenty-stage binomial bridge.
Figure \ref{fig:concurrence}(a) demonstrates that the per-action regret experienced by each agent decays more rapidly with time as the number of agents grows.
This is due to the fact that each agent's learning is accelerated by shared observations.
On the other hand, Figure \ref{fig:concurrence}(b) shows that per-action regret decays more slowly as a function of the number of actions taken
so far by the collective of agents.  This loss is due to fact that the the posterior distribution is updated only after $K$ concurrent actions are completed,
so actions are not informed by observations generated by concurrent ones as would be the case if the $K$ actions were applied sequentially.

As discussed in \citep{scott2010modern}, concurrence plays an important role in web services,
where at any time, a system may experiment by providing different versions of a service to different users.  Concurrent TS
offers a natural approach for such contexts.  The version discussed above involves synchronous action selection and
posterior updating.  In some applications, it is more appropriate to operate asynchronously, with actions
selected on demand and the posterior distribution updated as data becomes available.
The efficiency of synchronous and asynchronous variations of concurrent TS is studied in
\citep{kandasamy2018parallel}.  There are also situations where an agent can alter an action based
on recent experience of other agents, within a period before the action is complete.
For example, in the online shortest path problem, an agent may decide to change course to avoid an edge
if new observations made by other agents indicate a long expected travel time.
Producing a version of TS that effectively adapts to such information while still exploring in
a reliably efficient manner requires careful design, as explained in \citep{dimakopoulou2018}.

\chapter{Further Examples}

As contexts for illustrating the workings of TS, we have presented the Bernoulli bandit and variations of the online shortest path problem.
To more broadly illustrate the scope of TS and issues that arise in various kinds of applications, we present several additional examples in this section.

\section{News Article Recommendation}
\label{sec:contextualRecommendation}
Let us start with an online news article recommendation problem in which a website needs to learn to recommend \emph{personalized} and \emph{context-sensitive} news articles to its users, as has been discussed in \citep{LiNewsArticle2010} and \citep{chapelle2011empirical}. The website interacts with a sequence of users, indexed by $t\in \{1,2,\ldots \}$. In each round $t$,  it observes a feature vector $z_{t} \in \mathbb{R}^d$  associated with the $t$th user, chooses a news article $x_t$ to display from among a set of $k$ articles $\Xc=\{1,\ldots, k\}$, and then observes a binary reward $r_{t} \in \{0,1\}$ indicating whether the user liked this article. 

The user's feature vector might, for example, encode the following information:
\begin{itemize}
	\item The visiting user's recent activities, such as the news articles the user has read recently.
	\item The visiting user's demographic information, such as the user's gender and age.
	\item The visiting user's contextual information, such as the user's location and the day of week.
\end{itemize}
Interested readers can refer to Section 5.2.2 of \citep{LiNewsArticle2010} for an example of feature construction in a practical context.

Following section 5 of \citep{chapelle2011empirical}, we model the probability a user with features $z_t$ likes a given article $x_t$ through a logit model. Specifically, each article $x\in \Xc$ is associated with a $d$--dimensional parameter vector $\theta_x \in \mathbb{R}^d$. Conditioned on $x_t$, $\theta_{x_t}$ and $z_{t}$, a positive review occurs with probability $g(z_{t}^T \theta_{x_t})$, where $g$ is the logistic function, given by $g(a) = 1/(1+e^{-a})$. The per-period regret of this problem is defined by 
\[
\mathrm{regret}_t \left( \theta_1, \ldots, \theta_K \right)=\max_{x \in \mathcal{X}} g(z_{t}^T \theta_{x}) - g(z_{t}^T \theta_{x_t}) \quad \forall t=1,2,\ldots
\]
and measures the gap in quality between the recommended article $x_t$ and the best possible recommendation that could be made based on the user's features. This model allows for generalization across users, enabling the website to learn to predict whether a user with given features $z_t$ will like a news article based on experience recommending that article to different users.  

As in the path recommendation problem treated in Section \ref{se:approximations}, this problem is not amenable to efficient exact Bayesian inference. Consequently, we applied two approximate Thompson sampling methods: one samples from a Laplace approximation of the posterior (see Section~\ref{se:laplace}) and the other uses Langevin Monte Carlo to generate an approximate posterior sample (see Section~\ref{sec::langevin}).  To offer a baseline, we also applied the $\epsilon$-greedy algorithm, and searched over values of $\epsilon$ for the best performer.

We present simulation results for a simplified synthetic setting with $K=|\mathcal{X}|=3$ news articles and feature dimension $d=7$. At each time $t\in\{1,2,\cdots\}$, the feature vector $z_{t}$ has constant $1$ as its first component and each of its other components is independently drawn from a Bernoulli distribution with success probability $1/6$. Each components of $z_{t}$ could, for example, indicate presence of a particular feature, like whether the user is a woman or is accessing the site from within the United States, in which the corresponding component of $\theta_{x}$ would reflect whether users with this feature tend to enjoy article $x$ more than other users, while the first component of  $\theta_{x}$ reflects the article's overall popularity.


\begin{figure}[htpb]
\centering
\includegraphics[width=4in]{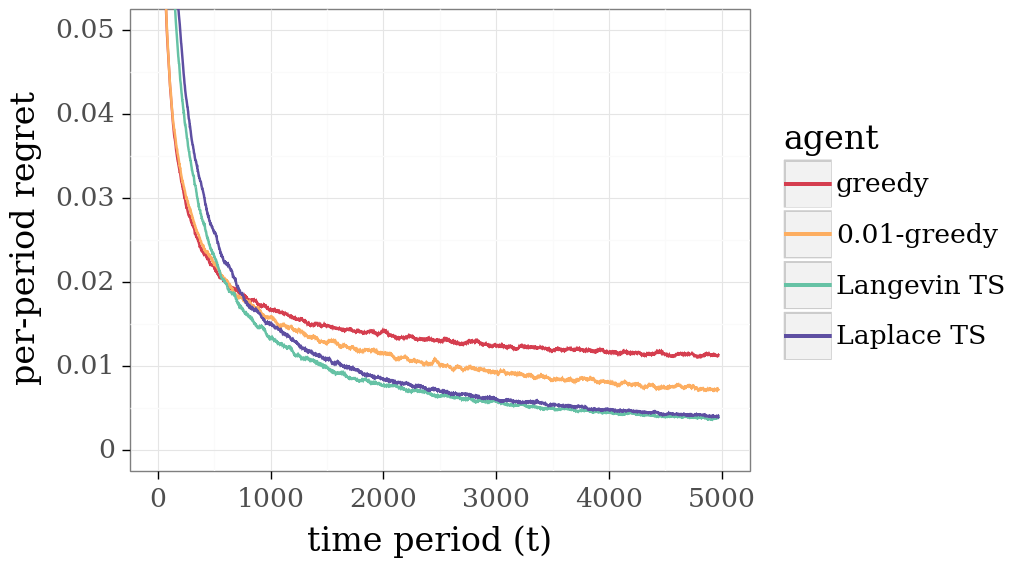}
\caption{Performance of different algorithms applied to the news article recommendation problem.}
\label{fig:news_recommendation}
\end{figure}


Figure~\ref{fig:news_recommendation} presents results from applying Laplace and Langevin Monte Carlo approximations of Thompson sampling as well as greedy and $\epsilon$-greedy algorithms. The plots in Figure \ref{fig:news_recommendation} are generated  by averaging over $2,000$ random problem instances.  In each instance, the $\theta_x$'s were independently sampled from $N(0, I)$, where  $I$ is the $7 \times 7$ identity matrix. Based on our simulations, the $\epsilon$-greedy algorithm incurred lowest regret with $\epsilon = 0.01$.  Even with this optimized value, it is substantially outperformed by Thompson sampling.


We conclude this section by discussing some extensions to the simplified model presented above.  One major limitation is  that the current model does not allow for generalization across news articles. The website needs to estimate $\theta_x$ separately for each article $x \in \mathcal{X}$, and can't leverage data on the appeal of other, related, articles when doing so. Since today's news websites have thousands or even millions of articles, this is a major limitation in practice. Thankfully, alternative models allow for generalization across news articles as well as users. One such model constructs a feature vector $z_{t,x}$ that encodes features of the $t$th user, the article $x$, and possibly interactions between these. Because the feature vector also depends on $x$, it is without loss of generality to restrict to a parameter vector $\theta_{x}=\theta$ that is common across articles. The probability user $t$ likes the article $x_t$ is given by 
$g(z_{t, x_t}^\top \theta)$. Such generalization models enable us to do ``transfer learning,'' i.e. to use information gained by recommending one article to reduce the uncertainty about the weight vector of another article.

Another limitation of the considered model is that the news article set $\mathcal{X}$ is time-invariant. In practice, the set of relevant articles will change over time as fresh articles become available or some existing articles become obsolete. 
%
%
 Even with generalization across news articles, a time-varying news article set,
 or both, the considered online news article recommendation problem is still a contextual bandit problem. As discussed in Section~\ref{se:constraints}, all the algorithms discussed in this subsection are also applicable to those cases, after some proper modifications.

\section{Product Assortment}
Let us start with an assortment planning problem.
Consider an agent who has an ample supply of each of $n$ different products, indexed by $i=1,2,\ldots,n$. The seller collects a profit of $p_i$ per unit sold of product type $i$. In each period, the agent has the option of offering a subset of the products for sale. Products may be substitutes or complements, and therefore the demand for a product  may be influenced by the other products offered for sale in the same period. In order to maximize her profit, the agent needs to carefully select the optimal set of products to offer in each period. We can represent the agent's decision variable in each period as a vector $x\in \{0,1\}^n$ where $x_i=1$ indicates that product $i$ is offered and $x_i = 0$ indicates that it is not. Upon offering an assortment containing product $i$ in some period, the agent observes a random log-Gaussian-distributed demand $d_i$. The mean of this log-Gaussian distribution depends on the entire assortment $x$ and an uncertain matrix $\theta\in \Re^{k\times k}$. In particular
\[
\log(d_i) \mid \theta, x \sim N\left((\theta x)_i , \sigma^2  \right)
\]
where $\sigma^2$ is a known parameter that governs the level of idiosyncratic randomness in realized demand across periods. For any product $i$ contained in the assortment $x$,
\[
(\theta x)_i = \theta_{ii} + \sum_{j\neq i} x_j \theta_{ij} ,
\]
where $\theta_{ii}$ captures the demand rate for item $i$ if it were the sole product offered and each $\theta_{ij}$ captures the effect availability of product $j$ has on demand for product $i$. When an assortment $x$ is offered, the agent earns expected profit
\begin{equation}\label{eq:totalprofit}
\E\left[\sum_{i=1}^n p_i x_i d_i \mid \theta, x \right]=\sum_{i=1}^n p_i x_i e^{(\theta x)_i + \frac{\sigma^2}{2}}.
\end{equation}
If $\theta$ were known, the agent would always select the assortment $x$ that maximizes her expected profit in \eqref{eq:totalprofit}. However, when $\theta$ is unknown, the agent needs to learn to maximize profit by exploring different assortments and observing the realized demands.

TS can be adopted as a computationally efficient solution to this problem. We assume the agent begins with a multivariate Gaussian prior over $\theta$.
Due to conjugacy properties of Gaussian and log-Gaussian distributions, the posterior distribution of $\theta$ remains Gaussian after any number of periods. At the beginning of each $t$'th period, the TS algorithm draws a sample $\hat\theta_t$ from this Gaussian posterior distribution. Then, the agent selects an assortment that would maximize her expected profit in period $t$ if the sampled  $\hat\theta_t$ were indeed the true parameter.

As in Examples \ref{ex:log-shortest-path} and \ref{ex:log-shortest-path-cf}, the mean and covariance matrix of the posterior distribution of $\theta$ can be updated in closed form. However, because $\theta$ is a matrix rather than a vector, the explicit form of the update is more complicated. To describe the update rule, we first introduce $\bar\theta$ as the vectorized version of $\theta$ which is generated by stacking the columns of $\theta$ on top of each other. Let  $x$ be the assortment selected in a period,  $i_1,i_2,\ldots,i_k$ denote the the products included in this assortment (i.e., $\text{supp}(x) = \{i_1,i_2,\ldots,i_k\}$)   and  $z\in\R^{k}$ be defined element-wise as
$$z_j = \ln(d_{i_j}), ~~j= 1,2,\ldots,k.$$
Let $S$ be a $k\times n$ ``selection matrix'' where $S_{j,i_j} = 1$ for $j=1,2,\ldots,k$ and all its other elements are 0. Also, define
$$W = x^\top\otimes S,$$
where $\otimes$ denotes the Kronecker product of matrices. At the end of current period, the posterior mean $\mu$ and covariance matrix $\Sigma$ of $\bar\theta$ are updated according to the following rules:
\begin{eqnarray*}
\mu &\leftarrow& \left(\Sigma^{-1}+ \frac{1}{\sigma^2}W^\top W\right)^{-1}\left(\Sigma^{-1}\mu + \frac{1}{\sigma^2}W^\top z\right), \\
\Sigma &\leftarrow& \left(\Sigma^{-1}+ \frac{1}{\sigma^2}W^\top W\right)^{-1}.
\end{eqnarray*}

\begin{figure}[htpb]
\centering
\includegraphics[scale=0.4]{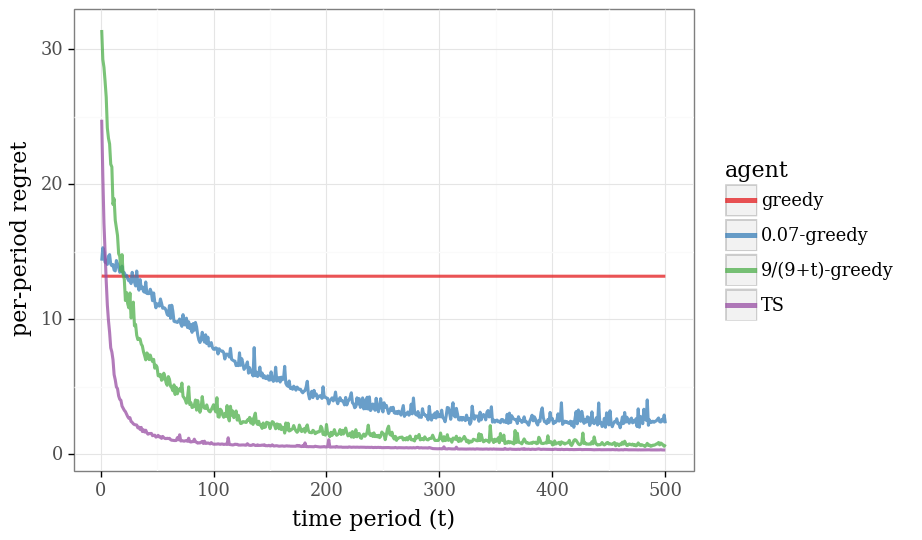}
\caption{Regret experienced by different learning algorithms applied to product assortment problem.}
\label{fig:product_assortment}
\end{figure}

To investigate the performance of TS in this problem, we simulated a scenario with  $n=6$ and $\sigma^2 = 0.04$. We take the profit associated to each product $i$ to be $p_i = 1/6$.  As the prior distribution, we assumed that each  element of $\theta$ is independent and Gaussian-distributed with mean $0$, the diagonal elements have a variance of $1$, and the off-diagonal elements have a variance of $0.2$. To understand this choice, recall the impact of diagonal and off-diagonal elements of $\theta$. The diagonal element $\theta_{ii}$ controls the mean demand when only product $i$ is available, and reflects the inherent quality or popularity of that item. The off-diagonal element $\theta_{ij}$ captures the influence availability of product $j$ has on mean demand for product $i$. Our choice of prior covariance encodes that the dominant effect on demand of a product is likely its own characteristics, rather than its interaction with any single other product. Figure \ref{fig:product_assortment} presents the performance of different learning algorithms in this problem. In addition to TS, we have simulated the greedy and $\epsilon$-greedy algorithms for various values of $\epsilon$. We found that $\epsilon = 0.07$ provides the best performance for $\epsilon$-greedy in this problem.

As illustrated by this figure, the greedy algorithm performs poorly in this problem while $\epsilon$-greedy presents a much better performance. We found that the performance of $\epsilon$-greedy can be improved by using an annealing $\epsilon$  of $\frac{m}{m+t}$ at each period $t$. Our simulations suggest  using  $m=9$ for the best performance in this problem.
 Figure \ref{fig:product_assortment} shows that TS outperforms both variations of $\epsilon$-greedy in this problem.

\section{Cascading Recommendations}
\label{sec:cascadingRecommendation}

We consider an online recommendation problem in which an agent learns to recommend a desirable list of \emph{items} to a \emph{user}.
As a concrete example, the agent could be a search engine and the items could be web pages.  We consider formulating this problem as
a {\it cascading bandit}, in which user selections are governed by a {\it cascade model},
as is commonly used in the fields of information retrieval and online advertising \citep{craswell2008experimental}.

A {\it cascading bandit} model is identified by a triple $(K, J, \theta)$, where $K$ is the number of items, 
$J \leq K$ is the number of items recommended in each period, and $\theta \in [0,1]^K$ 
is a vector of {\it attraction probabilities} .
At the beginning of each $t$th period, the agent selects and presents to the user an ordered list $x_t \in \{1,\ldots,K\}^J$.
The user \emph{examines} items in $x_t$ sequentially, starting from $x_{t,1}$.  Upon examining item $x_{t,j}$,
the user finds it attractive with probability $\theta_{x_{t,j}}$.  In the event that the user finds the item attractive, he 
selects the item and leaves the system.  Otherwise, he carries on to examine the next item in the list, unless 
$j=J$, in which case he has already considered all recommendations and leaves the system.

The agent observes $y_t=j$ if the user selects $x_{t,j}$ and $y_t =\infty$ if the user does not click any item.  
The associated reward $r_t = r(y_t)=\mathbf{1} \{ y_t \leq J\}$ indicates whether any item was selected. 
For each list $x=(x_1, \ldots, x_J)$ and $\theta' \in [0,1]^K$, let
\[
h(x, \theta')=1 - \textstyle \prod_{j=1}^J \left[ 1 -\theta'_{x_j}\right].
\]
Note that 
the expected reward at time $t$ is $\E \left[r_t \middle | x_t , \theta \right] = h(x_t, \theta)$.  The optimal solution
$x^* \in \argmax_{x: \, |x|=J} h(x, \theta)$ consists of the $J$ items with largest attraction probabilities. 
Per-period regret is given by
$
\mathrm{regret}_t(\theta)=h(x^*, \theta) - h(x_t, \theta)$.

\noindent
\begin{minipage}[t]{2.3in}
  \vspace{0pt}
\begin{algorithm}[H]
\begin{scriptsize}
\caption{{\footnotesize $\text{CascadeUCB}(K, J, \alpha, \beta)$}}\label{alg:cascadeucb}
\begin{algorithmic}[1]
\For{$t=1, 2,\ldots $}
\State \textcolor{blue}{\#compute itemwise UCBs:}
\For{$k=1, \ldots, K$}
\State Compute UCB $\mathrm{U}_t(k)$
\EndFor\\
\State \textcolor{blue}{\#select and apply action:}
\State $x_t \leftarrow \argmax_{x: |x|=J} h(x, \mathrm{U}_t)$
\State Apply $x_t$ and observe $y_t$ and $r_t$ \\
\State \textcolor{blue}{\#update sufficient statistics:}
\For{$j=1, \ldots, \min \{ y_t, J \}$}
\State $\alpha_{x_{t,j}} \leftarrow \alpha_{x_{t,j}} + \mathbf{1}(j=y_t)$
\State $\beta_{x_{t,j}} \leftarrow \beta_{x_{t,j}} + \mathbf{1}(j<y_t)$
\EndFor
\EndFor
\end{algorithmic}
\end{scriptsize}
\end{algorithm}
\end{minipage}%
\hspace{0.1in}
\begin{minipage}[t]{2.22in}
  \vspace{0pt}
\begin{algorithm}[H]
\begin{scriptsize}
\caption{{\footnotesize $\text{CascadeTS}(K, J, \alpha, \beta)$}}\label{alg:cascadets}
\begin{algorithmic}[1]
\For{$t=1,2,\ldots $}
\State \textcolor{blue}{\#sample model:}
\For{$k=1, \ldots, K$}
\State Sample $\hat{\theta}_k \sim \text{Beta}(\alpha_k, \beta_k)$
\EndFor \\
\State \textcolor{blue}{\#select and apply action:}
\State $x_t \leftarrow \argmax_{x: |x|=J} h(x, \hat{\theta})$
\State Apply $x_t$ and observe $y_t$ and $r_t$ \\
\State \textcolor{blue}{\#update posterior:}
\For{$j=1, \ldots, \min \{ y_t, J \}$}
\State $\alpha_{x_{t,j}} \leftarrow \alpha_{x_{t,j}} + \mathbf{1}(j=y_t)$
\State $\beta_{x_{t,j}} \leftarrow \beta_{x_{t,j}} + \mathbf{1}(j<y_t)$
\EndFor
\EndFor
\end{algorithmic}
\end{scriptsize}
\end{algorithm}
\end{minipage}
\vspace{0.3in}

\citet{kveton2015cascading} proposed learning algorithms for cascading bandits based on itemwise upper confidence bound (UCB) estimates.
CascadeUCB (Algorithm~\ref{alg:cascadeucb}) is a practical variant that allows for specification of prior parameters $(\alpha,\beta)$
that guide the early behavior of the algorithm.  CascadeUCB computes a UCB $\mathrm{U}_t(k)$ for each item $k \in \{1,\ldots,K\}$ and then chooses a list 
that maximizes $h(\cdot, \mathrm{U}_t)$, which represents an upper confidence bound on the list attraction probability.  The list $x_t$ can be efficiently generated 
by choosing the $J$ items with highest UCBs.
Upon observing the user's response, the algorithm updates the sufficient statistics $(\alpha, \beta)$, which count clicks and views for all the examined items. 
CascadeTS (Algorithm~\ref{alg:cascadets}) is a Thompson sampling algorithm for cascading bandits. 
CascadeTS operates in a manner similar to CascadeUCB except that
$x_t$ is computed based on the sampled attraction probabilities $\hat{\theta}$, rather than the itemwise UCBs $\mathrm{U}_t$.

In this section, we consider a specific form of UCB, which is defined by
\[
\mathrm{U}_t(k)=\frac{\alpha_k}{\alpha_k+\beta_k}+ c \sqrt{\frac{1.5 \log(t)}{\alpha_k+\beta_k}},
\]
for $k \in \{1,\ldots,K\}$, where $\alpha_k / (\alpha_k + \beta_k)$ represents the expected value of the attraction probability $\theta_k$, while
the second term represents an optimistic boost that encourages exploration. Notice that the parameter $c \geq 0$ controls the \emph{degree of optimism}.
When $c=1$, the above-defined UCB reduces to the standard UCB1, which is considered and analyzed in the context of cascading bandits in \citep{kveton2015cascading}. In practice, we can select $c$ through simulations to optimize performance.

\begin{figure}[htpb]
\centering
\includegraphics[width=4in]{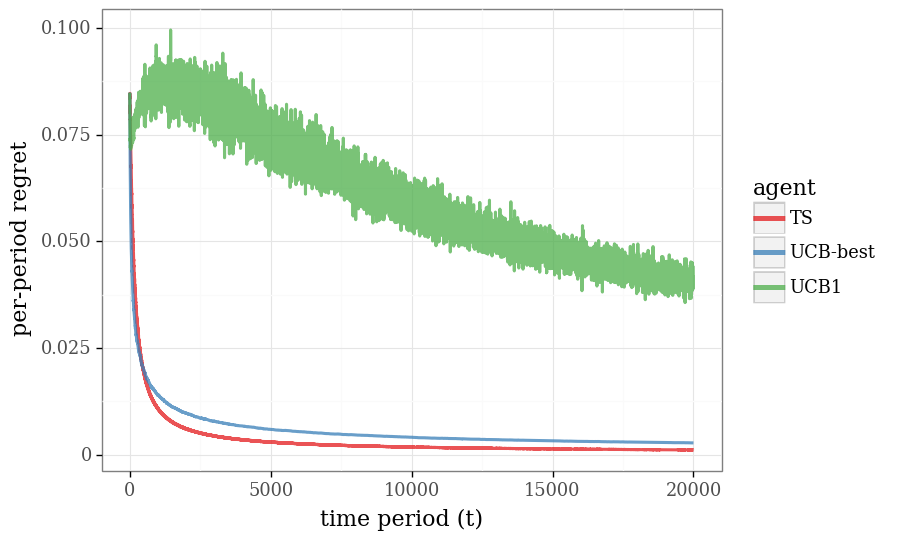}
\caption{Comparison of CascadeTS and CascadeUCB with $K=1000$ items and $J=100$ recommendations per period.}
\label{fig:cascade_coherent}
\end{figure}

Figure \ref{fig:cascade_coherent} presents results from applying CascadeTS and CascadeUCB based on UCB1.
These results are generated by randomly sampling $1000$ cascading bandit instances, $K=1000$ and $J=100$, in each case sampling each attraction probability $\theta_k$ independently from $\mathrm{Beta}(1, 40)$.  For each instance, CascadeUCB and CascadeTS are applied over 20000 time periods, initialized with $(\alpha_k, \beta_k)=(1,40)$.  The plots are of per-period regrets averaged over the $1000$ simulations.  

The results demonstrate that TS far outperforms this version of CascadeUCB.  Why?
An obvious reason is that $h(x, \mathrm{U}_t)$ is far too optimistic. In particular, $h(x, \mathrm{U}_t)$ represents the probability of a click if \emph{every} item in $x$ \emph{simultaneously} takes on the largest attraction probability that is statistically plausible.  However, due to the statistical independence of item attractions,
the agent is unlikely to have substantially under-estimated the attraction probability of every item in $x$. As such, $h(x, \mathrm{U}_t)$ tends to be far too optimistic. CascadeTS, on the other hand, samples components $\hat{\theta}_k$ independently across items. While any sample $\hat{\theta}_k$ might deviate substantially from its mean, it is unlikely that the sampled attraction probability of every item in $x$ greatly exceeds its mean.  As such, the variability in $h(x, \hat{\theta})$ provides a much more accurate reflection of the magnitude of uncertainty.

The plot labeled ``UCB-best'' in Figure \ref{fig:cascade_coherent}
illustrates performance of CascadeUCB with $c = 0.05$, which approximately minimizes cumulative regret over 20,000 time periods.  
It is interesting
that even after being tuned to the specific problem and horizon, the performance of CascadeUCB falls short of Cascade TS.
A likely source of loss stems from the shape of confidence sets used by CascadeUCB.  Note that the algorithm uses
hyper-rectangular confidence sets, since the set of statistically plausible attraction probability vectors is characterized by a Cartesian product 
item-level confidence intervals.  However, the Bayesian central limit theorem suggests that ``ellipsoidal" confidence sets offer a more suitable choice. 
Specifically, as data is gathered, the posterior distribution over $\theta$ can be well approximated by a multivariate Gaussian, for which level sets are ellipsoidal. 
Losses due to the use of hyper-rectangular confidence sets have been studied through regret analysis in \citep{dani2008stochastic} and through
simple analytic examples in \citep{OsbandRLDM2017}.

\begin{figure}[htpb]
\centering
\includegraphics[width=4in]{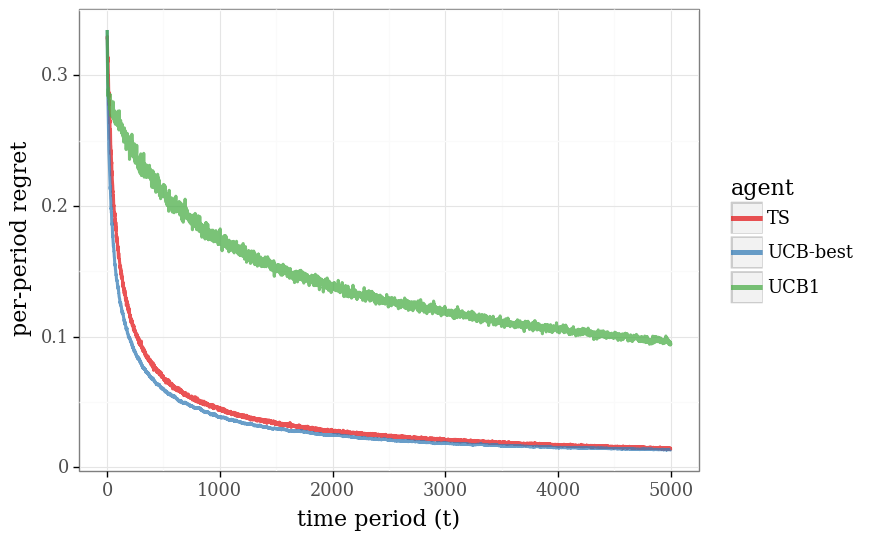}
\caption{Comparison of CascadeTS and CascadeUCB with $K=50$ items and $J=10$ recommendations per period.}
\label{fig:cascade_small}
\end{figure}

It is worth noting that tuned versions of CascadeUCB do sometimes perform as well or better than CascadeTS.  
Figure~\ref{fig:cascade_small} illustrates an example
of this.  The setting is identical to that used to generate the results of Figure~\ref{fig:cascade_coherent}, except that $K=50$ and $J=10$, and cumulative regret
is approximately optimized with $c=0.1$.  CascadeUCB with the optimally tuned $c$ outperforms CascadeTS.  This qualitative difference from the case of $K=1000$ and $J=100$
is likely due to the fact that hyper-rectangular sets offer poorer approximations of ellipsoids as the dimension increases.  This phenomenon
and its impact on regret aligns with theoretical results of \citep{dani2008stochastic}.  That said, CascadeUCB is somewhat advantaged in this comparison 
because it is tuned specifically for the setting and time horizon.


\section{Active Learning with Neural Networks}
\label{se:ensemble}

Neural networks are widely used in supervised learning, where given an existing set of predictor-response data pairs, the objective is to produce
a model that generalizes to accurately predict future responses conditioned on associated predictors.  They are also increasingly being used
to guide actions ranging from recommendations to robotic maneuvers.  Active learning is called for to close the loop by
generating actions that do not solely maximize immediate performance but also probe the environment to generate
data that accelerates learning.  TS offers a useful principle upon which such active learning algorithms can be developed.

With neural networks or other complex model classes, computing the posterior distribution over models becomes intractable.  Approximations
are called for, and incremental updating is essential because fitting a neural network is a computationally intensive task in its own right.
In such contexts, ensemble sampling offers a viable approach \citep{LuVR2017}.  In Section \ref{se:incremental},
we introduced a particular mechanism for ensemble sampling based on the bootstrap.  In this section, we consider an alternative
version of ensemble sampling and present results from \citep{LuVR2017} that demonstrate its application to active learning with neural networks.

To motivate our algorithm, let us begin by discussing how it can be applied to the linear bandit problem.
\begin{example}
\label{ex:lb}
{(Linear Bandit)}  Let $\theta$ be drawn from $\Re^M$ and distributed according to a $N(\mu_0, \Sigma_0)$ prior.  There is a set of $K$ actions $\mathcal{X} \subseteq \Re^M$.  At each time $t = 1,\ldots,T$, an action $x_t \in \mathcal{X}$ is selected, after which a reward $r_t = y_t = \theta^\top x_t + w_t$ is observed, where $w_t \sim N(0, \sigma^2_w)$.
\end{example}
In this context, ensemble sampling is unwarranted, since exact Bayesian inference can be carried out efficiently via Kalman filtering.
Nevertheless, the linear bandit offers a simple setting for explaining the workings of an ensemble sampling algorithm.

Consider maintaining a covariance matrix updated according to
\[ \Sigma_{t+1} = \left(\Sigma_t^{-1} + x_t x_t^\top / \sigma_w^2\right)^{-1}, \]
and $N$ models $\overline{\theta}_t^1,\ldots,\overline{\theta}_t^N$, initialized with $\overline{\theta}_1^1,\ldots,\overline{\theta}_1^N$ each
drawn independently from $N(\mu_0,\Sigma_0)$ and updated incrementally according to
\[ \overline{\theta}_{t+1}^n = \Sigma_{t+1} \left(\Sigma_t^{-1} \overline{\theta}_t^n + x_t (y_t + \tilde{w}_t^n) / \sigma_w^2\right), \]
for $n=1,\ldots,N$, where $(\tilde{w}_t^n: t=1,\ldots,T, n=1,\ldots,N)$ are independent $N(0,\sigma_w^2)$ random samples drawn by the updating algorithm.
It is easy to show that the resulting parameter vectors satisfy
\[ \overline{\theta}_t^n = \arg\min_{\nu} \left(\frac{1}{\sigma_w^2} \sum_{\tau=1}^{t-1} (y_\tau + \tilde{w}_{\tau}^n - x_\tau^\top \nu)^2 + (\nu - \overline{\theta}_1^n)^\top \Sigma_0^{-1} (\nu - \overline{\theta}_1^n)\right). \]
Thich admits an intuitive interpretation: each $\overline{\theta}_t^n$ is a model fit to a randomly perturbed prior and randomly perturbed observations.
As established in \citep{LuVR2017}, for any deterministic sequence $x_1,\ldots,x_{t-1}$, conditioned
on the history, the models $\overline{\theta}_t^1, \ldots, \overline{\theta}_t^N$ are independent and identically distributed according to the posterior distribution of $\theta$.
In this sense, the ensemble approximates the posterior.

The ensemble sampling algorithm we have described for the linear bandit problem motivates an analogous approach for the following neural network
model.
\begin{example}
\label{ex:nn}
{(Neural Network)}  Let $g_\theta:\Re^M \mapsto \Re^K$ denote a mapping induced by a neural network with weights $\theta$.  Suppose there are $K$ actions $\mathcal{X} \subseteq \Re^M$, which serve as inputs to the neural network, and the goal is to select inputs that yield desirable outputs.  At each time $t = 1,\ldots,T$, an action $x_t \in \mathcal{X}$ is selected, after which $y_t = g_\theta(x_t) + w_t$ is observed, where $w_t \sim N(0, \sigma^2_w I)$.  A reward $r_t = r(y_t)$ is associated with each observation.  Let $\theta$ be distributed according to a $N(\mu_0, \Sigma_0)$ prior.  The idea here is that data pairs $(x_t,y_t)$ can be used to fit a neural network model, while actions are selected to trade off between generating data pairs that reduce uncertainty in neural network weights and those that offer desirable immediate outcomes.
\end{example}
Consider an ensemble sampling algorithm that once again begins with $N$ independent models with connection weights
$\overline{\theta}_1^1, \ldots, \overline{\theta}_1^N$ sampled from a $N(\mu_0, \Sigma_0)$ prior.  It could be natural here to let $\mu_0 = 0$
and $\Sigma_0 = \sigma_0^2 I$ for some variance $\sigma_0^2$ chosen so that the range of probable models spans plausible outcomes.
To incrementally update parameters, at each time $t$, each $n$th model applies some number of stochastic gradient descent iterations to reduce a
loss function of the form
\[ \mathcal{L}_t(\nu) = \frac{1}{\sigma_w^2} \sum_{\tau=1}^{t-1} (y_\tau + \tilde{w}_\tau^n - g_\nu(x_\tau))^2 + (\nu - \overline{\theta}_1^n)^\top \Sigma_0^{-1} (\nu - \overline{\theta}_1^n). \]

\begin{figure}[h]
\centering
\begin{subfigure}{0.28\linewidth}
    \centering
    \includegraphics[height=5.8cm]{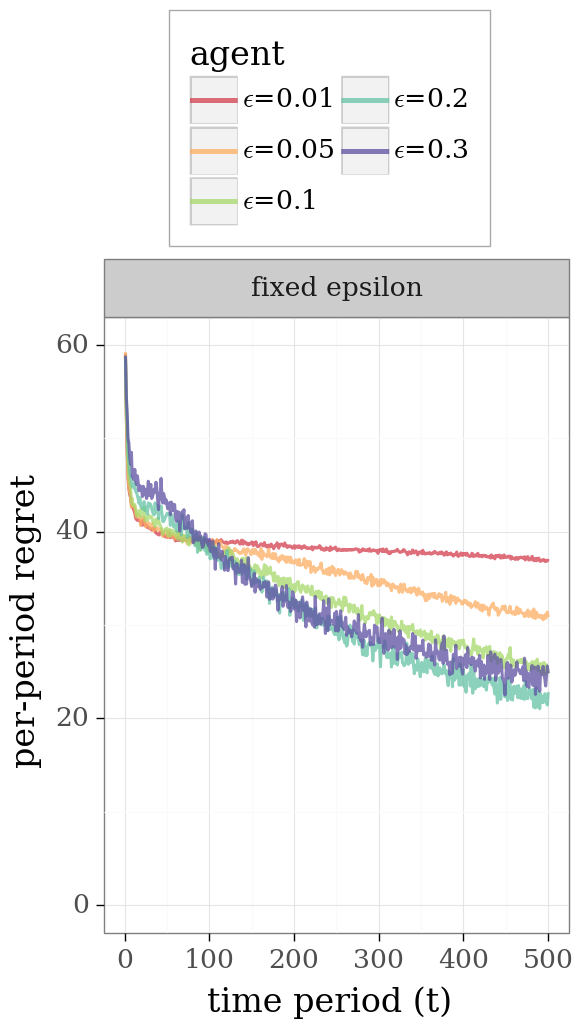}
    \caption{Fixed $\epsilon$-greedy.}
    \label{fig:nn-fixed}
\end{subfigure}
\quad
\begin{subfigure}{0.3\linewidth}
    \centering
    \includegraphics[height=5.8cm]{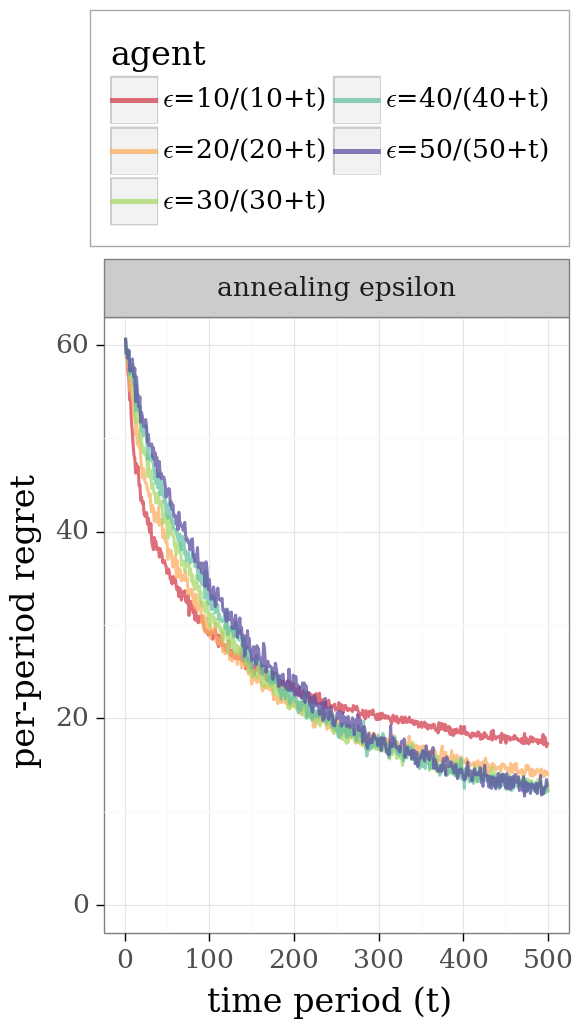}
    \caption{Annealing $\epsilon$-greedy.}
    \label{fig:nn-anneal}
\end{subfigure}
\quad
\begin{subfigure}{0.3\linewidth}
    \centering
    \includegraphics[height=5.8cm]{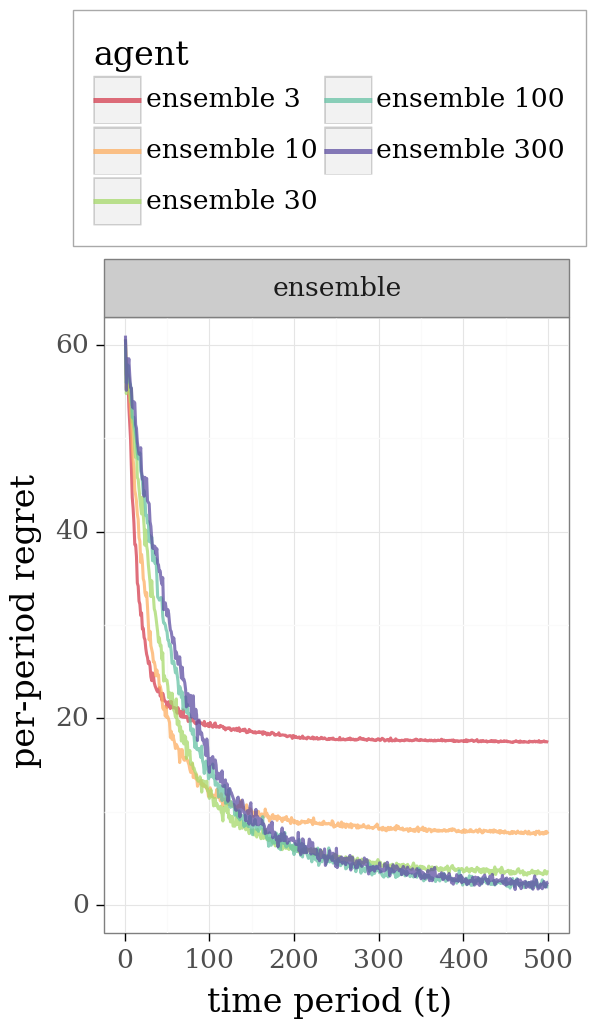}
    \caption{Ensemble TS.}
    \label{fig:nn-ensemble}
\end{subfigure}
\caption{Bandit learning with an underlying neural network.}
\label{fig:neural-net}
\end{figure}

Figure \ref{fig:neural-net} present results from simulations involving a two-layer neural network, with a set of $K$ actions, $\mathcal{X} \subseteq \Re^M$.  The weights of the neural network, which we denote by $w_1 \in \Re^{D \times N}$ and $w_2 \in \Re^{D}$, are each drawn from $N(0, \lambda)$. Let $\theta \equiv (w_1, w_2)$. The mean reward of an action $x \in  \mathcal{X}$ is given by $g_\theta(x) = w_2^\top \max(0, w_1 a) $. At each time step, we select an action $x_t \in \mathcal{X}$ and observe reward $y_t = g_\theta(x_t) + z_t$, where $z_t \sim N(0, \sigma_z^2)$.
We used $M = 100$ for the input dimension, $D = 50$ for the dimension of the hidden layer, number of actions $K = 100$, prior variance $\lambda = 1$, and noise variance $\sigma_z^2 = 100$. Each component of each action vector is sampled uniformly from $[-1, 1]$, except for the last component, which is set to 1 to model a constant offset.   All results are averaged over 100 realizations.

In our application of the ensemble sampling algorithm we have described, to facilitate gradient flow, we use leaky rectified linear units of the form $\max(0.01x, x)$ during training, though the target neural network is made up of regular rectified linear units as indicated above. In our simulations, each update was carried out with 5 stochastic gradient steps, with a learning rate of $10^{-3}$ and a minibatch size of 64.

\label{se:computations_nn}

Figure \ref{fig:neural-net} illustrates the performance of several learning algorithms with an underlying neural network.
Figure \ref{fig:nn-fixed} demonstrates the performance of an $\epsilon$-greedy strategy across various levels of $\epsilon$.
We find that we are able to improve performance with an annealing schedule $\epsilon=\frac{k}{k+t}$ (Figure \ref{fig:nn-anneal}).
However, we find that an ensemble sampling strategy outperforms even the best tuned $\epsilon$-schedules (Figure \ref{fig:nn-ensemble}).
Further, we see that ensemble sampling strategy can perform well with remarkably few members of this ensemble.
Ensemble sampling with fewer members leads to a greedier strategy, which can perform better for shorter horizons, but is prone to premature and suboptimal convergence compared to true TS \citep{LuVR2017}.
In this problem, using an ensemble of as few as 30 members provides very good performance.

\section{Reinforcement Learning in Markov Decision Processes}

Reinforcement learning (RL) extends upon contextual online decision problems to allow for delayed feedback and long term consequences \citep{sutton1998reinforcement, littman2015reinforcement}.
Concretely (using the notation of Section \ref{se:constraints}) the response $y_t$ to the action $x_t$ depends on a context $z_t$; but we no longer assume that the evolution of the context $z_{t+1}$ is independent of $y_t$.
As such, the action $x_t$ may affect not only the reward $r(y_t)$ but also, through the effect upon the context $z_{t+1}$ the rewards of future periods $\left(r(y_{t'})\right)_{t' > t}$.
As a motivating example, consider a problem of sequential product recommendations $x_t$ where the customer response $y_t$ is influenced not only by the quality of the product, but also the history of past recommendations.
The evolution of the context $z_{t+1}$ is then directly affected by the customer response $y_t$; if a customer watched `The Godfather' and loved it, then chances are probably higher they may enjoy `The Godfather 2.'

Maximizing cumulative rewards in a problem with long term consequences can require planning with regards to future rewards, rather than optimizing each period myopically.
Similarly, efficient exploration in these domains can require balancing not only the information gained over a single period; but also the potential for future informative actions over subsequent periods.
This sophisticated form of temporally-extended exploration, which can be absolutely critical for effective performance, is sometimes called \textit{deep exploration} \citep{osband2017deep}.
TS can be applied successfully to reinforcement learning \citep{osband2013more}.
However, as we will discuss, special care must be taken with respect to the notion of a {\em time period} within TS to preserve deep exploration.

Consider a finite horizon Markov decision process (MDP) $M = (\Sc, \Ac, R^M \hspace{-1mm}, P^M\hspace{-1mm}, H, \rho)$,
where $\Sc$ is the state space, $\Ac$ is the action space, and $H$ is the horizon.  The agent begins in a state $s_0$, sampled from $\rho$, 
and over each timestep $h=0,..,H-1$ the agent selects action $a_h \in \Ac$, receives a reward $r_h \sim R^M_{s_h,a_h}$, and transitions to a new state $s_{h+1} \sim P^M_{s_h, a_h}$.
Here, $R^M_{s_h,a_h}$ and $P^M_{s_h, a_h}$ are probability distributions.
A policy $\mu$ is a function mapping each state $s \in \Sc$ and timestep $h=0,..,H-1$ to an action $a \in \Ac$.
The value function $V^M_{\mu,h}(s) = \E [ \sum_{j=h}^{H-1} r_j(s_j, \mu(s_j,j)) \mid s_h=s ]$ encodes the expected reward accumulated under $\mu$ over the remainder of the episode when starting from state $s$ and timestep $h$. Finite horizon MDPs model delayed consequences of actions through the evolution of the state, but the scope of this influence is limited to within an individual episode.

Let us consider an episodic RL problem, in which an agent learns about $R^M$ and $P^M$ over episodes of interaction with an MDP.
In each episode, the agent begins in a random state, sampled from $\rho$, and follows a trajectory, selecting actions and observing rewards and transitions over $H$
timesteps.
Immediately we should note that we have already studied a finite horizon MDP under different terminology in Example \ref{ex:shortest-path}: the online shortest path problem.
To see the connection, simply view each vertex as a state and the choice of edge as an action within a timestep.
With this connection in mind we can express the problem of maximizing the cumulative rewards $\sum_{k=1}^K \sum_{h=0}^{H-1} r(s_{kh}, a_{kh})$ in a finite horizon MDP equivalently as an online decision problem over periods $k=1,2,..,K$, each involving the selection of a policy $\mu_k$ for use over an episode of interaction between the agent and the MDP.
By contrast, a naive application of TS to reinforcement learning that samples a new policy for each timestep within an episode could be extremely inefficient as it does not perform deep exploration.

\begin{figure}[h!]
\centering
\includegraphics[width=0.7\linewidth]{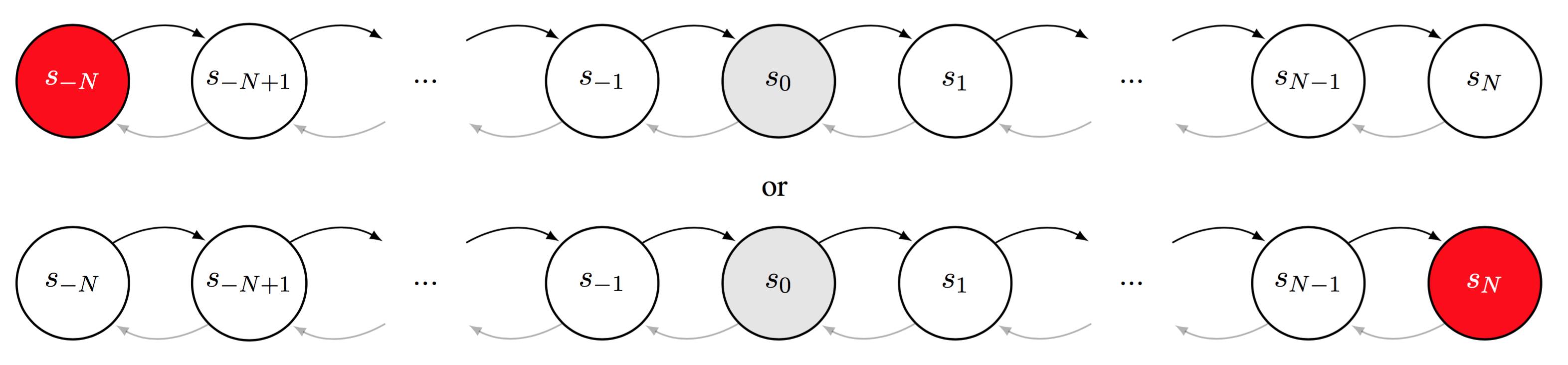}
\vspace{-2mm}
\caption{MDPs where TS with sampling at every timestep within an episode leads to inefficient exploration.}
\label{fig: thompson bad}
\vspace{-2mm}
\end{figure}

Consider the example in Figure \ref{fig: thompson bad} where the underlying MDP is characterized by a long chain of states $\{s_{-N},..,s_{N}\}$ and only the one of the far left or far right positions are rewarding with equal probability; all other states produce zero reward and with known dynamics.
Learning about the true dynamics of the MDP requires a consistent policy over $N$ steps right or $N$ steps left; a variant of TS that resamples after each step would be exponentially unlikely to make it to either end within $N$ steps \citep{osband2017deep}.
By contrast, sampling only once prior to each episode and holding the policy fixed for the duration of the episode demonstrates deep exploration and results in learning the optimal policy within a single episode.

In order to apply TS to policy selection we need a way of sampling from the posterior distribution for the optimal policy.
One efficient way to do this, at least with tractable state and action spaces, is to maintain a posterior distribution over the rewards $R^M$ and the transition dynamics $P^M$ at each state-action pair $(s,a)$.
In order to generate a sample for the optimal policy, simply take a single posterior sample for the reward and transitions and then solve for the optimal policy for this \textit{sample}.
This is equivalent sampling from the posterior distribution of the optimal policy, but may be computationally more efficient than maintaining that posterior distribution explicitly.
Estimating a posterior distribution over rewards is no different from the setting of bandit learning that we have already discussed at length within this paper.
The transition function looks a little different, but for transitions over a finite state space the Dirichlet distribution is a useful conjugate prior.
It is a multi-dimensional generalization of the Beta distribution from Example \ref{ex:beta-bernoulli}.
The Dirichlet prior over outcomes in $\Sc = \{1,..,S\}$ is specified by a positive vector of pseudo-observations $\alpha \in \R_+^S$; updates to the Dirichlet posterior can be performed simply by incrementing the appropriate column of $\alpha$ \citep{strens2000bayesian}.

\begin{figure}[htpb]
\centering
    \begin{subfigure}{.49\textwidth}
        \centering
  \includegraphics[width=\linewidth]{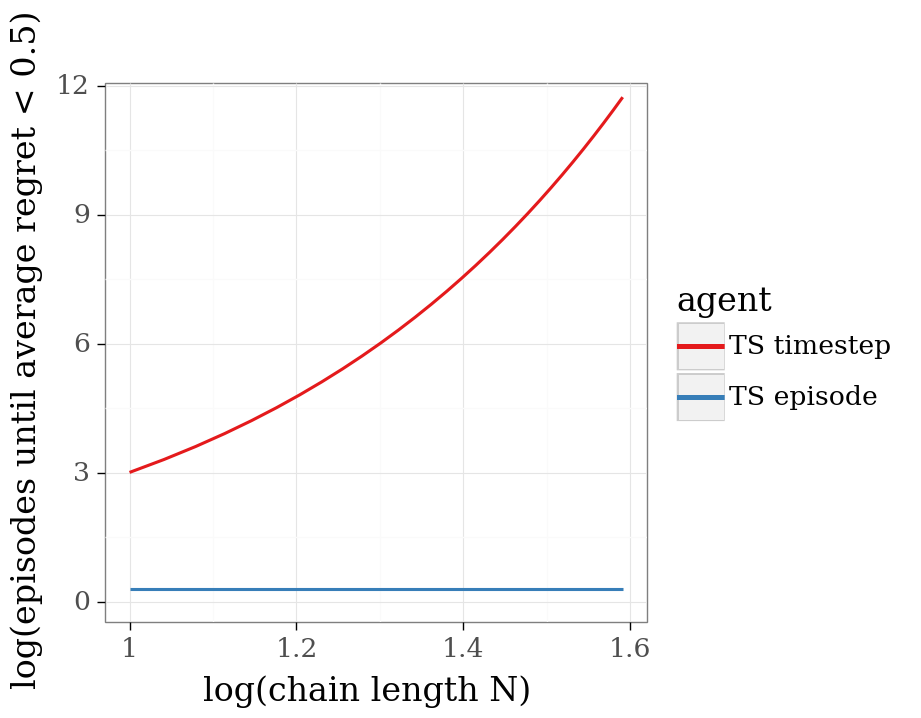}
  \caption{Using informed prior.}
   \label{fig:thompson_mdp_extreme}
    \end{subfigure}
    \begin{subfigure}{.49\textwidth}
        \centering
  \includegraphics[width=\linewidth]{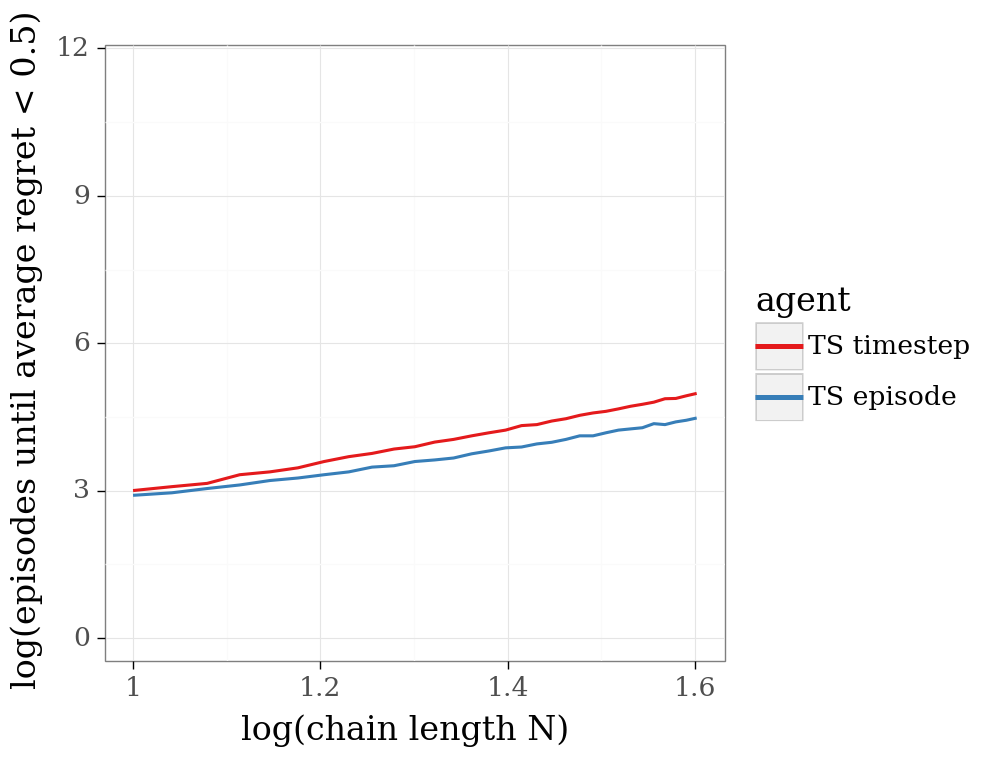}
  \caption{Using uninformed prior.}
   \label{fig:thompson_mdp_dirichlet}
    \end{subfigure}
\caption{TS with sampling per timestep versus per episode.}
\label{fig:thompson_mdp}
\end{figure}

In Figure \ref{fig:thompson_mdp} we present a computational comparison of TS with sampling per timestep versus per episode, applied to the example of Figure \ref{fig: thompson bad}.
Figure \ref{fig:thompson_mdp_extreme} compares the performance of sampling schemes where the agent has an informative prior that matches the true underlying system.
As explained above, sampling once per episode TS is guaranteed to learn the true MDP structure in a single episode.
By contrast, sampling per timestep leads to uniformly random actions until either $s_{-N}$ or $s_{N}$ is visited.
Therefore, it takes a minimum of $2^N$ episodes for the first expected reward.

The difference in performance demonstrated by Figure \ref{fig:thompson_mdp_extreme} is particularly extreme because the prior structure means that there is only value to deep exploration, and none to `shallow' exploration \citep{osband2017deep}.
In Figure \ref{fig:thompson_mdp_dirichlet} we present results for TS on the same environment but with a uniform Dirichlet prior over transitions and a standard Gaussian prior over rewards for each state-action pair.
With this prior structure sampling per timestep is not as hopeless, but still performs worse than sampling per episode.
Once again, this difference increases with MDP problem size.
Overall, Figure \ref{fig:thompson_mdp} demonstrates that the benefit of sampling per episode, rather than per timestep, can become arbitrarily large.
As an additional benefit this approach is also more computationally efficient, since we only need to solve for the optimal policy once every episode rather than at each timestep.

This more nuanced application of TS to RL is sometimes referred to as \textit{posterior sampling for reinforcement learning} (PSRL) \citep{strens2000bayesian}.
Recent work has developed a theoretical analyses of PSRL that guarantee strong expected performance over a wide range of environments \citep{osband2013more,osband2014near,osband2014model,osband2017posterior,ouyang2017learning}.  This work builds on and extends
theoretical results that will be discussed in Section \ref{se:regret}.
It is worth mentioning that PSRL fits in the broader family of Bayesian approaches to efficient reinforcement learning; we refer interested readers to the survey paper \citep{ghavamzadeh2015bayesian}.


\chapter{Why it Works, When it Fails, and Alternative Approaches}

Earlier sections demonstrate that TS approaches can be adapted to address a number of problem classes of practical import. In this section, we provide intuition for why TS explores efficiently, and briefly review theoretical work that formalizes this intuition. We  will then highlight problem classes for which TS is poorly suited, and refer to some alternative algorithms.

\section{Why Thompson Sampling Works}

To understand whether TS is well suited to a particular application, it is useful to develop a high level understanding of why it works. As information is gathered, beliefs about action rewards are carefully tracked. By sampling actions according to the posterior probability that they are optimal, the algorithm continues to sample all actions that could plausibly be optimal, while shifting sampling away from those that are unlikely to be optimal.  Roughly speaking, the algorithm tries all promising actions while gradually discarding those that are believed to underperform.This intuition is formalized in recent theoretical analyses of Thompson  sampling, which we now review.

\subsection{Regret Analysis for Classical Bandit Problems}

\paragraph{Asymptotic Instance Dependent Regret Bounds.}
Consider the classical beta-Bernoulli bandit problem of Example \ref{ex:bernoulli}. For this problem, sharp results on the asymptotic scaling of regret are available. The cumulative regret of an algorithm over $T$ periods is
\[
{\rm Regret}(T)= \sum_{t=1}^{T} \left( \max_{ 1\leq k \leq K } \theta_k - \theta_{x_t} \right),
\]
where $K$ is the number of actions, $x_t\in \{1,\ldots, K\}$ is the action selected at time $t$, and $\theta=(\theta_1,\ldots, \theta_K)$ denotes action success probabilities. For each time horizon $T$, $\E[{\rm Regret}(T) \mid \theta ]$  measures the expected $T$-period regret on the problem instance $\theta$. The conditional expectation integrates over the noisy realizations of rewards and the algorithm's random action selection, holding fixed the success probabilities $\theta=(\theta_1, \ldots, \theta_K)$.  Though this is difficult to evaluate, one can show that
\begin{equation}\label{eq: lai robbins optimality}
\lim_{T\to \infty} \frac{\E[{\rm Regret}(T) \mid \theta ]}{\log(T)} = \sum_{k \neq k^*} \frac{\theta_{k^*} -\theta_{k}}{d_{\text{KL}}(\theta_{k^*} \, || \, \theta_{k} )},
\end{equation}
assuming that there is a unique optimal action $k^*$.  Here, $d_{\text{KL}}(\theta \, || \, \theta') = \theta \log\left( \frac{\theta}{\theta'}  \right) + (1-\theta)\log\left( \frac{1-\theta}{1-\theta'}  \right)$ is the Kullback-Leibler divergence between Bernoulli distributions. The fundamental lower bound of \citep{lai1985asymptotically} shows no algorithm can improve on the scaling in \eqref{eq: lai robbins optimality}, establishing a sense in which the algorithm is asymptotically optimal. That the regret of TS exhibits this scaling was first observed empirically by \citep{chapelle2011empirical}. A series of papers provided proofs that formalize this finding \citep{agrawal2012analysis, agrawal2013further, kaufmann2012thompson}.

This result has been extended to cases where reward distributions are Gaussian or, more generally, members of a canonical one-dimensional exponential family  \citep{honda2014optimality}. It has also been extended to the case of Gaussian distributions with unknown variance by \citep{honda2014optimality}, which further establishes that
this result can fail to hold for a particular improper prior distribution. Although, intuitively, the effects of the prior distribution should wash out as $T\to \infty$, all of these results apply to specific choices of uninformative prior distributions.  Establishing asymptotic optimality of TS for broader classes of prior distributions remains an interesting open issue.

\paragraph{Instance-Independent Regret bounds.}
While the results discussed in the previous section establishes that the regret of TS is optimal in some sense, it is important to understand that this result is asymptotic.
Focusing on this asymptotic scaling enables sharp results, but even for problems with long time horizons, there are substantial performance differences among algorithms known to be asymptotically optimal in the sense of \eqref{eq: lai robbins optimality}.  The bound essentially focuses on a regime in which the agent is highly confident of which action is best but continues to occasionally explore in order to become even more confident. In particular, the bound suggests that for sufficiently large $T$, regret scales like
$$\E[{\rm Regret}(T) \mid \theta ] \approx \sum_{k \neq k^*} \frac{\theta_{k^*} -\theta_{k}}{d_{\text{KL}}(\theta_{k^*} \, || \, \theta_{k} )} \log(T).$$
This becomes easier to interpret if we specialize to the case in which rewards, conditioned on $\theta$, are Gaussian with unit variance, for which
$d_{\text{KL}}(\theta || \theta' ) = (\theta-\theta')^2/2$, and therefore,
\begin{equation}\label{eq: gauss log bound}
\E[{\rm Regret}(T) \mid \theta ] \approx \sum_{k \neq k^*} \frac{2}{\theta_{k^*} -\theta_{k}} \log(T).
\end{equation}
The fact that the final expression is dominated by near-optimal actions reflects that in the relevant asymptotic regime other actions can be essentially ruled out using
far fewer samples.

A more subtle issue is that $O(\log(T))$ regret bounds like those described above become vacuous for problems with nearly-optimal actions,
since the right-side of \ref{eq: gauss log bound} can become arbitrarily large.
This issue is particularly limiting for complex structured online decision problems, where there are often a large or even infinite number of near-optimal actions.

 For the Bernoulli bandit problem of Example \ref{ex:bernoulli},
\citep{agrawal2013further} establishes that when TS is initialized with a uniform prior,
\begin{equation}\label{eq: minimax regret}
\max_{\theta'} \E[{\rm Regret}(T) \mid \theta= \theta' ] = O\left(\sqrt{KT\log(T)}\right).
\end{equation}
This regret bounds holds uniformly over all problem instances, ensuring that there are no instances of bandit problems with binary rewards that will cause the regret of TS to explode.  This bound is nearly order-optimal, in the sense that there exists a distribution over problem instances under which the expected regret of any algorithm is at least $\Omega(\sqrt{KT})$ \citep{bubeck2012regret}.

\subsection{Regret Analysis for Complex Online Decision Problems}
\label{se:regret}

This tutorial has covered the use of TS to address an array of complex online decision problems. In each case, we first modeled the problem at hand, carefully encoding prior knowledge. We then applied TS, trusting it could leverage this structure to accelerate learning. The results described in the previous subsection are deep and interesting, but do not justify using TS in this manner.

We will now describe alternative theoretical analyses of TS that apply very broadly. These analyses point to TS's ability to exploit problem structure and prior knowledge, but also to settings where TS performs poorly.

\subsubsection{Problem Formulation}

Consider the following general class of online decision problems. In each period $t\in \mathbb{N}$, the agent selects an action $x_t \in \Xc$, observes an outcome $y_t$, and associates this with a real-valued reward $r(y_t)$ that is a known function of the outcome.  In the shortest path problem of Examples \ref{ex:log-shortest-path} and \ref{ex:log-shortest-path-cf}, $x_t$ is a path, $y_t$ is a vector encoding the time taken to traverse each edge in that path, and $r_t=r(y_t)$ is the negative sum of these travel times. More generally, for each $t$, $y_t=g\left(x_t, \theta, w_t\right)$ where $g$ is some known function and $(w_t : t\in \mathbb{N})$ are i.i.d and independent of $\theta$. This can be thought of as a Bayesian model, where the random variable $\theta$ represents the uncertain true characteristics of the system and $w_t$ represents idiosyncratic randomness
influencing the outcome in period $t$. Let
\[
\mu(x, \theta) = \E[r\left(g( x, \theta, w_t) \right) \mid \theta ]
\]
denote the expected reward generated by the action $x$ under the parameter $\theta$, where this expectation is taken over the disturbance $w_t$. The agent's uncertainty about $\theta$ induces uncertainty about the identity of the optimal action $x^* \in \argmax_{x\in \Xc} \mu(x, \theta)$.

An algorithm is an adaptive, possibly randomized, rule for selecting an action as a function of the history of actions and observed outcomes. The expected cumulative regret of an algorithm over $T$ periods is
\[
\E\left[{\rm Regret}(T) \right] = \E\left[\sum_{t=1}^{T} \left( \mu(x^*, \theta) - \mu(x_t, \theta) \right) \right].
\]
This expectation is taken over draws of $\theta$, the idiosyncratic noise terms $(w_t,\ldots, w_T)$, and the algorithm's internal randomization over actions. This is sometimes called the algorithm's \emph{Bayesian regret}, since it is integrated over the prior distribution.

It is worth briefly discussing the interpretation of this regret measure. No single algorithm can minimize conditional expected regret $\E[{\rm Regret}(T) \mid \theta = \theta']$ for every problem instance $\theta'$. As discussed in Section \ref{se:prior-specification}, one algorithm may have lower regret than another for one problem instance but have higher regret for a different problem instance. In order to formulate a coherent optimization problem, we must somehow scalarize this objective. We do this here by aiming to minimize integrated regret $\E\left[{\rm Regret}(T)\right] =\E\left[\E[{\rm Regret}(T) \mid \theta] \right]$. Under this objective, the prior distribution over $\theta$ directs the algorithm to prioritize strong performance in more likely scenarios. Bounds on expected regret help certify that an algorithm has efficiently met this objective. An alternative choice is to bound worst-case regret $\max_{\theta'} \E[{\rm Regret}(T) \mid \theta=\theta']$. Certainly, bounds on worst-case regret imply bounds on expected regret, but targeting this objective will rule out the use of flexible prior distributions, discarding one of the TS's most useful features.   In particular, designing an algorithm to minimize worst-case regret typically entails substantial sacrifice of performance with likely values of $\theta$.

\subsubsection{Regret Bounds via UCB}

One approach to bounding expected regret relies on the fact that TS shares a property of UCB algorithms that underlies many of their theoretical guarantees.  Let us begin by discussing how regret bounds are typically established for UCB algorithms.

A prototypical UCB algorithm generates a function $U_t$ based on the history $\hist_{t-1}$
such that, for each action $x$, $U_t(x)$ is an optimistic but statistically plausible estimate of the expected reward, referred to as an upper-confidence bound.
Then, the algorithm selects an action $\overline{x}_t$ that maximizes $U_t$.
There are a variety  of proposed approaches to generating $U_t$ for specific models.  For example, \citep{kaufmann2012bayesian} suggest taking $U_t(x)$
to be the $(1 - 1/t)$th quantile of the posterior distribution of $\mu(x, \theta)$.  A simpler heuristic, which is nearly identical to the UCB1 algorithm presented and analyzed
in \citep{auer2002finite}, selects actions to maximize
$U_t(x) = \E[\mu(x, \theta) |\hist_{t-1}] + \sqrt{2 \ln(t) / t_x}$, where $t_x$ is the number of times action $x$ is selected prior to period $t$.
If $t_x=0$, $\sqrt{2 \ln(t) / t_x} = \infty$, so each action is selected at least once.  As experience with an action accumulates and $\ln(t) / t_x$ vanishes,
$U_t(x)$ converges to $\E[\mu(x, \theta) |\hist_{t-1}]$, reflecting increasing confidence.

With any choice of $U_t$, regret over the period decomposes according to
\begin{eqnarray*}
\mu(x^*, \theta) - \mu(\overline{x}_t, \theta)
&=&  \mu(x^*, \theta) - U_t(\overline{x}_t) + U_t(\overline{x}_t) - \mu(\overline{x}_t, \theta) \\
&\leq&  \underbrace{\mu(x^*, \theta) - U_t(x^*)}_{\text{pessimism}} + \underbrace{U_t(\overline{x}_t) - \mu(\overline{x}_t, \theta)}_{\text{width}}.
\end{eqnarray*}
The inequality follows from the fact that $\bar{x}_t$ is chosen to maximize
$U_t$. If $U_t(x^*) \geq \mu(x^*, \theta)$, which an upper-confidence bound should satisfy with high probability, the pessimism term is negative.
The width term, penalizes for slack in the confidence interval at the selected action $\bar{x}_t$.  For reasonable proposals of $U_t$, the width vanishes over time for actions that are selected repeatedly.  Regret bounds for UCB algorithms are obtained by characterizing the rate at which this slack diminishes as actions are applied.

As established in \citep{russo2014learning}, expected regret bounds for TS can be produced in a similar manner.
To understand why, first note that for any function $U_t$ that is determined by the history $\hist_{t-1}$,
\begin{equation}\label{eq: TS regret decomposition}
\E[U_t(x_t)] = \E[\E[U_t(x_t) | \hist_{t-1}]] = \E[\E[U_t(x^*) | \hist_{t-1}]] = \E[U_t(x^*)].
\end{equation}
The second equation holds because TS samples $x_t$ from the posterior distribution of $x^*$. Note that for this result, it is important that $U_{t}$ is determined by $\hist_{t-1}$. For example, although $x_{t}$ and $x^*$ share the same marginal distribution, in general $\E[\mu(x^*, \theta)] \neq \E[\mu(x_t, \theta) ]$ since the joint distribution of $(x^*, \theta)$ is not identical to that of $(x_t, \theta)$.

From Equation (\ref{eq: TS regret decomposition}), it follows that
\begin{eqnarray*}
 \E\left[\mu(x^*, \theta) - \mu(x_t, \theta)  \right]
&=& \E\left[\mu(x^*, \theta) - U_t(x_t)\right] + \E\left[U_t(x_t) - \mu(x_t, \theta) \right] \\
&=& \underbrace{\E\left[\mu(x^*, \theta) - U_t(x^*)\right]}_{\text{pessimism}} + \underbrace{\E\left[U_t(x_t) - \mu(x_t, \theta) \right]}_{\text{width}}.
\end{eqnarray*}
If $U_t$ is an upper-confidence bound, the pessimism term should be negative, while the width term can be bounded by
arguments identical to those that would apply to the corresponding UCB algorithm.
Through this relation, many regret bounds that apply to UCB algorithms translate immediately to expected regret bounds for TS.

An important difference to take note of is that UCB regret bounds depend on the specific choice of $U_t$
used by the algorithm in question.  With TS, on the other hand, $U_t$ plays no role in the algorithm and appears only
as a figment of regret analysis.  This suggests that, while the regret of a UCB algorithm depends critically on the
specific choice of upper-confidence bound, TS depends only on the best possible choice.  This is
a crucial advantage when there are complicated dependencies among actions,
as designing and computing with appropriate upper-confidence bounds present significant challenges.

Several examples provided in \citep{russo2014learning} demonstrate how UCB regret bounds translate to TS expected regret bounds.  These include a bound that applies to all problems with a finite number of actions, as well as stronger bounds that apply when the reward function $\mu$ follows a linear model, a generalized linear model, or is sampled from a Gaussian process prior.  As an example, suppose mean rewards follow the linear model $\mu(x, \theta)=x^\top \theta$ for $x\in \mathbb{R}^{d}$ and $\theta \in \mathbb{R}^d$ and that reward noise is sub-Gaussian.
It follows from the above relation that existing analyses \citep{dani2008stochastic, rusmevichientong2010linearly, abbasi2011improved} of UCB algorithms imply that under TS
 \begin{equation}\label{eq: linear regret bound}
\E\left[{\rm Regret}(T) \right]= O(d \sqrt{T} \log(T)) .
\end{equation}
This bound applies for any prior distribution over a compact set of parameters $\theta$. The big-$O$ notation assumes several quantities are bounded by constants: the magnitude of feasible actions, the magnitude of $\theta$ realizations, and the variance proxy of the sub-Gaussian noise distribution. An important feature of this bound is that it depends on the complexity of the parameterized model through the dimension $d$, and not on the number of actions. Indeed, when there are a very large, or even infinite, number of actions, bounds like \eqref{eq: minimax regret} are vacuous, whereas \eqref{eq: linear regret bound} may still provide a meaningful guarantee.

In addition to providing a means for translating UCB to TS bounds, results of \citep{russo2014learning,russo2013nipsEluder} unify many of these bounds.
In particular, it is shown that across a very broad class of online decision problems, both TS and well-designed UCB algorithms satisfy
 \begin{equation}\label{eq: eluder regret bound}
\E\left[{\rm Regret}(T) \right] = \tilde{O}\left(\sqrt{ \underbrace{{\rm dim}_E\left(\mathcal{F}, T^{-2} \right)}_{\text{eluder dimension}}\underbrace{\log\left( N\left(\mathcal{F}, T^{-2}, \left\Vert \cdot \right\Vert_\infty \right) \right)}_{\text{log-covering number}} T}\right),
\end{equation}
where $\mathcal{F} = \{\mu(\cdot, \theta): \theta \in \Theta\}$ is the set of possible reward functions, $\Theta$ is the set of possible parameter vectors $\theta$, and $\tilde{O}$ ignores logarithmic factors.
This expression depends on the class of reward functions $\mathcal{F}$ through two measures of complexity. Each captures the approximate structure of the class of functions at a scale $T^{-2}$ that depends on the time horizon. The first measures the growth rate of the covering numbers of $\mathcal{F}$ with respect to the maximum norm, and is closely related to measures of complexity that are common in the supervised learning literature. This quantity  roughly captures the sensitivity of $\mathcal{F}$ to statistical overfitting.  The second measure, the {\it eluder dimension}, captures how effectively the value of unobserved actions can be inferred from observed samples.  This bound can be specialized to particular function classes.  For example, when specialized to the aforementioned linear model, ${\rm dim}_E\left(\mathcal{F}, T^{-2} \right) = O(d \log(T))$ and
$\log\left( N\left(\mathcal{F}, T^{-2}, \left\Vert \cdot \right\Vert_\infty \right) \right) = O(d \log(T))$, and it follows that
$$\E\left[{\rm Regret}(T) \right]= \tilde{O}(d \sqrt{T}).$$
It is worth noting that, as established in \citep{russo2014learning,russo2013nipsEluder}, notions of complexity common to the supervised learning literature such as covering numbers and Kolmogorov and Vapnik-Chervonenkis dimensions are insufficient for bounding regret in online decision problems.  As such, the new notion of eluder dimension introduced in \citep{russo2014learning,russo2013nipsEluder} plays an essential role in (\ref{eq: eluder regret bound}).

\subsubsection{Regret Bounds via Information Theory}

Another approach to bounding regret, developed in \citep{russo2016info}, leverages the tools of information theory.
The resulting bounds more clearly reflect the benefits of prior knowledge, and the analysis points to shortcomings of TS and how they can
addressed by alternative algorithms.  A focal point in this analysis is the notion of an {\it information ratio}, which for any model and online decision algorithm is
defined by
 \begin{equation}\label{eq:info ratio}
\Gamma_t  = \frac{\left(\E\left[\mu(x^*, \theta) - \mu(x_t, \theta)\right]\right)^2 }{I\left(x^* ; (x_t, y_t) | \hist_{t-1} \right)}.
\end{equation}
The numerator is the square of expected single-period regret, while in the denominator, the conditional mutual information $I(x^*; (x_t, y_t) | \hist_{t-1})$
between the uncertain optimal action $x^*$ and the impending observation $(x_t, y_t)$ measures expected information gain.\footnote{An alternative definition of the information ratio -- the expected regret $\left(\E\left[\mu(x^*, \theta) - \mu(x_t, \theta) \mid \hist_{t-1} = h_{t-1}\right] \right)^2$  divided by the mutual information $I\left(x^* ; (x_t, y_t) \mid \hist_{t-1}=h_{t-1} \right)$, both conditioned on a particular history $h_{t-1}$ -- was used in the original paper on this topic \citep{russo2016info}.  That paper established bounds on the information ratio that hold uniformly over possible realizations of $h_{t-1}$. It was observed in \citep{russo2018time} that the same bounds apply with the information ratio defined as in \eqref{eq:info ratio}, which integrates over $h_t$.  The presentation here mirrors the more elegant treatment of these ideas in \citep{russo2018time}.}

The information ratio depends on both the model and algorithm and can be interpreted as an expected ``cost'' incurred per bit of information acquired.
If the information ratio is small, an algorithm can only incur large regret when it is expected to gain a lot of information about which action is optimal.
This suggests that expected regret is bounded in terms of the maximum amount of information any algorithm could expect to acquire,
which is at most the entropy of the prior distribution of the optimal action.  The following regret bound from \citep{russo2016info}, which applies to any model and algorithm,
formalizes this observation:
\begin{equation}\label{eq:info theoretic regret bound}
\E\left[{\rm Regret}(T)  \right] \leq  \sqrt{\overline{\Gamma} H(x^*) T},
\end{equation}
where $\overline{\Gamma} = \max_{t \in \{1,\ldots,T\}} \Gamma_t$.  An important feature of this bound
is its dependence on initial uncertainty about the optimal action $x^*$, measured in terms of the entropy $H(x^*)$.  This captures the benefits of prior information in a way that is missing from previous regret bounds.

A simple argument establishes bound (\ref{eq:info theoretic regret bound}):
\begin{eqnarray*}
\E \left[{\rm Regret}(T)  \right]
&=& \sum_{t=1}^{T} \E\left[\mu(x^*, \theta) - \mu(x_t, \theta)\right] \\
&=& \sum_{t=1}^{T} \sqrt{\Gamma_{t} I\left(x^* ; (x_t, y_t)  | \hist_{t-1}\right)} \\
&\leq& \sqrt{\overline{\Gamma}T \sum_{t=1}^{T} I\left( x^* ; (x_t, y_t) | \hist_{t-1} \right)},
\end{eqnarray*}
where the inequality follows from Jensen's inequality and the fact that $\Gamma_t \leq \overline{\Gamma}$.  Intuitively, $I(x^*, (x_t,y_t) | \hist_{t-1})$ represents the expected information gained about $x^*$, and the sum over periods cannot exceed the entropy $H(x^*)$.  Applying this relation, which is formally established in \citep{russo2016info} via the chain rule of mutual information, we obtain (\ref{eq:info theoretic regret bound}).

It may be illuminating to interpret the bound in the case of TS applied to a shortest path problem.  Here, $r_t$ is the negative travel time of the path selected in period $t$ and we assume the problem has been appropriately normalized so that $r_t \in [-1,0]$ almost surely. For a problem with $d$ edges, $\theta \in \mathbb{R}^d$ encodes the mean travel time along each edge, and $ x^* = x^*(\theta)$ denotes the shortest path under $\theta$.  As established in \citep{russo2016info}, the information ratio can be bounded above by $d/2$,
and therefore, (\ref{eq:info theoretic regret bound}) specializes to $\E[{\rm Regret}(T)] \leq \sqrt{d H(x^*) T / 2}.$
Note that the number of actions in the problem is the number of paths, which can be exponential in the number of edges.
This bound reflects two ways in which TS is able to exploit the problem's structure to nevertheless learn efficiently.  First, it depends on the number of edges $d$ rather than the number of paths. Second, it depends on the entropy $H(x^*)$ of the decision-maker's prior over which path is shortest. Entropy is never larger than the logarithm of the number of paths, but can be much smaller if the agent has informed prior over which path is shortest. Consider for instance the discussion following  Example \ref{ex:log-shortest-path}, where the agent had knowledge of the distance of each edge and believed {\it a priori} that longer edges were likely to require greater travel time; this prior knowledge reduces the entropy of the agent's prior, and the bound formalizes that this prior knowledge improves performance.
 Stronger bounds apply when the agent receives richer feedback in each time period.  At one extreme, the agent observes the realized travel time along every edge in that period, including those she did not traverse. In that case, \citep{russo2016info} establishes that the information ratio is bounded by $1/2$, and therefore, $\E[{\rm Regret}(T)] \leq \sqrt{H(x^*) T/2}$.   The paper also defines a class of problems where the agent observes the time to traverse each individual edge along the chosen path and establishes that the information ratio is bounded by $d/2m$ and $\E[{\rm Regret}(T)] \leq \sqrt{d H(x^*) T/ 2m}$, where $m$ is the maximal number of edges in a path.

The three aforementioned bounds of $d/2$, $1/2$ and $d/2m$ on information ratios reflect the impact of each problem's information structure on the regret-per-bit of information acquired by TS about the optimum. Subsequent work has established bounds on the information ratio for problems with convex reward functions \citep{bubeck2015multi} and  for problems with graph structured feedback \citep{liu2017information}.

The bound of (\ref{eq:info theoretic regret bound}) can become vacuous as the number of actions increases due to the dependence on entropy.
In the extreme, the entropy $H(x^*)$ can become infinite when there are an infinite number of actions.  It may be possible to derive alternative information-theoretic bounds
that depend instead on a rate-distortion function.  In this context, a rate-distortion function should capture the amount of information
required to deliver near-optimal performance.  Connections between rate-distortion theory and online decision problems
have been established in \citep{russo2018time}, which studies a variation of TS that aims to learn satisficing actions.  Use of rate-distortion
concepts to analyze the standard version of TS remains an interesting direction for further work.

\subsubsection{Further Regret Analyses}

Let us now discuss some alternatives to the regret bounds described above. For linear bandit problems, \citep{agrawal2013linear} provides an analysis of TS with an uninformative Gaussian prior. Their results yield a bound on worst-case expected regret of $\min_{\theta': \|\theta'\|_2 \leq 1 }\E[{\rm Regret}(T) \mid \theta = \theta' ]= \tilde{O}\left(d^{3/2}\sqrt{T}\right)$. Due to technical challenges in the proof, this bound does not actually apply to TS with proper posterior updating, but instead to a variant that inflates the variance of posterior samples. This leads to an additional $d^{1/2}$ factor in this bound relative to that in \eqref{eq: linear regret bound}. It is an open question whether a worst-case regret bound can be established for standard TS in this context, without requiring any modification to the posterior samples. Recent work has revisited this analysis and provided improved proof techniques \citep{abeille2017linear}. Furthering this line of work, \citep{agrawal2017thompson} study an assortment optimization problem and provide worst-case regret bounds for an algorithm that is similar to TS but samples from a modified posterior distribution. Following a different approach, \citep{gopalan2014thompson} provides an asymptotic analysis of Thomson sampling for parametric problems with finite parameter spaces.  Another recent line of theoretical work treats  extensions of TS to reinforcement learning \citep{osband2013more, gopalan2015thompson, osband2016generalization, kim2017thompson}.

\subsection{Why Randomize Actions}

TS is a {\it stationary randomized strategy}: randomized in that each action is randomly sampled from a distribution and stationary in that this action distribution is determined by the posterior distribution of $\theta$ and otherwise independent of the time period.  It is natural to wonder whether randomization plays a fundamental role or if a stationary deterministic strategy can offer similar behavior.  The following example from \citep{russo2018IDS} sheds light on this matter.
\begin{example}(A Known Standard)
Consider a problem with two actions $\mathcal{X} = \left\{1, 2 \right\}$ and a binary parameter $\theta$ that is distributed Bernoulli($p_0$).  Rewards from action $1$ are known to be distributed Bernoulli($1/2$).  The distribution of rewards from action $2$ is Bernoulli($3/4$) if $\theta = 1$ and Bernoulli($1/4$) if $\theta = 0$.
\end{example}
Consider a stationary deterministic strategy for this problem.  With such a strategy, each action $x_t$ is a deterministic function of $p_{t-1}$, the probability that $\theta = 1$ conditioned $\hist_{t-1}$.  Suppose that for some $p_0 > 0$, the strategy selects $x_1 = 1$.  Since the resulting reward is uninformative, $p_t = p_0$ and $x_t = 1$ for all $t$, and thus, expected cumulative regret grows linearly with time.  If, on the other hand, $x_1 = 2$ for all $p_0 > 0$, then $x_t = 2$ for all $t$, which again results in expected cumulative regret that grows linearly with time.  It follows that, for any deterministic stationary strategy, there exists a prior probability $p_0$ such that expected cumulative regret grows linearly with time.  As such, for expected cumulative regret to exhibit a sublinear horizon dependence, as is the case with the bounds we have discussed, a stationary strategy must randomize actions.  Alternatively, one can satisfy such bounds via a strategy that is deterministic but nonstationary, as is the case with typical UCB algorithms.

\section{Limitations of Thompson Sampling}

TS is effective across a broad range of problems, but there are contexts in which TS leaves a lot of value on the table.  We now highlight four problem features that are not adequately addressed by TS.

\subsection{Problems that do not Require Exploration}

We start with the simple observation that TS is a poor choice for problems where learning does not require active exploration.  In such contexts, TS is usually outperformed by greedier algorithms that do not invest in costly exploration. As an example, consider the problem of selecting a portfolio made up of publicly traded financial securities. This can be cast as an online decision problem. However, since historical returns are publicly available, it is possible to backtest trading strategies, eliminating the need to engage in costly real-world experimentation.  Active information gathering may become important, though, for traders who trade large volumes of securities over short time periods, substantially influencing market prices, or when information is more opaque, such as in dark pools.

In contextual bandit problems, even when actions influence observations,
randomness of context can give rise to sufficient exploration so that additional active exploration
incurs unnecessary cost.  Results of \citep{bastani2018} formalize conditions under which greedy behavior is effective because of passive exploration induced by contextual randomness. The following example captures the essence of this phenomenon.
\begin{example}{(Contextual Linear Bandit)}
Consider two actions $\mathcal{X}=\{1,2\}$ and parameters $\theta_1$ and $\theta_2$ that are independent and standard-Gaussian-distributed.
A context $z_{t}$ is associated with each time period $t$ and is drawn independently from a standard Gaussian.
In period $t$, the agent selects an action $x_t$ based on prevailing context $z_{t}$, as well as observed history, and then
observes a reward $r_t = z_{t} \theta_{x_t} + w_t$, where $w_t$ is i.i.d. zero-mean noise.
\end{example}
Consider selecting a greedy action $x_t$ for this problem.  Given point estimates $\hat{\theta}_1$ and $\hat{\theta}_2$,
assuming ties are broken randomly, each action is selected with equal probability, with the choice determined by the random context.
This probing of both actions alleviates the need for active exploration, which would decrease immediate reward.
It is worth noting, though, that active exploration can again become essential if the context variables are binary-valued with $z_{t}\in \{0,1\}$. In particular, if the agent converges on a point estimate $\hat{\theta}_1 = \theta_1 > 0$, and action $2$ is optimal but with an erroneous negative point estimate $\hat{\theta}_2 < 0 < \theta_2$, a greedy strategy may repeatedly select action $1$ and never improve its estimate for action $2$.
The greedy strategy faces similar difficulties with a reward function of the form $r_t = z_{t, x_t} \theta_{x_t} + \overline{\theta}_{x_t} + w_t$,
that entails learning offset parameters $\overline{\theta}_1$ and $\overline{\theta}_2$, even if context variables are standard-Gaussian-distributed.
For example, if $\overline{\theta}_1 < \overline{\theta}_2$ and $\overline{\theta}_2$ is sufficiently underestimated, as the distributions of $\theta_1$ and $\theta_2$
concentrate around $0$, a greedy strategy takes increasingly long to recover.  In the extreme case where $\theta_1 = \theta_2 = 0$ with probability $1$,
the problem reduces to one with independent actions and Gaussian noise, and the greedy policy may never recover.
It is worth noting that the news recommendation problem of Section \ref{sec:contextualRecommendation} involves a contextual bandit that
embodies both binary context variables and offset parameters.

\subsection{Problems that do not Require Exploitation}

At the other extreme, TS may also be a poor choice for problems that do not require exploitation. For example, consider a classic simulation optimization problem. Given a realistic simulator of some stochastic system, we  may like to identify, among a finite set of actions, the best according to a given objective function.  Simulation can be expensive, so we would like to intelligently and adaptively allocate simulation effort so the best choice can be rapidly identified. Though this problem requires intelligent exploration, this does not need to be balanced with a desire to accrue high rewards while experimenting. This problem is called \emph{ranking and selection} in the simulation optimization community and either \emph{best arm identification} or a \emph{pure-exploration} problem in the multi-armed bandit literature.  It can be possible to perform much better than TS for such problems. The issue is that once TS is fairly confident of which action is best, it exploits this knowledge and plays that action in nearly all periods. As a result, it is very slow to refine its knowledge of alternative actions. Thankfully, as shown by \citep{russo2016simple}, there is a simple modification to TS that addresses this issue. The resulting pure exploration variant of TS dramatically outperforms standard TS, and is in some sense asymptotically optimal for this best-arm identification problem. It is worth highlighting that although TS is often applied to A/B testing problems, this pure exploration variant of the algorithm may be a more appropriate choice.

\subsection{Time Sensitivity}

TS is effective at minimizing the exploration costs required to converge on an optimal action. It may perform poorly, however, in time-sensitive learning problems where it is better to exploit a high performing suboptimal action than to invest resources exploring actions that might offer slightly improved performance.
The following example from \citep{russo2018time} illustrates the issue.
\begin{example}{(Many-Armed Deterministic Bandit)}
	Consider an action set $\mathcal{X}=\{1,\ldots,K\}$ and a $K$-dimensional parameter vector $\theta$ with independent components, each distributed uniformly over $[0,1]$.  Each action $x$ results in reward $\theta_x$, which is deterministic conditioned on $\theta$.
	As $K$ grows, it takes longer to identify the optimal action $x^* = \argmax_{x \in \mathcal{X}} \theta_x$.  Indeed, for any algorithm, $\Prob(x^* \in \{x_1,\ldots x_t\}) \leq t/K$.
	Therefore, no algorithm can expect to select $x^*$ within time $t\ll K$.  On the other hand, by simply selecting actions in order, with $x_1 = 1, x_2 = 2, x_3 = 3, \ldots$,
	the agent can expect to identify an $\epsilon$-optimal action within $t = 1/\epsilon$ time periods, independent of $K$.
\end{example}
Applied to this example, TS is likely to sample a new action in each time period so long as $t \ll K$.  The problem with this is most pronounced in the asymptotic regime of $K \to \infty$,
for which TS never repeats any action because, at any point in time, there will be actions better than those previously selected.
It is disconcerting that TS can be so dramatically outperformed by a simple variation: settle for the first action $x$ for which $\theta_x \geq 1-\epsilon$.

While stylized, the above example captures the essence of a basic dilemma faced in all decision problems and not adequately addressed by TS.
The underlying issue is time preference.  In particular, if an agent is only concerned about performance over an
asymptotically long time horizon, it is reasonable to aim at learning $x^*$, while this can be a bad idea if shorter term performance matters
and a satisficing action can be learned more quickly.


Related issues also arise in the nonstationary learning problems described in Section \ref{se:nonstationary}.  As a nonstationary system evolves, past observations become irrelevant to optimizing future performance. In such cases, it may be impossible to converge on the current optimal action before the system changes substantially, and the algorithms presented in Section \ref{se:nonstationary} might perform better if they are modified to explore less aggressively.

Interestingly, the information theoretic regret bounds described in the previous subsection also point to this potential shortcoming of TS. Indeed, the regret bounds there depend on the entropy of the optimal action $H(A^*)$, which may tend to infinity as the number of actions grows, reflecting the enormous quantity of information needed to identify the exact optimum.  This issue is discussed further in \citep{russo2018time}. That paper proposes and analyzes \emph{satisficing TS}, a variant of TS that is designed to minimize exploration costs required to identify an action that is sufficiently close to optimal.

\subsection{Problems Requiring Careful Assessment of Information Gain}

TS is well suited to problems where the best way to learn which action is optimal is to test the most promising actions. However, there are natural problems where such a strategy is far from optimal, and efficient learning requires a more careful assessment of the information actions provide. The following example from \citep{russo2018IDS} highlights this point.
\begin{example}(A Revealing Action)
	Suppose there are $k+1$ actions $\{0,1,...,k\}$, and $\theta$ is an unknown parameter drawn uniformly at random from $\Theta=\{1,..,k\}$. Rewards are deterministic conditioned on $\theta$, and when played action $i \in \{1,...,k\}$ always yields reward 1 if $\theta=i$ and $0$ otherwise. Action 0 is a special ``revealing'' action that yields reward $1/2\theta$ when played.
\end{example}
Note that action 0 is known to never yield the maximal reward, and is therefore never selected by TS. Instead, TS will select among actions $\{1, ..., k\}$, ruling out only a single action at a time until a reward 1 is earned and the optimal action is identified. A more intelligent algorithm for this problem would recognize that although action 0 cannot yield the maximal reward, sampling it is valuable because of the information it provides about other actions. Indeed, by sampling action $0$ in the first period, the decision maker immediately learns the value of $\theta$, and can exploit that knowledge to play the optimal action in all subsequent periods.

The shortcoming of TS in the above example can be interpreted through the lens of the information ratio \eqref{eq:info ratio}. For this problem, the information ratio when actions are sampled by TS is far from the minimum possible, reflecting that it is possible to a acquire information at a much lower cost per bit.  The following two examples, also from \citep{russo2018IDS}, illustrate a broader range of problems for which TS suffers in this manner.  The first illustrates issues that arise with sparse linear models.
\begin{example}
\label{ex:sparse}
(Sparse Linear Model)
Consider a linear bandit problem where $\mathcal{X} \subset \mathbb{R}^d$ and the reward from an action $x \in \mathcal{X}$ is $x^T \theta$, which is deterministic conditioned on $\theta$. The true parameter $\theta$ is known to be drawn uniformly at random from the set of one-hot vectors $\Theta=\{ \theta' \in \{0,1 \}^d : \| \theta' \|_0 =1 \}$. For simplicity, assume $d$ is an integer power of two. The action set is taken to be the set of nonzero vectors in $\{ 0,1 \}^d$, normalized so that components of each vector sum to one: $\mathcal{X} = \left\{ \frac{x}{\|x\|_1} : x\in \{0,1  \}^d, x\neq 0 \right\}$.
\end{example}
Let $i^*$ be the index for which $\theta_{i^*} = 1$.  This bandit problem amounts to a search for $i^*$.
When an action $x_t$ is selected the observed reward $r_t=x_t^T \theta$ is positive if $i^*$ is in the support of $x_t$ or $0$ otherwise.  Given that actions in $\mathcal{X}$ can support any subset of indices, $i^*$ can be found via a bisection search, which requires $\log(d)$ periods in expectation.  On the other hand, TS selects exclusively from the set of actions that could be optimal.  This includes only one-hot vectors.  Each such action results in either ruling out one index or identifying $i^*$.  As such, the search carried out by TS requires $d/2$ periods in expectation.

Our final example involves an assortment optimization problem.
\begin{example}
(Assortment Optimization)
\label{ex:assortment}
Consider the problem of repeatedly recommending an assortment of products to a customer. The customer has unknown type $\theta \in \Theta$ where $|\Theta|=n$. Each product is geared toward customers of a particular type, and the assortment of $m$ products offered is characterized by the vector of product types $x \in \mathcal{X} = \Theta^m$. We model customer responses through a random utility model in which customers are more likely to derive high value from a product geared toward their type.  When offered an assortment of products $x$, the customer associates with the $i$th product utility $u_{\theta i t}(x)= \beta \mathbf{1}_\theta(x_i) + w_{it}$, where $\mathbf{1}_\theta$ indicates whether its argument is $\theta$, $w_{it}$ follows a standard Gumbel distribution, and $\beta \in \mathbb{R}$ is a known constant. This is a standard multinomial logit discrete choice model.  The probability a customer of type $\theta$ chooses product $i$ is given by
$$\frac{\exp\left(\beta \mathbf{1}_\theta(x_i)\right)}{\sum_{j=1}^m \exp\left(\beta \mathbf{1}_\theta(x_j)\right)}. $$
When an assortment $x_t$ is offered at time $t$, the customer makes a choice $i_t = \arg \max _i u_{\theta i t}(x)$ and leaves a review $u_{\theta i_t t}(x)$ indicating the utility derived from the product, both of which are observed by the recommendation system. The reward to the recommendation system is the normalized utility $u_{\theta i_t t}(x) / \beta$.
\end{example}
If the type $\theta$ of the customer were known, then the optimal recommendation would be $x^*=(\theta, \theta, \ldots,\theta)$, which consists only of products targeted at the customer's type. Therefore, TS would only ever offer assortments consisting of a single type of product. Because of this, TS requires $n$ samples in expectation to learn the customer's true type.
However, as discussed in \citep{russo2018IDS}, learning can be dramatically accelerated through offering {\it diverse} assortments.
To see why, suppose that $\theta$ is drawn uniformly at random from $\Theta$ and consider the limiting case where $\beta \rightarrow \infty$. In this regime, the probability a customer  chooses a product of type $\theta$ if it is available tends to $1$, and the normalized review $\beta^{-1} u_{\theta i_t t}(x)$ tends to  $\mathbf{1}_\theta(x_{i_t})$, an indicator for whether the chosen product is of type $\theta$.  While the customer type remains unknown, offering a diverse assortment, consisting of $m$ different and previously untested product types, will maximize both immediate expected reward and information gain, since this attains the highest probability of containing a product of type $\theta$. The customer's response almost perfectly indicates whether one of those items is of type $\theta$.  By continuing to offer such assortments until identifying the customer type, with extremely high probability, an algorithm can learn the type within $\left \lceil{n/m}\right \rceil$ periods.  As such, diversification can accelerate learning by a factor of $m$ relative to TS.

In each of the three examples of this section, TS fails to explore in any reasonably intelligent manner.
\citet{russo2018IDS} propose an alternative algorithm -- information-directed sampling -- that samples actions in a manner that minimizes the information ratio, and this addresses the shortcomings of TS in these examples.  It is worth mentioning, however, that despite possible advantages, information-directed sampling requires more complex computations and may not be practical across the range of applications for which TS is well-suited.

\section{Alternative Approaches}
Much of the the work on multi-armed bandit problems has focused on problems with a finite number of independent actions, like the beta-Bernoulli bandit of Example \ref{ex:beta-bernoulli}. For such problems, for the objective of maximizing expected discounted reward, the Gittins index theorem \citep{gittins1979dynamic} characterizes an optimal strategy.  This strategy can be implemented via solving a dynamic program for action in each period, as explained in \citep{katehakis1987bandit}, but this is computationally onerous relative to TS. For more complicated problems, the Gittins index theorem fails to hold, and computing optimal actions is typically infeasible. A thorough treatment of Gittins indices is provided in \citep{gittins2011multi}.

Upper-confidence-bound algorithms, as discussed in Section \ref{se:regret}, offer another approach to efficient exploration.  At a high level, these algorithms are similar to TS, in that they continue sampling all promising actions while gradually discarding those that underperform.  Section \ref{se:regret} also discusses a more formal relation between the two approaches, as originally established in \citep{russo2014learning}.  UCB algorithms have been proposed for a variety of problems, including bandit problems with independent actions \citep{lai1985asymptotically, auer2002finite, KL-UCB2013, kaufmann2012bayesian}, linear bandit problems \citep{dani2008stochastic, rusmevichientong2010linearly},  bandits with continuous action spaces and smooth reward functions \citep{kleinberg2008multi, bubeck2011xarmed, srinivas2012information}, and exploration in reinforcement learning \citep{jaksch2010near}.  As discussed, for example, in \citep{russo2014learning,OsbandRLDM2017,osband2017posterior}, the design of upper-confidence bounds that simultaneously accommodate both statistical and computational efficiency often poses a challenge, leading to use of UCB algorithms that sacrifice statistical efficiency relative to TS.

Information-directed sampling \citep{russo2014nipsIDS} aims to better manage the trade-off between immediate reward and information acquired by sampling an action through minimizing the information ratio.  The knowledge gradient algorithm \citep{frazier2008knowledge, frazier2009knowledge} and several other heuristics presented in \citep{francetich2017toolkitb,francetich2017toolkita} similarly aim to more carefully assess the value of information and also address time-sensitivity.
Finally, there is a large literature on online decision problems in adversarial environments, which we will not review here; see \citep{bubeck2012regret} for thorough coverage.


\begin{acknowledgements}
This work was generously supported by a research grant from Boeing, a Marketing Research Award from Adobe, and Stanford Graduate Fellowships courtesy of Burt and Deedee McMurty, PACCAR, and Sequoia Capital.  We thank Stephen Boyd, Michael Jordan, Susan Murphy, David Tse, and the anonymous reviewers for helpful suggestions, and Roland Heller, Xiuyuan Lu, Luis Neumann, Vincent Tan, and Carrie Wu for pointing out typos.
\end{acknowledgements}

\backmatter  

\printbibliography

\end{document}